\documentclass{article}
\usepackage{iclr2026_conference,times}
\usepackage{amsmath,amsfonts,bm}

\def\eqref#1{equation~\ref{#1}}
\def\1{\bm{1}}

\DeclareMathAlphabet{\mathsfit}{\encodingdefault}{\sfdefault}{m}{sl}
\SetMathAlphabet{\mathsfit}{bold}{\encodingdefault}{\sfdefault}{bx}{n}

\usepackage[utf8]{inputenc}
\usepackage[T1]{fontenc}
\usepackage{url}
\usepackage{booktabs}
\usepackage{amsfonts}
\usepackage{nicefrac}
\usepackage{microtype}
\usepackage{xcolor}      
\usepackage{graphicx}
\usepackage{bbm}
\usepackage{bm}
\usepackage{amsmath}
\usepackage{amssymb}
\usepackage{wrapfig}
\usepackage{multirow}
\usepackage{tabularx}
\usepackage{adjustbox}
\usepackage{subcaption}
\usepackage[font=small,labelfont=bf]{caption}
\usepackage{pifont}
\usepackage{gensymb}
\usepackage[accsupp]{axessibility}
\usepackage{svg}
\usepackage{cuted}
\usepackage{tabu}
\usepackage{mwe}
\usepackage{bm}
\usepackage{textcomp}
\usepackage{makecell}
\usepackage{grffile}
\usepackage{rotating}
\usepackage{colortbl}
\usepackage{siunitx}
\usepackage{ulem}
\usepackage{tocloft}
\usepackage{etoc} 
\usepackage{enumitem}
\usepackage[svgnames]{xcolor}
\definecolor{crimson}{rgb}{0.86, 0.08, 0.24}
\definecolor{lightblue}{rgb}{0.90, 0.95, 1}
\definecolor{lightred}{rgb}{1, 0.9, 0.95}
\definecolor{tabfirst}{rgb}{1, 0.7, 0.7}
\definecolor{tabsecond}{rgb}{1, 0.85, 0.7}
\definecolor{tabthird}{rgb}{1, 1, 0.7}
\definecolor{darkred}{RGB}{159,0,0} 
\usepackage[colorlinks=true,
            citecolor=cyan,
            pdfborder={0 0 0}
           ]{hyperref}
\usepackage{hyperref}
\hypersetup{linkcolor=red}%linkbordercolor=red,
\newcommand{\logo}{\raisebox{-0.5em}{\includegraphics[height=2em]{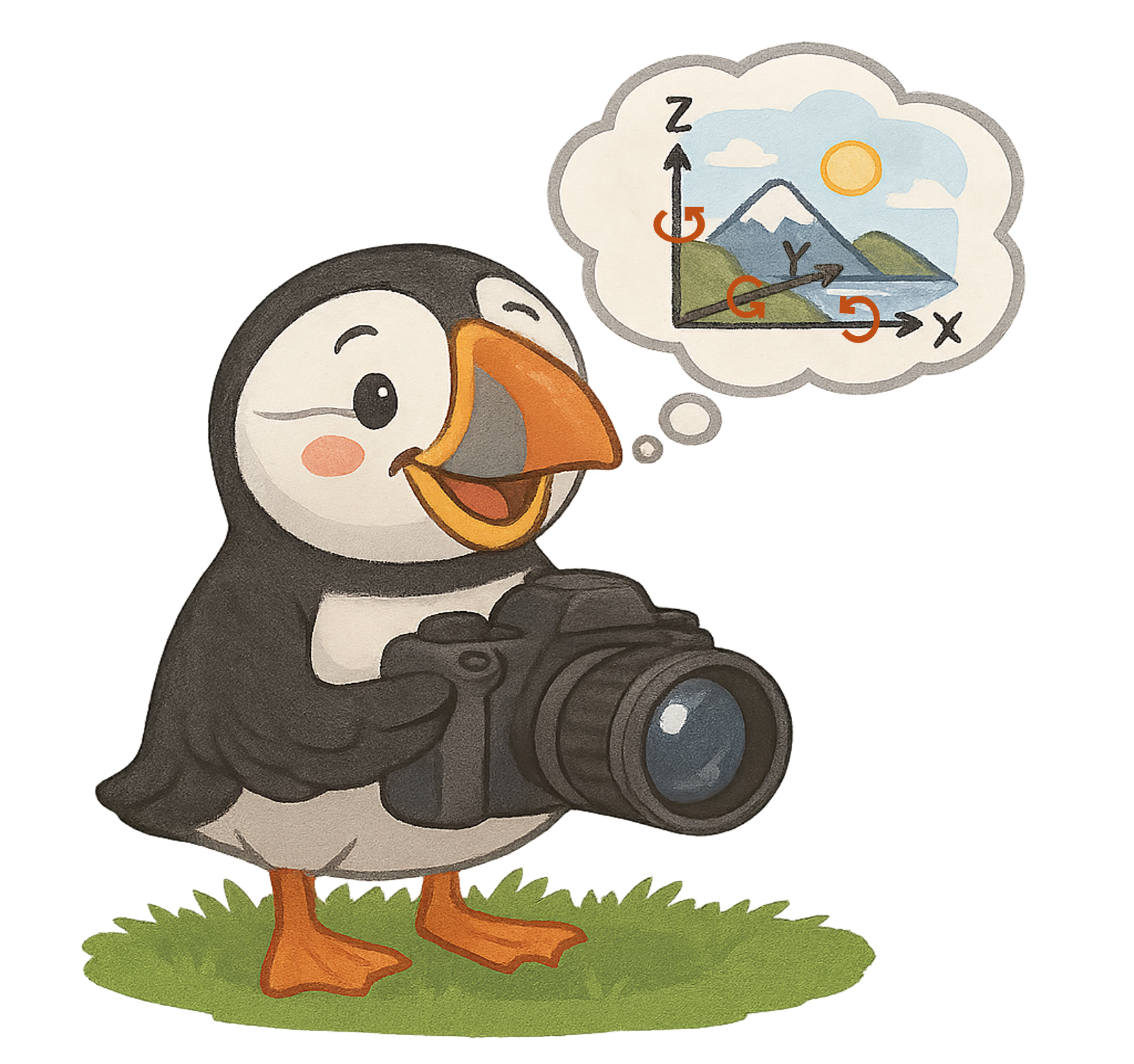}}}

\newcommand{\0}{\phantom{0}}

\renewcommand{\paragraph}[1]{\vskip4pt \noindent\textbf{#1}}

\definecolor{tabfirst}{rgb}{0.7, 0.85, 0.95}
\definecolor{tabsecond}{rgb}{0.88, 0.95, 1.0}
\definecolor{tabthird}{rgb}{1, 1, 0.7}
\newcommand{\cthird}{}
\newcommand{\cfirst}{\cellcolor{tabfirst}\bfseries}
\newcommand{\csecond}{\cellcolor{tabsecond}}
\newcommand{\greencheck}{\textcolor{green}{\ding{52}}}
\newcommand{\redcheck}{\textcolor{red}{\ding{55}}}

\makeatletter
\renewcommand\@fnsymbol[1]{}
\makeatother

\title{
\vspace{-0.5cm}
\begin{center}
\logo\
Thinking with Camera: A Unified Multimodal Model for Camera-Centric Understanding and Generation
\end{center}
}
\vspace{-3cm}
\author{
\centerline{
Kang Liao\textsuperscript{\rm 1}\qquad
Size Wu\textsuperscript{\rm 1}\qquad
Zhonghua Wu\textsuperscript{\rm 2}\qquad
Linyi Jin\textsuperscript{\rm 3} \qquad
}\\
\centerline{
\textbf{
Chao Wang\textsuperscript{\rm 4} \qquad
Yikai Wang\textsuperscript{\rm 1} \qquad
Fei Wang\textsuperscript{\rm 2} \qquad
Wei Li\textsuperscript{\rm 1 $\dagger$}\qquad
Chen Change Loy\textsuperscript{\rm 1 $\dagger$} \thanks{$^\dagger$ Corresponding authors}
}}\\
\centerline{
\textsuperscript{\rm 1}S-Lab, Nanyang Technological University \quad
\textsuperscript{\rm 2}SenseTime Research
}\\
\centerline{
\textsuperscript{\rm 3}University of Michigan \quad
\textsuperscript{\rm 4}Max-Planck Institute for Informatics
}\\
}

\iclrfinalcopy 
\begin{document}

\maketitle

\vspace{-0.9cm}
\begin{figure}[!htbp]
    \centering
    \begin{subfigure}{1\textwidth}
        \centering
        \includegraphics[width=1\linewidth]{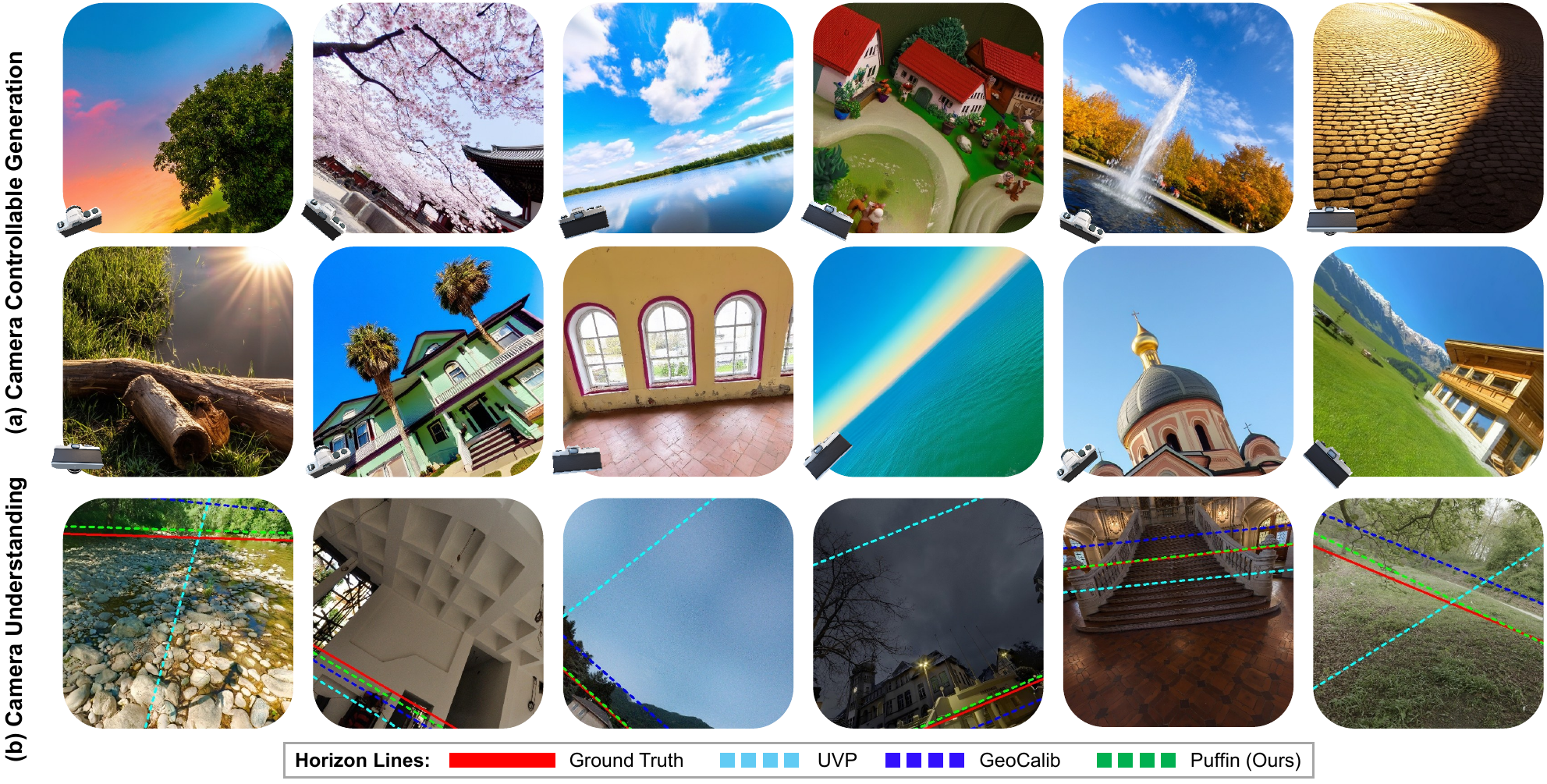}
    \end{subfigure}
    
    %\vspace{0.1cm}
    
    \begin{subfigure}{1\textwidth}
        \centering
        \includegraphics[width=1\linewidth]{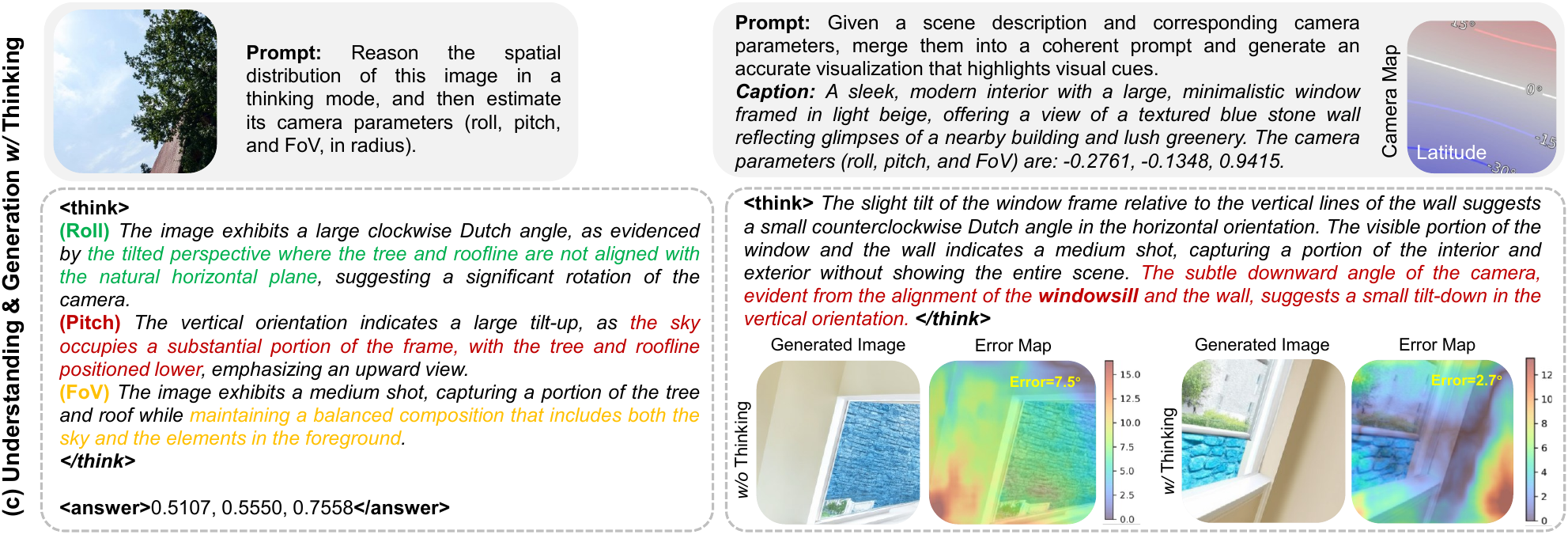}
    \end{subfigure}
    
    \vspace{-0.1cm}
    \begin{subfigure}{1\textwidth}
        \centering
        \includegraphics[width=1\linewidth]{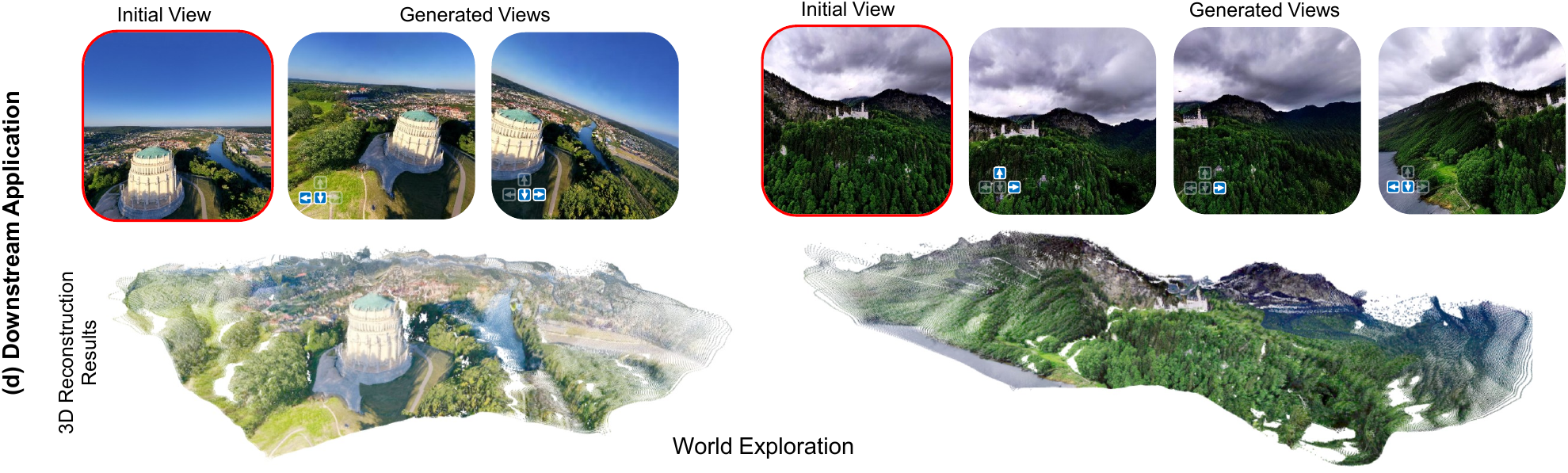}

    \end{subfigure}
    \vspace{-0.3cm}
    \caption{\textbf{Illustration of the versatile capabilities of our Puffin model. It unifies camera-centric generation (a) and understanding (b), supports the thinking mode (c), and enables diverse 3D applications (d).}
    }
    \label{fig:teaser}
\end{figure}

\begin{abstract}
Camera-centric understanding and generation are two cornerstones of spatial intelligence, yet they are typically studied in isolation.
We present \textbf{Puffin}, a unified camera-centric multimodal model that extends spatial awareness along the camera dimension. Puffin integrates language regression and diffusion-based generation to interpret and create scenes from arbitrary viewpoints.
To bridge the modality gap between cameras and vision-language, we introduce a novel paradigm that treats \textit{camera as language}, enabling \textit{thinking with camera}. This guides the model to align spatially grounded visual cues with photographic terminology while reasoning across geometric context.
Puffin is trained on \textbf{Puffin-4M}, a large-scale dataset of 4 million vision-language-camera triplets. We incorporate both global camera parameters and pixel-wise camera maps, yielding flexible and reliable spatial generation.
Experiments demonstrate Puffin’s superior performance over specialized models for camera-centric generation and understanding. With our designed instruction tuning, Puffin generalizes to diverse cross-view tasks such as spatial imagination, world exploration, and photography guidance. To advance multimodal spatial intelligence research, we released the code, models, dataset pipeline, and benchmark at \url{https://kangliao929.github.io/projects/puffin}.
\end{abstract}

\section{Introduction}

For machines, cameras serve as the primary interface to the physical world, enabling spatial intelligence that underlies applications such as robotics, AR/VR, and autonomous driving. In general, two principal camera-centric objectives work in tandem to enable machines to perceive and interact with their spatial context. On the one hand, \textit{understanding} the camera geometry from images~\citep{pollefeys1999self, hold2018perceptual, liao2023deep, zhang2024cameras, veicht2024geocalib, lin2025towards}, namely how the 3D world is projected onto the 2D image plane, lays the foundation for machines to recover spatial structure and navigate complex environments. On the other hand, by modulating intrinsic and extrinsic parameters, cameras encode spatial relationships and offer flexible control over spatial content \textit{generation}~\citep{he2024cameractrl, wang2024motionctrl, ren2025gen3c, yuan2025generative, bernal2025precisecam, genie3}, which simulates how the world appears from any viewpoint or orientation. To date, these two perspectives have been commonly treated as isolated problems and independently explored by the research community.

In this work, we make \textit{the first attempt} to unify camera-centric understanding and generation in a cohesive framework. Motivated by recent progress in unified understanding and generation with large multimodal models (LMMs)
~\citep{team2024chameleon, wu2025harmonizing, wu2024vila, pan2025transfer, wu2025janus}, we extend this paradigm to the spatial domain, where camera geometry plays a central role.
However, unlike language or images, camera parameters are abstract and non-intuitive: they describe field-of-view (FoV), orientation, or perspective in numerical form rather than semantic content. This discrepancy introduces a modality gap when integrating cameras into LMMs. For instance, when users specify ``20\degree \ roll'' or ``35mm lens'' for controllable generation, existing models often ignore or misinterpret such cues, pursuing semantic alignment while neglecting precise spatial control. Similarly, current LMMs tend to collapse geometric details into coarse representations when understanding camera information, leading to spatially inconsistent outputs. As a result, naïvely extending LMMs cannot resolve conflicts between modalities, producing suboptimal performance in both tasks.

To address this challenge, we introduce \textbf{Puffin}, a unified multimodal framework that interprets cameras as a first-class modality. Puffin combines autoregressive and diffusion modeling to jointly perform camera-centric understanding and generation\footnote{We mainly focus on single-view calibration and text-to-image controllable generation, but Puffin can be flexibly extended to cross-view understanding and generation via our designed instruction tuning (see Figure~\ref{fig:application}).}. Instead of treating camera parameters as auxiliary labels, Puffin introduces the notion of \textit{thinking with camera}, aligning spatially grounded visual cues with professional photographic terminology while reasoning over geometric context. This design provides a shared chain-of-thought across multimodal tasks, enabling spatially consistent understanding and controllably aligned generation.

To support this framework, we construct \textbf{Puffin-4M}, a large-scale dataset of 4 million vision-language-camera triplets. Puffin-4M includes single-view images with precise camera parameters, descriptive captions, pixel-wise camera maps, and spatial reasoning annotations across diverse indoor and outdoor scenarios. Beyond single views, it also incorporates cross-view and aesthetic images, making it a versatile benchmark for both understanding and generation tasks. 

Experimental results show Puffin outperforms specialized models for camera-centric understanding or generation, and can be adapted to diverse downstream applications. We illustrate the versatile capabilities of our Puffin model in Figure~\ref{fig:teaser}. In each generated image (a), the target camera is marked at the bottom left, and the horizon lines are visualized from the estimated camera parameters (b). For world exploration (d), we visualize 3D reconstruction results derived from the initial and generated views. Our main contributions are threefold:

\begin{itemize}
    \item We make \textit{the first attempt} to seamlessly integrate camera geometry into a unified multimodal model, introducing a camera-centric framework to advance multimodal spatial intelligence.
    \item We propose \textit{thinking with camera}, a novel mechanism that guides the model to align spatially grounded visual cues with photographic terminology, bridging the modality gap between camera and vision-language and enabling effective spatial reasoning.
    \item We construct \textbf{Puffin-4M}, a large-scale dataset of 4M vision-language-camera triplets spanning diverse indoor and outdoor scenes, and establish a comprehensive benchmark for evaluating camera-centric multimodal models.
\end{itemize}
\section{Related Work}
\label{sec:related_work}
\noindent\textbf{Large Multimodal Models.} Built upon a visual encoder~\citep{radford2021learning, zhai2023sigmoid, tschannen2025siglip} and a large language model (LLM)~\citep{touvron2023llama, qwen2024qwen2, cai2024internlm2, liu2024deepseek}, LMMs~\citep{liu2023visual, chen2024sharegpt4v, tong2024cambrian, bai2025qwen2, zhu2025internvl3} process mixed visual and textual inputs and perform understanding and reasoning via language generation. Fueled by large-scale pre-training of the vision and language models and sophisticated instruction-tuning, LMMs excel at high-level understanding tasks, such as object localization, counting, and optical character recognition. However, these models, optimized for semantic alignment between vision and language, remain limited in capturing image intrinsics (\textit{e.g.}, depth and geometry), which constrains their ability in camera understanding and spatial reasoning. To bridge this gap, it is crucial to enrich LMMs with geometry-aware prior knowledge that preserves structural details beyond semantics. Moreover, aligning such geometric cues with linguistic tokens provides a pathway to extend the reasoning capacity of LMMs from abstract semantics to physically grounded spatial understanding.

\noindent\textbf{Unified Multimodal Models.} As an extension of standard LMMs, unified multimodal models~\citep{team2024chameleon, wang2024emu3, tang2025ugen, wu2025harmonizing, lin2025toklip, wu2024vila, tong2024metamorph, pan2025transfer, lin2025uniworld, wu2025openuni, chen2025blip3, wu2025janus, xie2024show, xie2025show} jointly learn visual understanding and generation within a single framework. Two main design philosophies are typically adopted. One line of work formulates visual generation as autoregression over either discrete~\citep{team2024chameleon, wu2024vila, wang2024emu3, wu2025janus} or continuous~\citep{fan2025unified} image tokens, sharing LLM parameters for both understanding and generation. Another line~\citep{pan2025transfer, chen2025blip3, wu2025openuni, lin2025uniworld} aligns pre-trained LMMs with diffusion modules, enabling faster convergence and lower training cost. While both types of models advance general image understanding and generation, they are constrained to simplistic camera assumptions (\textit{e.g.}, fixed front-view, predefined FoVs), hindering their practical applicability to realistic and complex environments. To this end, we introduce a camera-centric framework that jointly performs camera understanding and controllable generation.

\noindent\textbf{Camera Geometry from Vision.} Tasks such as camera calibration and pose estimation have long been a central topic in 3D vision~\citep{pollefeys1999self, hartley2003multiple, liao2023deep, veicht2024geocalib, hold2018perceptual, jin2023perspective, zhang2024cameras, lin2025towards}. While earlier learning-based works attempted to directly regress camera parameters from input images~\citep{hold2018perceptual, workman2015deepfocal, bogdan2018deepcalib, zhai2016detecting, kendall2015posenet}, recent advances increasingly explore the use of intermediate representations or geometry fields to bridge the prediction gap. Representative approaches~\citep{lee2020neural, lee2021ctrl, Song2024MSCC, janampa2024sofi, yin2018fisheyerecnet} leverage geometric structures or semantic features to alleviate the inherent difficulty of inferring camera parameters from a few views. Building on priors of the camera model and the perspective properties of captured images, a growing body of methods proposes to learn dense geometry fields, such as distortion distribution maps~\citep{liao2020model, liao2021deep}, pixel displacement fields~\citep{li2019blind, liao2025mowa, xie2025aligndiff}, camera rays~\citep{zhang2024cameras}, perspective fields~\citep{jin2023perspective, veicht2024geocalib, tirado2025anycalib}, or incidence fields~\citep{zhu2023tame, he2025diffcalib, deng2024boost}. However, such representations typically emphasize low-/mid-level patterns, limiting their ability to capture a holistic and coherent spatial concept. Rather than pursuing better representations, this work explores an alternative perspective: interpreting the camera as language.

\section{Camera-Centric Unified Multimodal Model}

\begin{figure}
    \centering
    \includegraphics[width=1\linewidth]{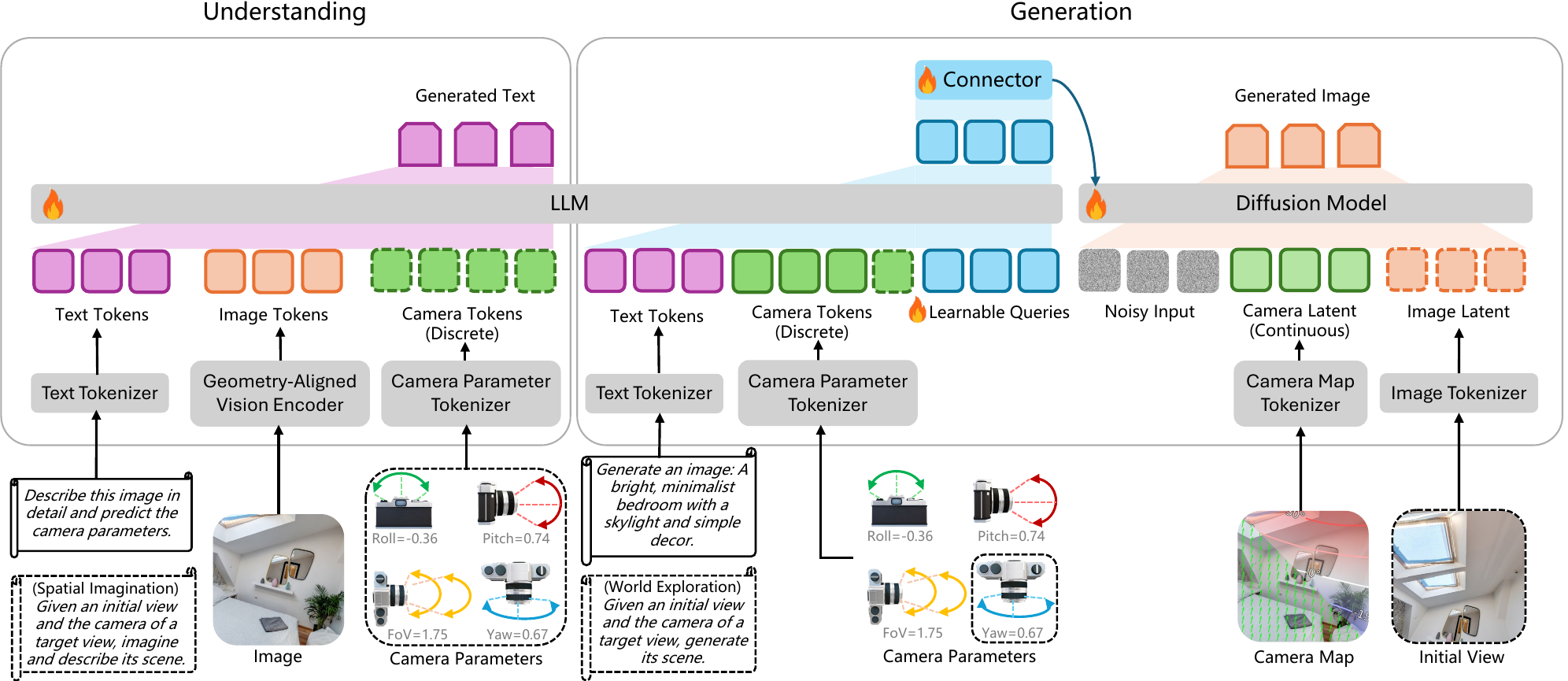}
    \caption{\textbf{Overview of the proposed Puffin.} It jointly learns the camera-centric understanding and generation tasks in a unified multimodal framework. The elements bounded with dotted boundaries represent the cross-view understanding and generation during instruction tuning, such as spatial imagination and world exploration.
    } 
    \label{fig:framework}
    %\vspace{-0.4cm}
\end{figure}

Puffin, as illustrated in Figure~\ref{fig:framework}, unifies camera-centric understanding and generation within a multimodal paradigm. For understanding, we introduce a geometry-aligned vision encoder to a large language model (LLM) to retain rich geometric features and enhance the model’s capacity for spatial analysis. For generation, a connector module learns to map the hidden states of the LLM (via a set of learnable queries) into conditioning signals that can be interpreted by the diffusion model. To facilitate the integration of camera geometry, apart from the discrete camera tokens derived from numerical camera parameters, we introduce continuous camera latent obtained from pixel-wise camera maps, allowing fine-grained spatial control in image generation.

\subsection{Camera Understanding}
\label{sec:thinking_und}
\noindent\textbf{Definition.} In this work, camera understanding is formulated as a question-answering task conditioned on image content. The generated text consists of a concise description or spatial reasoning along with the estimated camera parameters (\textit{i.e.}, roll, pitch, FoV) of the input image. Unlike previous methods that directly estimate the parameters from images, our approach integrates camera geometry within the text and performs next-token prediction in a multimodal sequence modeling paradigm.

\begin{figure}[t]
    \centering
    \includegraphics[width=1\linewidth]{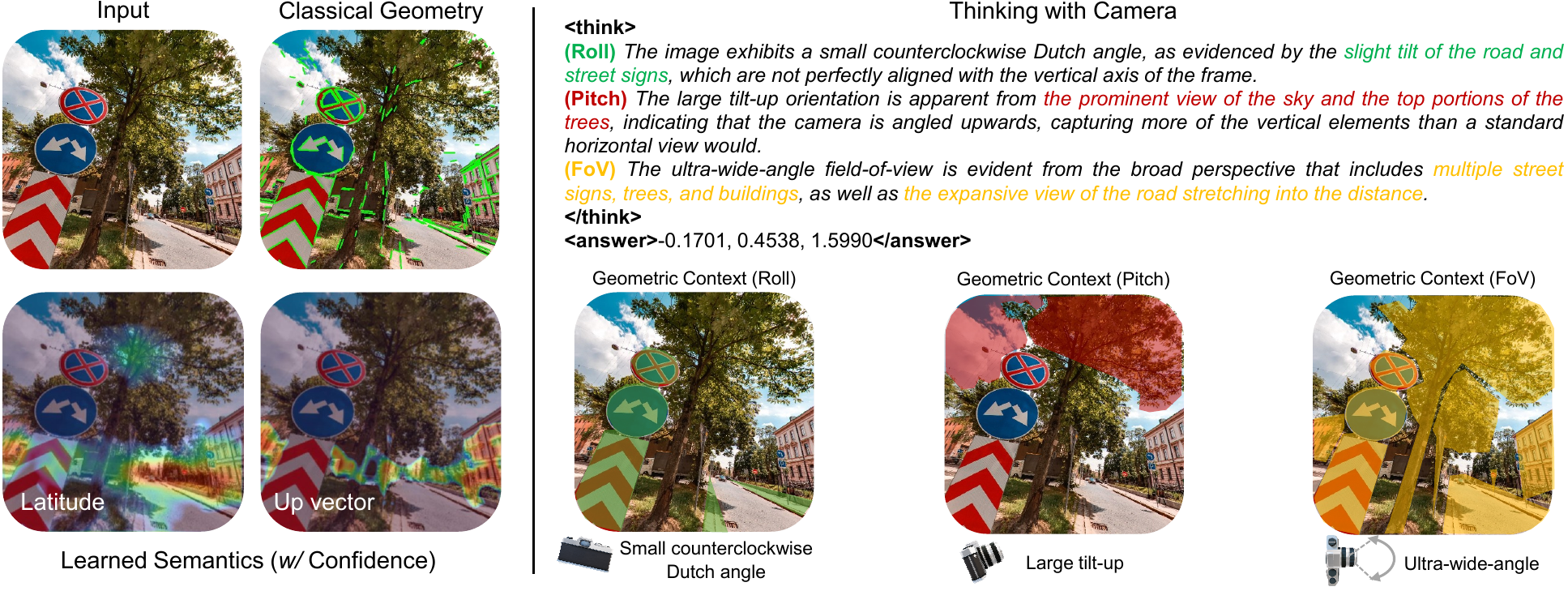}
     \caption{\textbf{Methods for learning camera geometry.} (Left) Previous classical and learning-based methods focused on extracting or learning representations such as geometric structures or semantic features (with confidence). (Right) We introduce the notion of \textit{thinking with camera} through LMMs. It first decouples the camera parameters across geometric context, establishing connections between spatially grounded visual cues (highlighted in the masked regions) and professional photographic terms. The camera parameters are then predicted within the \textbf{<answer></answer>} tag through this spatial reasoning process \textbf{<think></think>}.
     }
    \label{fig:thinking_comparison}
    %\vspace{-0.4cm}
\end{figure}

\noindent\textbf{Motivation.} As illustrated in Figure~\ref{fig:thinking_comparison} (left), previous classical and learning-based methods focus on extracting or learning representations to predict the camera parameters, such as geometric structures~\citep{pautrat2023vanishing} or semantic features with confidence estimates~\citep{veicht2024geocalib}. However, these representations often emphasize low-/mid-level patterns, limiting their ability to capture a holistic and coherent spatial concept. 
As a result, existing approaches tend to excel in scenarios with rich features but struggle to generalize across diverse visual environments. 

\noindent\textbf{Thinking.}
Instead of focusing on how to learn a representation, we propose to interpret the camera as language and introduce the notion of \textit{thinking with camera}. It guides the LMMs to align spatially grounded visual cues with photographic terminology while reasoning across geometric context. The details of each key element are elaborated below. 

\noindent\textbullet~\textit{{Spatially Grounded Visual Cues}.} 
The 3D world is governed by physical laws, where gravity and human design shape stable spatial regularities that serve as strong perceptual priors. Texture-less regions such as sky, ceilings, floors, or ground surfaces lack local features but encode vertical regularities critical for pitch estimation. Similarly, FoV estimation relies on perceiving spatial composition, including the foreground–background ratio, object scale, and depth distribution. While such properties are difficult to infer from purely visual representations, they are implicitly captured by LMMs as knowledge priors. Thus, we embed these spatially grounded visual cues into our thinking captions, enabling the model to perform explicit spatial reasoning about camera geometry. %The details of building thinking captions are presented in Section~\ref{sec:dataset}.
    
\noindent\textbullet~\textit{{Professional Photographic Terms}.} Existing LMMs typically acquire over-abstracted semantics, whereas the detailed numerical values of camera parameters are too fine-grained to estimate precisely. As a practical alternative, professional photographic terms (\textit{e.g.}, close-up, tilt-up, Dutch angle) are widely used in annotations and well aligned with LMM knowledge~\citep{liu2025shotbench, wang2025cinetechbench, lin2025towards}. Thus, we leverage them as intermediate supervisory signals to naturally bridge low-/mid-level camera geometry and high-level multimodal reasoning. These terms, derived as quantized abstractions of camera parameters, are merged with textual scene descriptions, making global spatial arrangements linguistically accessible. The parameter-to-term mapping can be formulated as $f: p \mapsto t$, in which the mapping $f$ is shown in Table~\ref{tab:photographic_term}.

\noindent\textbullet~\textit{{Geometric Context}.}
As shown in Figure~\ref{fig:thinking_comparison} (right), we decouple camera parameters across geometric context (roll, pitch, and FoV), which aligns specific spatially grounded visual cues such as sky, foreground composition, and object-level depth ordering with each professional photographic terminology. By anchoring numerical attributes to semantically meaningful descriptors, our framework bridges abstract visual features and physically interpretable geometry. The final parameters are predicted through this structured spatial reasoning.

With the above designs, we interpret the camera as language by grounding its physical attributes in stable spatial regularities. Numerical parameters are abstracted into professional photographic terms, providing a semantic vocabulary aligned with LMMs. Through this mapping, camera geometry becomes linguistically interpretable, allowing structured spatial reasoning for accurate camera parameter prediction. We visualize more reasoning results in Figure~\ref{fig:und_reasoning_vis}.

\noindent\textbf{Choosing a Suitable Vision Encoder.}
A straightforward approach to camera understanding is to fine-tune existing LMMs that couple a vision encoder with an LLM, but this naïve strategy faces two major limitations: (i) vision encoders in LMMs are primarily designed for recognition tasks and thus yield condensed features lacking geometric fidelity, and (ii) language components contain little prior knowledge of spatial perception, reducing adaptability to camera-centric tasks. As a result, such fine-tuning can lead to performance bottlenecks and even underperform pure vision-based methods (see Section~\ref{sec:ablation_study}). To overcome these issues, we introduce a \textit{geometry-aligned vision encoder} distilled from both semantic (\textit{e.g.}, CLIP, SigLIP) and vision-centric (\textit{e.g.}, DINO, SAM) teachers~\citep{heinrich2025radiov2}, offering versatile features that preserve geometric fidelity while maintaining strong semantic understanding. We then align this encoder with an LLM~\citep{qwen2024qwen2} via progressive unfreezing and joint fine-tuning. This staged optimization stabilizes training and fosters spatial awareness that bridges low-/mid-level structural cues with high-level linguistic reasoning. The detailed training recipe is provided in Section~\ref{sec:appendix_training}.

\subsection{Camera-Controllable Generation}
\textbf{Motivation.}
Unlike image understanding, image generation requires complex cross-modal alignment and the synthesis of fine-grained visual details. As discussed in Section~\ref{sec:thinking_und}, the detailed numerical values of camera parameters are too specific for current LMMs to interpret effectively, failing to faithfully capture the realistic spatial distribution required for camera-controllable generation. 

\textbf{Thinking.} To address this, we design a step-by-step process that integrates visual detail analysis with reasoning. The model first infers the potential visual cues from vanilla captions, and then uses this textual reasoning as a semantic planning stage to guide image generation. For instance, a large pitch value may correspond to an expansive sky with clouds in outdoor scenes or to pendant lights and uncluttered ceilings indoors. Beyond textual reasoning, numerical camera parameters are translated into professional photographic terms more suitable for LMMs, naturally aligning with the reasoning process in camera understanding. We therefore adopt a shared chain-of-thought mechanism between understanding and controllable generation. As shown in Figure~\ref{fig:teaser} (c), given a small pitch value and a caption describing a modern interior, our method translates the value into a photographic term (\textit{e.g.}, small tilt-down), imagines salient cues such as a windowsill, and produces more precise spatial simulation than the baseline.

\noindent\textbf{Flexible and Faithful Control.} 
The pipeline of camera-controllable generation is shown in Figure~\ref{fig:framework} (right). The key design is to incorporate pixel-wise camera maps as a continuous latent of camera geometry, apart from the discrete camera tokens derived from numerical parameters. Unlike tokens that capture only global attributes, these dense maps encode local geometric context at each pixel, including orientation and displacement cues~\citep{jin2023perspective}. By converting maps into continuous latent, the diffusion model receives fine-grained spatial priors that preserve global camera settings while adapting to subtle geometric variations, thus offering flexible control of spatial layout and viewpoint. Additionally, we introduce a connector module as an adaptive interface between the LLM and the diffusion model, where a set of learnable queries together with text and camera tokens extract and restructure LLM hidden representations, which are then projected into conditioning signals for generation~\cite{pan2025transfer, wu2025openuni}. This design enables semantic and geometric understanding from the LLM to faithfully guide the diffusion model. 

\subsection{Instruction Tuning}
Although our Puffin focuses on single-view camera calibration and text-to-image controllable generation, it can be flexibly extended to cross-view settings with only minor modifications, such as appending additional tokens and switching prompts according to the target task. As shown in Figure~\ref{fig:framework}, the dotted modules denote cross-view understanding and generation. We explore three tasks: (i) spatial imagination, where the model imagines the scene description of a target view given its camera parameters and an initial view; (ii) world exploration, where the model generates the target view, incorporating an additional yaw parameter to represent cross-view deviations and conditioning on both the target-view camera map and the VAE-encoded initial view (with text descriptions randomly dropped to support both text-conditioned and text-free generation); and (iii) photographic guidance, where the model suggests camera parameter adjustments from an initial view to achieve images with higher photographic aesthetics. Visualization results are presented in Figure~\ref{fig:application}.

\subsection{Training Recipe}
\label{sec:appendix_training}
\begin{table}[t]
\setlength\tabcolsep{11pt}%
\centering
%\footnotesize
\scriptsize
\caption{\textbf{Training recipe of Puffin.} For the data sampling ratio, we mark the data involving the spatial reasoning and instruction tuning in \colorbox{lightblue}{light blue} and \colorbox{lightred}{light red}, respectively. For clarity, we abbreviate the generation and understanding as \textit{Gen.} and \textit{Und.}.}
\begin{tabular}{l|cccc}

\toprule
 & \textbf{Stage I} & \textbf{Stage II} & \textbf{Stage III} & \textbf{Stage IV} \\
\hline
\textbf{Hyperparameters} \\
Learning rate  & $1\times10^{-4}$ & $2\times10^{-5}$ & $1\times10^{-5}$ & $5\times10^{-6}$ \\
LR Scheduler    & \multicolumn{4}{c}{Cosine} \\
Weight Decay    & \multicolumn{4}{c}{0.05} \\
Betas    & \multicolumn{4}{c}{(0.9, 0.95)} \\
Optimizer       & \multicolumn{4}{c}{AdamW} \\
Batch Size      & 1024 & 1024 & 512 & 256 \\
Training Steps  & 10K & 30K & 60K & 20K \\
Warm-up Steps   & 1K & 900 & 1.8K & 600 \\
LLM             & Frozen & Trainable & Trainable & Trainable \\
Diffusion Model & Frozen & Trainable & Trainable & Trainable \\
Vision Encoder  & Frozen & Trainable & Trainable & Frozen \\
\midrule
\textbf{Data Sampling Ratio} \\
Text-Camera$\rightarrow$Image (\textit{Gen.})           & 0.5 & 0.5 & 0.0 & 0.0 \\
Image$\rightarrow$Text-Camera (\textit{Und.})           & 0.5 & 0.5 & 0.0 & 0.0 \\
\rowcolor{lightblue}Text$\rightarrow$Text       & 0.0 & 0.0 & 0.33 & 0.0 \\
\rowcolor{lightblue}Text-Camera$\rightarrow$Image (\textit{Gen.})           & 0.0 & 0.0 & 0.33 & 0.0 \\
\rowcolor{lightblue}Image$\rightarrow$Text-Camera (\textit{Und.})           & 0.0 & 0.0 & 0.33 & 0.0 \\
\rowcolor{lightred}Image-Camera$\rightarrow$Text (Cross-view \textit{Und.})  & 0.0 & 0.0 & 0.0 & 0.4 \\
\rowcolor{lightred}Image-(Text)-Camera$\rightarrow$Image (Cross-view \textit{Gen.})     & 0.0 & 0.0 & 0.0 & 0.4 \\
\rowcolor{lightred}Image$\rightarrow$Camera (Photography \textit{Und.})  & 0.0 & 0.0 & 0.0 & 0.05 \\
\bottomrule
\end{tabular}
\label{tab:training_recipe}
\end{table}
We conduct a multi-stage training strategy, where the vision encoder, LLM, and the diffusion model are aligned in the first stage. Then, in the supervised fine-tuning (SFT) stage, the models are jointly optimized using both base and thinking datasets. Finally, an instruction-tuning stage is applied, involving various cross-view generation and understanding tasks. The details are listed in Table~\ref{tab:training_recipe}. We elaborate each training stage as follows. 

\begin{itemize}
    \item \textbf{Stage I-Alignment}. In this stage, we align the vision encoder with the LLM by training only the MLP projector for the understanding task, where the framework learns to predict both text descriptions and camera parameters from the input image. For generation, the framework takes text descriptions, camera parameters, and the camera map as inputs, and learns to synthesize the target image with the corresponding description and configuration. Specifically, we train learnable queries and a connector to bridge the LLM and the diffusion transformer, where the connector maps LLM hidden states into conditioning signals for the diffusion model. A cross-entropy loss and diffusion loss supervise the understanding and generation, respectively, while parameters of the vision encoder, LLM, and diffusion model remain frozen.
    
    \item \textbf{Stage II-SFT.} After aligning different modalities, we unfreeze all modules except the VAE and fine-tune the entire framework, using the same inputs and outputs as in Stage I. To stabilize training, we apply gradient scaling of 0.1 to the vision encoder.
    
    \item \textbf{Stage III-SFT \textit{w/} Thinking.} To further bridge the modality gap between the camera and vision-language, we introduce \textit{thinking with camera} in this stage. The implementation is the same as Stage-II, except that the training data contains spatial reasoning captions (the details of obtaining such captions are provided in Section~\ref{sec:appendix_dataset}). Beyond generation and understanding, this stage also learns the textual reasoning task, which enriches the vanilla captions with spatially grounded visual cues and translates specific camera parameter values into professional photographic terms.
    \item \textbf{Stage IV-Instruction Tuning.} Finally, we improve our model’s ability to adapt to diverse spatial configurations. In particular, three types of cross-view data are trained simultaneously, including the spatial imagination, world exploration, and photographic guidance. The KV cache mechanism is utilized in cross-view generation. The vision encoder is frozen while other modules are trainable.
\end{itemize}

We release three model variants: Puffin-Base, Puffin-Thinking, and Puffin-Instruct, to accommodate different application needs. Puffin-Base provides a foundation model for unified camera understanding and camera-controllable image generation; Puffin-Thinking enhances spatial reasoning and generation; and Puffin-Instruct is optimized by instruction tuning, supporting cross-view tasks and complex multimodal interactions.
\section{Dataset Construction}
\label{sec:dataset}
\begin{figure}[t]
    \centering
    \includegraphics[width=1\linewidth]{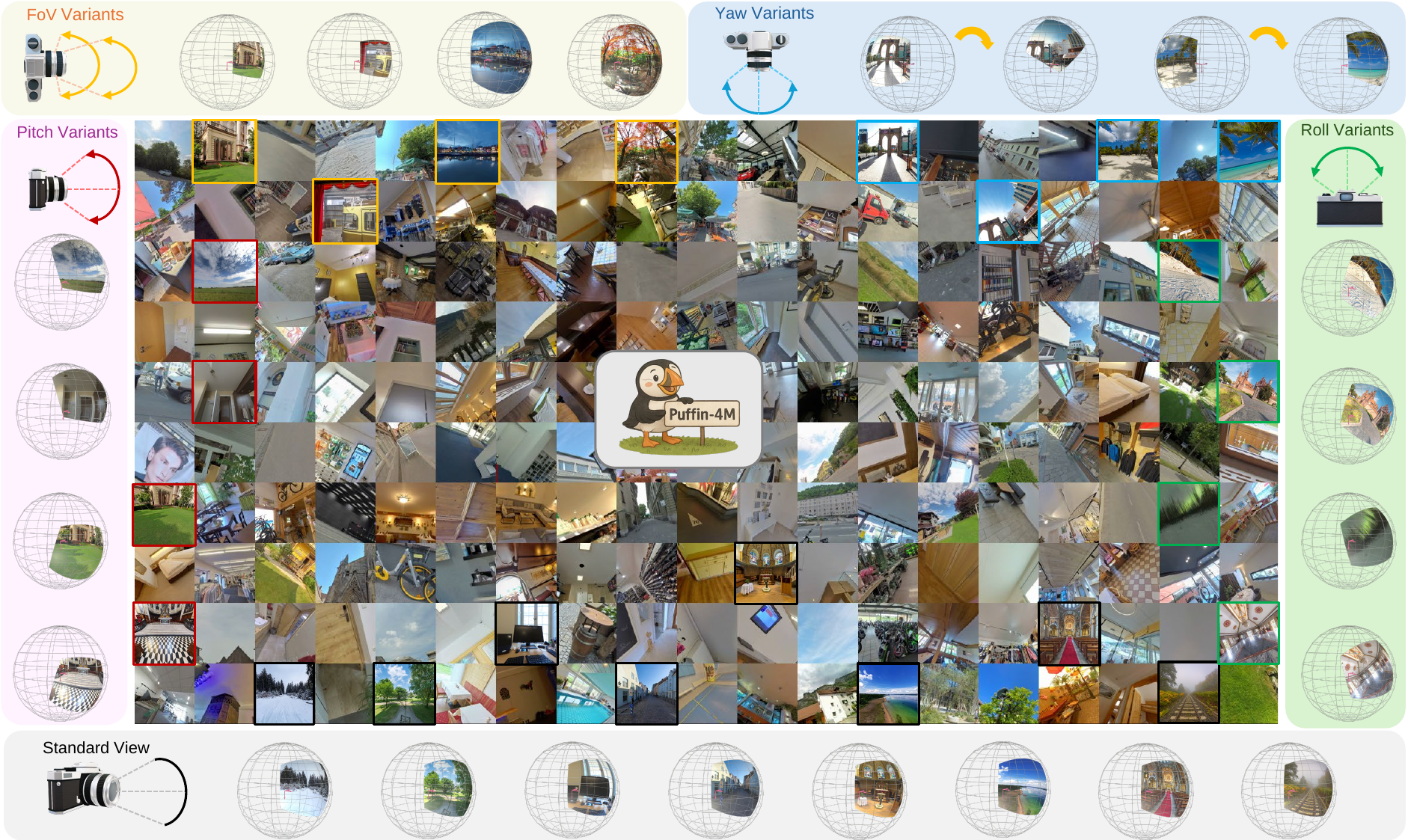}
    \caption{\textbf{Overview of the proposed Puffin-4M dataset.} It consists of 4 million vision-language-camera triplets under various scenarios and camera configurations. We mark the sample images with different colors, each denoting a different variant of the camera configurations.
    }
    \label{fig:dataset}
    %\vspace{-0.4cm}
\end{figure}
\label{sec:appendix_dataset}

\begin{figure}[t]
    \centering
    \includegraphics[width=1\linewidth]{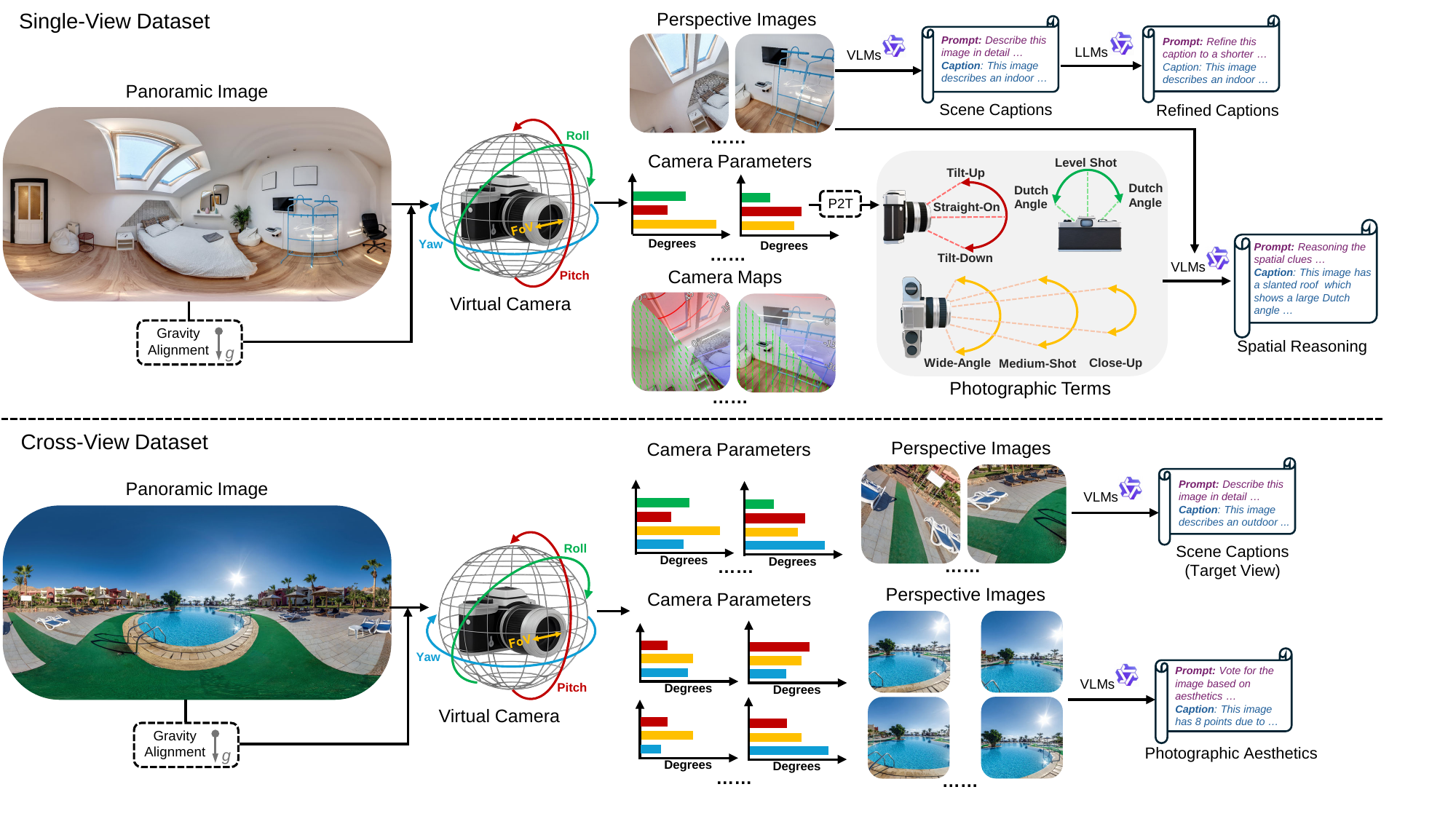}
    \caption{\textbf{Pipeline of the dataset construction.} P2T denotes the mapping from the numerical camera parameters to the professional photographic terms. For clarity, we omit the orientations ``clockwise'' and ``counterclockwise'' of the Dutch angle in photographic terms. 
    }
    \label{fig:dataset_pipeline}
    %\vspace{-0.4cm}
\end{figure}

Datasets and benchmarks that span vision, language, and camera modalities remain scarce in the domain of spatial multimodal intelligence. To address this gap, we introduce \textbf{Puffin-4M}, a large-scale, high-quality dataset comprising 4 million vision-language-camera triplets. 
The overview of our Puffin-4M is shown in Figure~\ref{fig:dataset}. The construction of this curated dataset consists of four stages: panoramic data collection and preprocessing, perspective image generation, scene and spatial reasoning captioning, and extensions for cross-view scenarios. Following previous works~\citep{veicht2024geocalib, jin2023perspective, bernal2025precisecam, hold2018perceptual}, we render the perspective images with various camera intrinsic and extrinsic parameters from $360^\circ$ panoramic images using a standard camera model. The pipeline of the dataset construction is illustrated in Figure~\ref{fig:dataset_pipeline} and the dataset comparison with previous works is listed in Table~\ref{tab:dataset_cp}. More details are described as follows.

\noindent\textbf{Panoramic Data Collection and Preprocessing.} We begin by collecting panoramic images from publicly available datasets~\citep{OmniPhotos, OmniBlender, PanDA, 360Loc, Pandora, F-360iSOD, D-SAV360, cheng2018learning, Laval, chang2018generating, veicht2024geocalib, Stanford2d3d} as well as from online platforms~\citep{Flickr360, StreetView360AtoZ, Wikimedia-Commons, HDRMAPS, Poly-Haven, AmbientCG, BlenderKit, Youtube}. In addition, we acquire a large volume of outdoor panoramic data from Google Street View~\citep{Google_street_view}, spanning 12 cities across Asia, Europe, and North America. In total, our curated dataset comprises approximately 200K high-quality panoramic images with substantial diversity. A significant portion of these images reaches 4K resolution or higher, up to 10K. However, due to variations in $360^\circ$ camera calibration and acquisition stability, some panoramas exhibit geometric distortions and misalignment. To mitigate this, we apply geometric correction techniques based on line segmentation and vanishing point estimation~\citep{jiang2022lgt, zou2018layoutnet}, aligning the panoramas with the gravity direction and improving structural consistency.

\begin{table*}[t]
\captionsetup{aboveskip=2pt}
\centering
\footnotesize
\caption{\textbf{Dataset Comparisons.} The datasets proposed in previous individual models tailored for camera understanding~\citep{lee2021ctrl, bogdan2018deepcalib, veicht2024geocalib, jin2023perspective, hold2018perceptual} or camera-controllable image generation~\citep{bernal2025precisecam} \textit{vs.} our Puffin-4M for the camera-centric unified multimodal model. In addition to its larger scale, our dataset also offers advantages in spatial reasoning captions, and cross-view image pairs. For the camera parameters, we denote the intrinsic parameters: focal length ($f$), radial distortion coefficient ($\xi$); and the extrinsic parameters: roll ($\phi$), pitch ($\theta$), yaw ($\psi$).}
\resizebox{\textwidth}{!}{%
\begin{tabular}{lcccr|ccccc}
\toprule[0.17em]
\multirow{2}{*}{\textbf{Dataset}} & \multirow{2}{*}{\textbf{Task Type}} & \multirow{2}{*}{\textbf{Intrinsics}} & \multirow{2}{*}{\textbf{Extrinsics}} & \multirow{2}{*}{\textbf{\# Frames}} & \multicolumn{5}{c}{\textbf{Details}}             \\
                         &                              &                         &                              &                            & Camera & Text & Reasoning  & Single-View & Cross-View \\ \midrule
GeoCalib~\citep{veicht2024geocalib}             & Understanding   & \{$f, \xi$\} & \{$\phi, \theta$\} & 37K        & \greencheck & \redcheck & \redcheck & \greencheck & \redcheck \\
CTRL-C~\citep{lee2021ctrl}     & Understanding &  $f$ & \{$\phi, \theta$\}  & 45K       & \greencheck & \redcheck & \redcheck  & \greencheck & \redcheck \\
Deepcalib~\citep{bogdan2018deepcalib}     & Understanding  & \{$f, \xi$\} & - & 67K       & \greencheck & \redcheck & \redcheck  & \greencheck & \redcheck \\
ParamNet~\citep{jin2023perspective}            & Understanding    & $f$ & \{$\phi, \theta$\}  & 190K    & \greencheck & \redcheck & \redcheck & \greencheck & \redcheck \\
Perceptual~\citep{hold2018perceptual}            & Understanding   & $f$ & \{$\phi, \theta$\}  & 390K       & \greencheck & \redcheck & \redcheck & \greencheck   & \redcheck   \\
PreciseCam~\citep{bernal2025precisecam}             & Generation  & \{$f, \xi$\} & \{$\phi, \theta$\}  & 57K       & \greencheck & \greencheck & \redcheck & \greencheck   & \redcheck \\
\textbf{Puffin-4M (Ours)}     & Unified  & $f$ & \{$\phi, \theta, \psi$\} & 4M     & \greencheck & \greencheck & \greencheck & \greencheck & \greencheck \\
\bottomrule[0.17em]
\end{tabular}%
}% end resizebox
\label{tab:dataset_cp}
\captionsetup{justification=raggedright,singlelinecheck=false}
\end{table*}

\noindent\textbf{Perspective Image Generation.} We adopt the pinhole camera model with varying intrinsic parameters (vertical FoV) and extrinsic parameters (roll and pitch) to synthesize perspective images, following the protocol established in recent state-of-the-art camera calibration works~\citep{veicht2024geocalib, jin2023perspective}. For each panoramic image, we generate multiple perspective crops by uniformly sampling roll, pitch, and vertical FoV within the ranges $[-45^\circ, 45^\circ]$, $[-45^\circ, 45^\circ]$, and $[20^\circ, 105^\circ]$, respectively. The number of crops is adaptively determined based on the resolution of the original panorama. While our current setup assumes an ideal pinhole model, incorporating radial distortion effects via an additional distortion parameter $\mathbf{k}$ is left as future work. After generating the perspective images, we further convert the corresponding camera parameters into a pixel-wise Perspective Field~\citep{jin2023perspective} as camera map, where each pixel is annotated with its up-vector $\mathbf{u}_{\mathbf{x}}$ and latitude angle $\varphi_{\mathbf{x}}$ to enable fine-grained spatial encoding:
{%\small
\begin{equation}
    \mathbf{u_{x}}=\lim _{c \rightarrow 0} \frac{\mathcal{P}(\mathbf{X}-c\mathbf{g})-\mathcal{P}(\mathbf{X})}{\|(\mathcal{P}(\mathbf{X})-c\mathbf{g})-\mathcal{P}(\mathbf{X})\|_2}, \, \varphi_{\mathbf{x}}=\arcsin \left(\frac{\mathbf{R} \cdot \mathbf{g}}{\|\mathbf{R}\|_2}\right),
    \label{eq:upVector}
\end{equation}
}where $\mathcal{P}(\mathbf{X}) = \mathbf{x}$ denotes the mapping of a 3D point $\mathbf{X}$ to an image pixel $\mathbf{x}$, and each pixel $\mathbf{x}$ corresponds to a light ray $\mathbf{R}$ originating from $\mathbf{X}$.

\noindent\textbf{Scene and Spatial Reasoning Captions.} Captions are essential for multimodal understanding and generation. To construct high-quality descriptions, we first utilize Qwen2.5-VL-7B-Instruct~\citep{bai2025qwen2} to generate semantically rich captions for each perspective image. These are subsequently distilled using Qwen2.5-7B-Instruct~\citep{qwen2024qwen2} into shorter, vivid, and visually expressive sentences.

To bridge the modality gap between camera geometry and vision-language representations, we introduce the notion of thinking with camera that explicitly guides multimodal tasks using spatially grounded visual cues and professional photographic terms. For captioning such a spatial reasoning process, we propose a two-step strategy: first, we map the numerical camera parameters to their corresponding semantic photographic terms (see Table~\ref{tab:photographic_term}); then, for each camera parameter, we use its corresponding photographic terms to \textit{retrieve} and prompt out relevant visual concepts from the LMM’s prior knowledge. For the trade-off between accuracy and efficacy, we employ Qwen2.5-VL-32B-Instruct to generate the final spatial reasoning captions.

\noindent\textbf{Cross-View Extension.} In addition to the single-view dataset construction described above, we further enrich our dataset with cross-view data involving various tasks. Specifically, during cross-view perspective image generation, we extend the camera’s degrees of freedom by incorporating the yaw angle in addition to roll and pitch, sampling it uniformly from $[0^\circ, 360^\circ)$. The initial view is initialized with a standard camera pose (roll = pitch = yaw = $0^\circ$), and the target view is rendered using a random camera pose sampled within the aforementioned ranges. The target view in each cross-view pair is then captioned using Qwen2.5-VL-7B-Instruct.

For the photographic guidance task, we first consulted photography experts and enthusiasts to identify four key aspects of photographic aesthetics: viewpoint creativity, subject emphasis, compositional balance, and spatial harmonization. These are then formulated into four criteria that serve as aesthetic rating prompts for LMMs. We focus on pitch and yaw as the key controllable camera parameters\footnote{Both professional and amateur users generally prefer near level-shot photography, as humans are highly sensitive to horizontal perturbations~\citep{Howard1966SpatialOrientation, dyde2006subjective}. Thus, we fix roll at $0^\circ$ for this task.}. An initial view is generated with a random pitch in $[-20^\circ, 20^\circ]$, and $N$ neighboring views are sampled by perturbing pitch and yaw within the same range. All rendered views are evaluated by Qwen2.5-VL-32B-Instruct using the aesthetic prompts, and scores are assigned through voting. The pitch and yaw offsets between the initial view and the highest-scoring view are taken as labels for the photographic guidance task.

\noindent\textbf{Prompt Design.} For the scene caption of each image, the prompt is formatted as: ``\texttt{Describe this image in 3-4 sentences}''. Then, we refine the caption into a more compact description using: ``\texttt{Here is a detailed image description: <caption>. Rewrite it into a much shorter, vivid, and visually rich sentence (one or two sentences) that captures only the most essential elements and atmosphere of the scene. Ensure the description is concise, clear, and optimized for use with a text-to-image generation model}''. For captioning the spatial reasoning and photographic aesthetics of each image, we show the corresponding prompts and example results in Figure~\ref{fig:prompt_dataset}.
\section{Experiments}

\subsection{Implementation Details}
\noindent\textbf{Network Configuration.} For the architecture of our Puffin, we use the pretrained C-RADIOv3-H~\citep{heinrich2025radiov2}, Qwen2.5-1.5B-Instruct~\citep{qwen2024qwen2}, and SD3-Medium~\citep{esser2024scaling} to initialize our geometry-aligned vision encoder, LLM, and diffusion model, respectively. Learnable queries with the number of 64 and a lightweight connector comprising six transformer layers are exploited to translate the LLM hidden states to conditioning signals for the diffusion model. The resolutions of the image and camera maps are set to $512\times512$ for all tasks. For tokenization, the camera parameter tokenizer follows the same procedure as the text tokenizer. Since camera parameters are numerical, we first serialize them into discrete tokens, which are naturally handled by the standard text tokenizer without requiring any extra module. Introducing an additional tokenizer (or a separate text encoder) would substantially increase the alignment burden across modules and modalities in a unified multimodal model. For this reason, we keep the vanilla decoder-only LLM backbone and its tokenizer to process both language and camera parameters. For the camera map, we adopt Perspective Field~\citep{jin2023perspective}. We first normalize its values to the range $[-1, 1]$, and then reuse the image tokenizer (\textit{i.e.}, the VAE encoder), as the camera map also has three channels. Pretraining a specialized tokenizer for camera maps is left as future work.

\noindent\textbf{Training Setting.} The whole training process includes four stages as listed in Table~\ref{tab:training_recipe} and takes around 4 days with 64 NVIDIA A100 (80 GB) GPUs. In reference, we use greedy search for text generation in camera understanding and set the CFG weight as 4.5 for the camera-controllable image generation. Prompt template for each task is shown in Section~\ref{sec:appendix_implementation_prompt}.

\begin{table}[t]
\centering
%\footnotesize
\scriptsize
\renewcommand{\arraystretch}{1.0}%
\setlength\tabcolsep{3.9pt}%
\caption{\textbf{Evaluation results on camera understanding.} The comparison methods are evaluated on the public datasets: MegaDepth~\citep{Megadepth}, TartanAir~\citep{Tartanair}, and LaMAR~\citep{sarlin2022lamar}, and our constructed challenging benchmark Puffin-\textit{Und}. We color the \colorbox{tabfirst}{\textbf{best}} and \colorbox{tabsecond}{second best} results.}
\begin{tabular}{clcccccccccccc}
\toprule
&\multirow{2}{*}[-.4em]{Approach} 
& \multicolumn{4}{c}{Roll [degrees]} 
& \multicolumn{4}{c}{Pitch [degrees]} 
& \multicolumn{4}{c}{FoV [degrees]}\\
\cmidrule(lr){3-6}
\cmidrule(lr){7-10}
\cmidrule(lr){11-14}
&& error\,$\downarrow$ & \multicolumn{3}{c}{AUC\,$\triangleright$\,1/5/10\degree\,$\uparrow$}
& error\,$\downarrow$ & \multicolumn{3}{c}{AUC\,$\triangleright$\,1/5/10\degree\,$\uparrow$}
& error\,$\downarrow$ & \multicolumn{3}{c}{AUC\,$\triangleright$\,1/5/10\degree\,$\uparrow$} \\
\midrule
\multirow{9}{*}{\begin{sideways}\textbf{MegaDepth}\end{sideways}}
&{DeepCalib}~\citep{lopez2019deep}      &          \01.41 &          34.6 &          65.4 &          79.4 &          \05.19 &          11.9 &          27.8 &          44.8 &           11.14 &          \05.6 &           12.1 &          22.9 \\
&{Perceptual}~\citep{hold2018perceptual}              &          \01.07 &          47.9 &          72.4 &          83.2 & \cthird \03.49 & \cthird  19.8 & \cthird 39.1 & \cthird  54.2 &           13.40 &          \02.9 &          \08.2 &          16.8 \\
&{CTRL-C}~\citep{lee2021ctrl}                       & \cthird  \00.88 & \cthird  54.5 & \cthird  75.0 & \cthird  84.2 &          \04.80 &          16.6 &          33.2 &          46.5 &           18.65 &          \02.0 &          \05.8 &          12.8 \\
&{MSCC}~\citep{Song2024MSCC}                  &          \00.90 &          53.1 &          72.8 &          82.1 &          \05.73 &          19.0 &          33.2 &          44.3 & \cthird  10.80 &          \06.0 &           14.6 & \cthird  26.2 \\
&{ParamNet}~\citep{jin2023perspective} &          \01.17 &          43.4 &          70.7 &          82.2 &          \03.99 &          15.4 &          34.5 &          53.3 &           11.01 &          \03.2 &           10.1 &          21.3 \\
&{SVA}~\citep{lochman2021minimal}                            &              -  &          31.9 &          35.0 &          36.2 &              -  &          13.6 &          20.6 &          24.9 &              -  & \cthird \09.4 & \cthird   16.1 &          21.1 \\
&{UVP}~\citep{pautrat2023vanishing}    & \cthird \00.51 & \cthird 69.2 & \cthird 81.6 & \cthird 86.9 &          \04.59 & \cthird 21.6 &          36.2 &          47.4 & \cthird   10.92 & \cthird  \08.2 & \cthird  18.7 & \cthird 29.8 \\
&{GeoCalib}~\citep{veicht2024geocalib}                                & \csecond  \00.36 & \csecond  82.6 & \csecond  90.6 & \csecond  94.0 & \csecond  \01.94 & \csecond  32.4 & \csecond  53.3 & \csecond  67.5 & \csecond  \04.46 & \csecond   13.6 & \csecond   31.7 & \csecond  48.2 \\
&\textbf{Puffin (Ours)}                                   & \cfirst  \00.32 & \cfirst  84.9 & \cfirst  93.4 & \cfirst  96.2 & \cfirst  \01.08 & \cfirst  47.6 & \cfirst  68.2 & \cfirst  79.4 & \cfirst  \02.42 & \cfirst  23.9 & \cfirst   47.8 & \cfirst  64.1 \\
\midrule
%%%%%%%%%%%%%
\multirow{9}{*}{\begin{sideways}\textbf{TartanAir}\end{sideways}}
&{DeepCalib}~\citep{lopez2019deep}      &          \01.95 &          24.7 &          55.4 &          71.5 &          \03.27 &          16.3 &          38.8 &          58.5 &          \08.07 &          \01.5 &          \08.8 &          27.2 \\
&{Perceptual}~\citep{hold2018perceptual}              &          \02.24 &          23.2 &          48.6 &          66.7 &          \02.86 &          23.5 &          44.6 &          61.5 &           15.06 &          \05.1 &          \08.9 &          17.1 \\
&{CTRL-C}~\citep{lee2021ctrl}                       &          \01.68 &          32.8 &          59.1 & \cthird 74.1 & \cthird \02.39 & \cthird  24.6 & \cthird  48.6 & \cthird 65.2 & \csecond \05.64 & \cthird   10.7 & \cthird   25.4 & \csecond 43.5 \\
&{MSCC}~\citep{Song2024MSCC}                  &          \03.50 &          15.0 &          37.2 &          57.7 &          \03.48 &          18.8 &          38.6 &          54.3 &           11.18 &          \04.4 &           11.8 &          23.0 \\
&{ParamNet}~\citep{jin2023perspective} & \cthird  \01.63 & \cthird  34.5 & \cthird  59.2 & \cthird  73.9 &          \03.05 &          19.4 &          42.0 &          60.3 &          \08.21 &          \06.0 &           16.8 &          31.6 \\
&{SVA}~\citep{lochman2021minimal}                            &          \09.48 &          32.4 &          39.6 &          44.1 &           18.46 &          21.2 &          28.8 &          34.5 &           43.01 &          \08.8 &           16.1 &          21.6 \\
&{UVP}~\citep{pautrat2023vanishing}    & \cthird \00.89 & \cthird 52.1 & \cthird 64.8 &          71.9 & \cthird  \02.48 & \cthird 36.2 & \cthird 48.8 &          58.6 &          \09.15 & \cthird   15.8 & \cthird  25.8 &          35.7 \\
&{GeoCalib}~\citep{veicht2024geocalib}                                   & \csecond  \00.43 & \csecond  71.3 & \csecond  83.8 & \csecond  89.8 & \csecond  \01.49 & \csecond  38.2 & \csecond  62.9 & \csecond  76.6 & \cfirst  \04.90 & \cthird  14.1 & \cfirst   30.4 & \cfirst  47.6 \\
&\textbf{Puffin (Ours)}                                & \cfirst  \00.40 & \cfirst  71.7 & \cfirst  86.2 & \cfirst  92.1 & \cfirst  \00.95 & \cfirst  51.0 & \cfirst  68.2 & \cfirst  79.3 & \cthird  \07.48 & \cfirst  16.3 & \csecond   28.5 & \cthird  39.0 \\
\midrule
%%%%%%%%%%%%%
\multirow{9}{*}{\begin{sideways}\textbf{LaMAR}\end{sideways}}
&{DeepCalib}~\citep{lopez2019deep}      &          \01.15 &           44.1 &           73.9 &           84.8 &          \04.68 &           10.8 &           28.3 &           49.8 &           10.93 &          \00.7 &           13.0 &           24.0 \\
&{Perceptual}~\citep{hold2018perceptual}              &          \01.29 &           40.0 &           68.9 &           81.6 &          \02.83 &           21.2 &           44.7 &           62.6 &           17.78 &          \03.0 &          \05.3 &           10.7 \\
&{CTRL-C}~\citep{lee2021ctrl}                       &          \01.20 &           43.5 &           70.9 &           82.5 & \cthird  \01.94 & \cthird   27.6 & \cthird   54.7 & \cthird  70.2 & \cthird  \05.64 & \cthird  \09.8 & \cthird   24.6 & \cthird   43.2 \\
&{MSCC}~\citep{Song2024MSCC}                  &          \01.44 &           39.6 &           60.7 &           72.8 &          \03.02 &           20.9 &           41.8 &           55.7 &           14.78 &          \03.2 &          \08.3 &           16.8 \\
&{ParamNet}~\citep{jin2023perspective} & \cthird  \00.93 & \cthird   51.7 & \cthird   77.0 & \cthird  86.0 &          \02.15 &           27.0 &           52.7 & \cthird  70.2 &           14.71 &          \02.8 &          \06.8 &           14.3 \\
&{SVA}~\citep{lochman2021minimal}                            &              -  &          \08.6 &          \09.2 &          \09.7 &              -  &          \03.4 &          \05.7 &          \07.0 &              -  &          \01.2 &          \02.7 &          \04.1 \\
&{UVP}~\citep{pautrat2023vanishing}    & \csecond \00.38 & \cthird  72.7 & \cthird  81.8 & \cthird   85.7 & \cthird \01.34 & \cthird  42.3 & \cthird  59.9 & \cthird   69.4 & \cthird \05.57 & \cthird  15.6 & \cthird  30.6 & \cthird  43.5 \\
&{GeoCalib}~\citep{veicht2024geocalib}                                 & \cfirst  \00.28 & \cfirst   86.4 & \cfirst   92.5 & \cfirst   95.0 & \csecond  \00.87 & \csecond   55.0 & \csecond   76.9 & \csecond   86.2 & \cfirst  \03.03 & \cfirst   19.1 & \cfirst   41.5 & \cfirst   60.0 \\
&\textbf{Puffin (Ours)}
& \csecond  \00.38 & \csecond  80.6 & \csecond  89.8 & \csecond  93.5 & \cfirst  \00.71 & \cfirst  61.7 & \cfirst  78.9 & \cfirst  86.4 & \csecond  \03.62 & \csecond  17.0 & \csecond   37.3 & \csecond  53.1 \\
\midrule
%%%%%%%%%%%%%
\multirow{7}{*}{\begin{sideways}\textbf{Puffin-\textit{Und}}\end{sideways}}
&{DeepCalib}~\citep{lopez2019deep}      &          \01.90 &           29.3 &           56.2 &           71.7 &          \03.71 &           15.3 &           36.0 &           54.9 &           \07.43 &          \09.0 &           19.4 &           34.8 \\
&{CTRL-C}~\citep{lee2021ctrl}                       &          \04.69 &           20.3 &           35.2 &           46.7 & \cthird  \08.43 & \cthird   10.8 & \cthird   24.6 & \cthird  36.1 & \cthird  11.70 & \cthird  \05.3 & \cthird   12.7 & \cthird   23.5 \\
&{MSCC}~\citep{Song2024MSCC}                  &          \04.40 &           17.4 &           34.7 &           47.9 &          \06.87 &           13.1 &           26.3 &           38.9 &           \09.79 &          \06.8 &          16.3 &           29.0 \\
&{ParamNet}~\citep{jin2023perspective} & \cthird  \02.11 & \cthird   24.9 & \cthird   53.6 & \cthird  71.5 &          \03.40 &          16.1 &           38.7 & \cthird  58.6 &           \06.21 &          \09.4 &          22.3 &           39.8 \\
&{UVP}~\citep{pautrat2023vanishing}    & \cthird \02.03 & \cthird  32.7 & \cthird  46.4 & \cthird   54.9 & \cthird \09.04 & \cthird  11.4 & \cthird  22.6 & \cthird   32.5 & \cthird 18.80 & \cthird  \05.0 & \cthird  12.1 & \cthird  19.9 \\
&{GeoCalib}~\citep{veicht2024geocalib}                                 & \csecond  \00.92 & \csecond   53.6 & \csecond   73.9 & \csecond   82.6 & \csecond  \02.18 & \csecond   28.9 & \csecond   52.5 & \csecond   69.6 & \csecond  \05.04 & \csecond   12.4 & \csecond   28.0 & \csecond   45.8 \\
&\textbf{Puffin (Ours)}                                   & \cfirst  \00.41 & \cfirst  78.3 & \cfirst  91.0 & \cfirst  95.2 & \cfirst  \00.74 & \cfirst  60.2 & \cfirst  81.2 & \cfirst  90.0 & \cfirst  \01.21 & \cfirst  42.4 & \cfirst   70.5 & \cfirst  84.3 \\
%%%%%%%%%%%%%
\bottomrule
\end{tabular}
\label{tab:cam_und_evaluation}
\end{table}
\subsection{Evaluations on Camera Understanding} 
\label{sec:eva_und}
\noindent\textbf{Settings.} Following prior works, we compare our method against a range of learning-based camera calibration approaches, including DeepCalib~\citep{lopez2019deep}, Perceptual~\citep{hold2018perceptual}, CTRL-C~\citep{lee2021ctrl}, MSCC~\citep{Song2024MSCC}, ParamNet~\citep{jin2023perspective}, and GeoCalib~\citep{veicht2024geocalib}, as well as traditional methods such as SVA~\citep{lochman2021minimal} and UVP~\citep{pautrat2023vanishing}. For each image, gravity estimation is evaluated using the angular errors in roll and pitch, while focal length is evaluated through the error in vertical FoV. For all metrics, we report both the median error and the Area Under the Recall Curve (AUC) at thresholds of $1^\circ$, $5^\circ$, and $10^\circ$. We conduct evaluations on three common datasets, MegaDepth~\citep{Megadepth}, TartanAir~\citep{Tartanair}, and LaMAR~\citep{sarlin2022lamar}. Notably, images from these datasets are primarily captured or simulated in well-structured environments, where buildings, rooms, or trees occupy a substantial portion of the scene. Moreover, the camera parameters in some datasets are limited in distribution; for instance, TartanAir uses a single FoV for all images. To complement these settings, we construct a more challenging dataset, Puffin-\textit{Und}, designed for a comprehensive assessment of camera understanding. This dataset contains 1,000 images spanning diverse camera configurations and scenarios.

\noindent\textbf{Comparison Results.} As shown in Table~\ref{tab:cam_und_evaluation}, our method outperforms the baselines on MegaDepth and Puffin-\textit{Und}, and achieves comparable results on TartanAir and LaMAR. Due to the fixed image resolution in our training data ($512\times512$), we adopt a central cropping strategy followed by resizing to rectangular inputs for evaluating non-square images. The vertical FoV is then computed from the predicted value and scaled according to the crop ratio. Nevertheless, this procedure may discard semantically valid content and thereby degrade our camera understanding performance, particularly when the aspect ratio deviates substantially from unity, as in LaMAR, where Puffin slightly underperforms the state-of-the-art method~\citep{veicht2024geocalib}. While this limitation is orthogonal to our current exploration, it could be potentially mitigated in future work by constructing a multi-scale training dataset. 

We present the horizon lines derived from the predicted camera parameters of different methods in Figure~\ref{fig:horizon}. Compared to prior approaches, our Puffin demonstrates strong performance not only in common scenarios such as architectural and indoor scenes, but also in challenging cases characterized by limited geometric features or significantly tilted camera poses. These results highlight the robustness of the proposed method. 

\noindent\textbf{Discussion.}
From the quantitative evaluation of existing methods, we observe that estimating pitch and FoV is considerably more challenging than estimating roll. This difference arises from the nature of the underlying visual cues. Roll estimation is supported by low-/mid-level geometric representations, such as edges and vanishing lines, which are directly embedded in the image structure and thus relatively straightforward to learn. In contrast, accurate estimation of pitch and FoV requires more extensive contextual priors. Unlike previous vision-based approaches, our method explicitly models the relationship between physical camera parameters and spatial context using a large multimodal model. This integration allows the model to capture spatially contextual knowledge that cannot be sufficiently represented through visual features alone. We visualize additional camera understanding results (with camera maps converted from the predicted camera parameters) on diverse inputs, including AIGC images~\citep{GTP-4o} and real-world photographs, in Figure~\ref{fig:cam_und_map}.

\begin{figure}[t]
    \centering
    \begin{subfigure}{1\textwidth}
        \centering
        \includegraphics[width=1\linewidth]{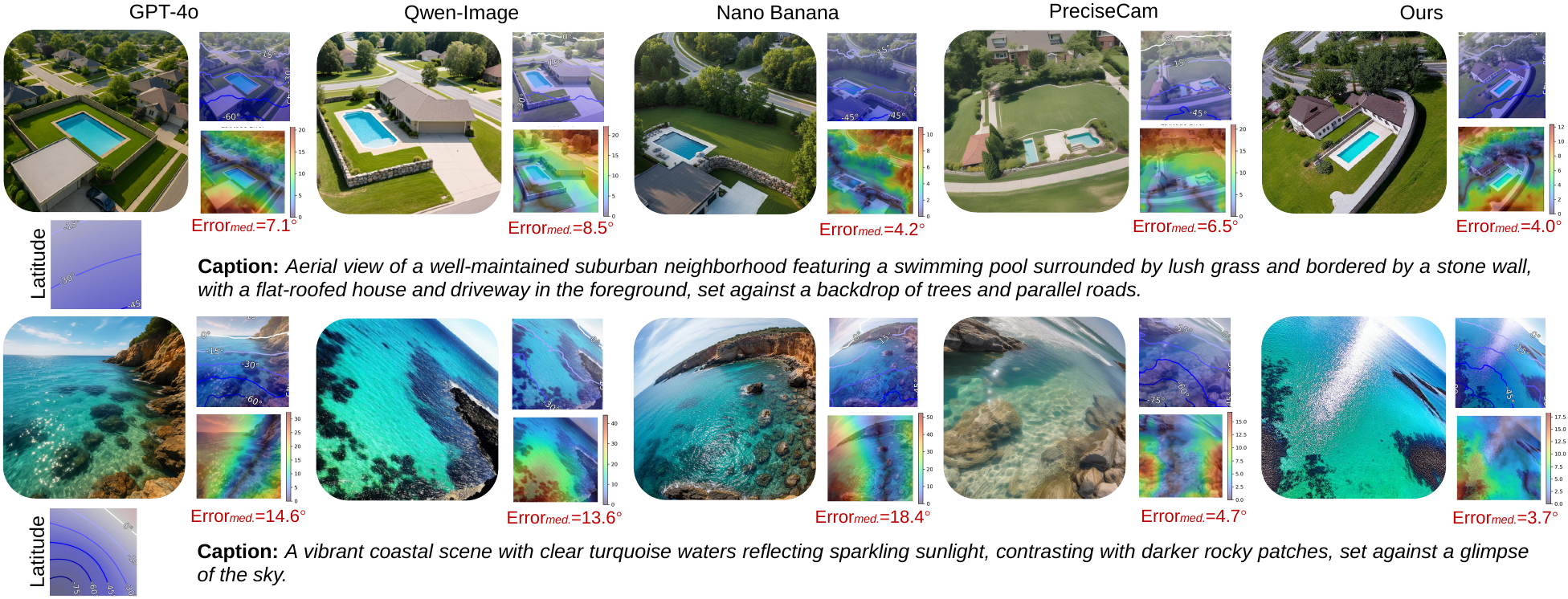}
    \end{subfigure}
    
   % \vspace{-0.1em}
    
    \begin{subfigure}{1\textwidth}
        \centering
        \includegraphics[width=1\linewidth]{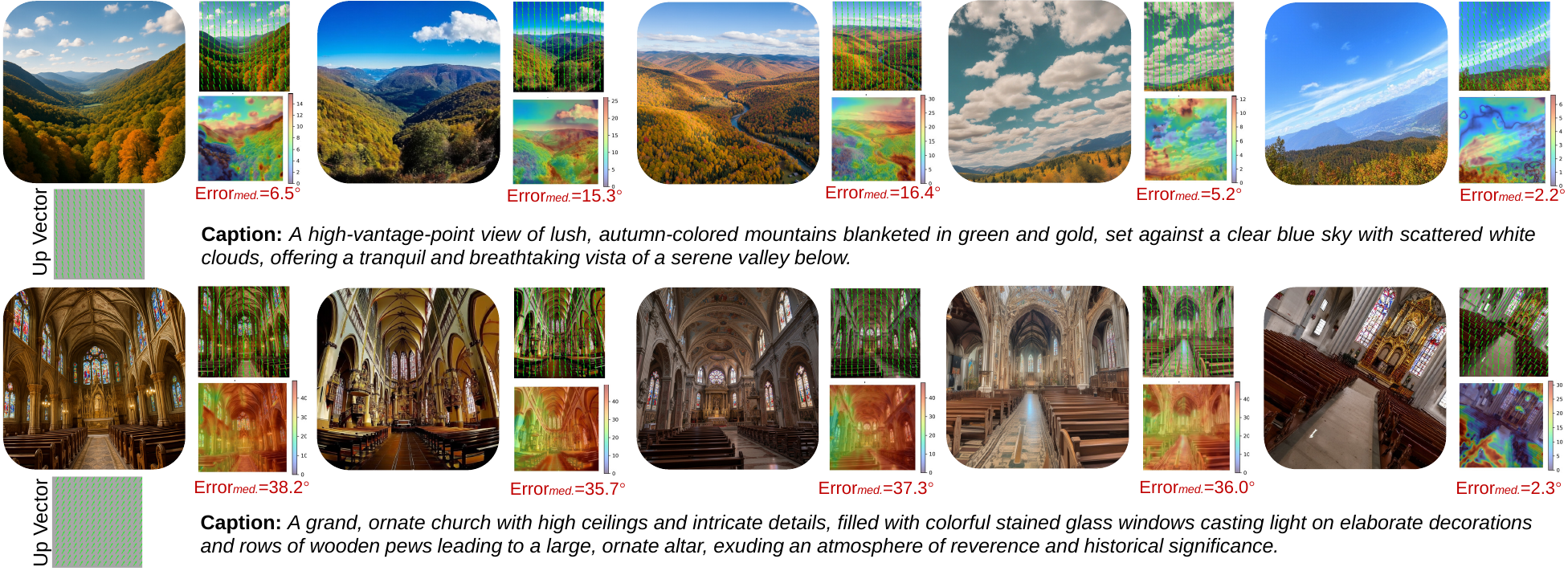}
    \end{subfigure}

    \caption{\textbf{Comparison results on controllable generation.} We visualize the generated image along with its camera map (latitude or up vector, estimated by ~\citep{veicht2024geocalib}), error map to the GT camera map, and the median error. The caption and target camera map are presented at the bottom of each comparison.}
    \label{fig:gen_vis}
\end{figure}

\begin{table}[t]
\centering
%\footnotesize
\scriptsize
\renewcommand{\arraystretch}{1.0}%
\setlength\tabcolsep{0.3pt}%
\caption{\textbf{Camera-controllable generation evaluation on Puffin-\textit{Gen}.} When evaluating multimodal models, we convert the camera parameters from radians to degrees* or express them using standard photographic terms$^\dagger$.}
\begin{tabular}{clcccccccc}
\toprule
& \multirow{2}{*}[-.4em]{Approach}
& \multicolumn{2}{c}{Up Vector [degrees]}
& \multicolumn{2}{c}{Latitude [degrees]}
& \multicolumn{2}{c}{Gravity [degrees]}
& \multicolumn{1}{c}{{Visual Quality}}\\
\cmidrule(lr){3-4}\cmidrule(lr){5-6}\cmidrule(lr){7-8}\cmidrule(lr){9-10}
&& mean error $\downarrow$ & median error $\downarrow$
& mean error $\downarrow$ & median error $\downarrow$
& mean error $\downarrow$ & median error $\downarrow$
& {FID $\downarrow$} & \\
\midrule
\multirow{1}{*}{\begin{sideways}\textbf{}\end{sideways}}
& GPT-4o*~\citep{GTP-4o}
& 24.11 & 22.86 & 15.87 & 13.67 &  28.08 &  27.39 &  95.92 &  \\
& GPT-4o$^\dagger$~\citep{GTP-4o}
& 24.07 & 22.10 & 14.67 & 12.43 &  27.19 &  26.32 &  94.43 &  \\
& Qwen-Image*~\citep{wu2025qwen}
&  23.80 &  22.73 &  15.76 &  13.90 &  27.75 &  27.22 &  \csecond 83.31 &  \\
& Qwen-Image$^\dagger$~\citep{wu2025qwen}
&  23.98 &  22.60 &  15.92 &  13.92 &  27.86 &  26.45 &  83.37 &  \\
& Nano Banana*~\citep{gemini25flashimage}
&  24.08 &  23.13 &  16.66 &  15.05 &  28.78 &  28.22 &  91.66 &  \\
& Nano Banana$^\dagger$~\citep{gemini25flashimage}
&  24.65 &  23.50 &  15.80 &  13.98 &  28.22 &  26.73 &  88.02 &  \\
& PreciseCam~\citep{bernal2025precisecam}
& \csecond 18.66 & \csecond 17.47 & \csecond 12.49 & \csecond \09.99 & \csecond 18.39 & \csecond 15.34 &  90.91 &  \\
& \textbf{Puffin (Ours)}
& \cfirst 11.94 & \cfirst 10.12 & \cfirst \06.34 & \cfirst \04.04 & \cfirst \06.79 & \cfirst \03.43 &  \cfirst 69.46 & \\
\bottomrule
\end{tabular}
\label{tab:cam_gen_evaluation_fid}
\end{table}

\subsection{Evaluations on Camera-Controllable Generation}
\noindent\textbf{Settings.}  We evaluate our generation performance against the state-of-the-art method PreciseCam~\citep{bernal2025precisecam}. In addition, we compare our approach with recent powerful multimodal models, including GPT-4o~\citep{GTP-4o}, Qwen-Image~\citep{wu2025qwen}, and Nano Banana~\citep{gemini25flashimage}, using the same captions as our method. The prompt templates are shown in Section~\ref{sec:appendix_implementation_prompt}. To mitigate the data gap for the multimodal models, we convert the camera parameters in captions from radians to degrees or express them using professional photographic terms. For quantitative evaluation, we adopt an offline method~\citep{veicht2024geocalib} to estimate pixel-wise camera maps. Using the ground truth maps, we then compute the mean and median errors of the up vector, latitude, and gravity, all measured in degrees. We also incorporate the standard FID metric to assess the overall visual quality of the generated images. Since no benchmark dataset exists for text-to-image generation with precise camera parameters, we construct Puffin-\textit{Gen} to fill this gap. The dataset consists of 650 caption–camera pairs spanning diverse scenarios and camera configurations. We will release Puffin-\textit{Gen} and Puffin-\textit{Und} to support standardized evaluation and facilitate subsequent works.

\begin{figure}
    \centering
    \includegraphics[width=1\linewidth]{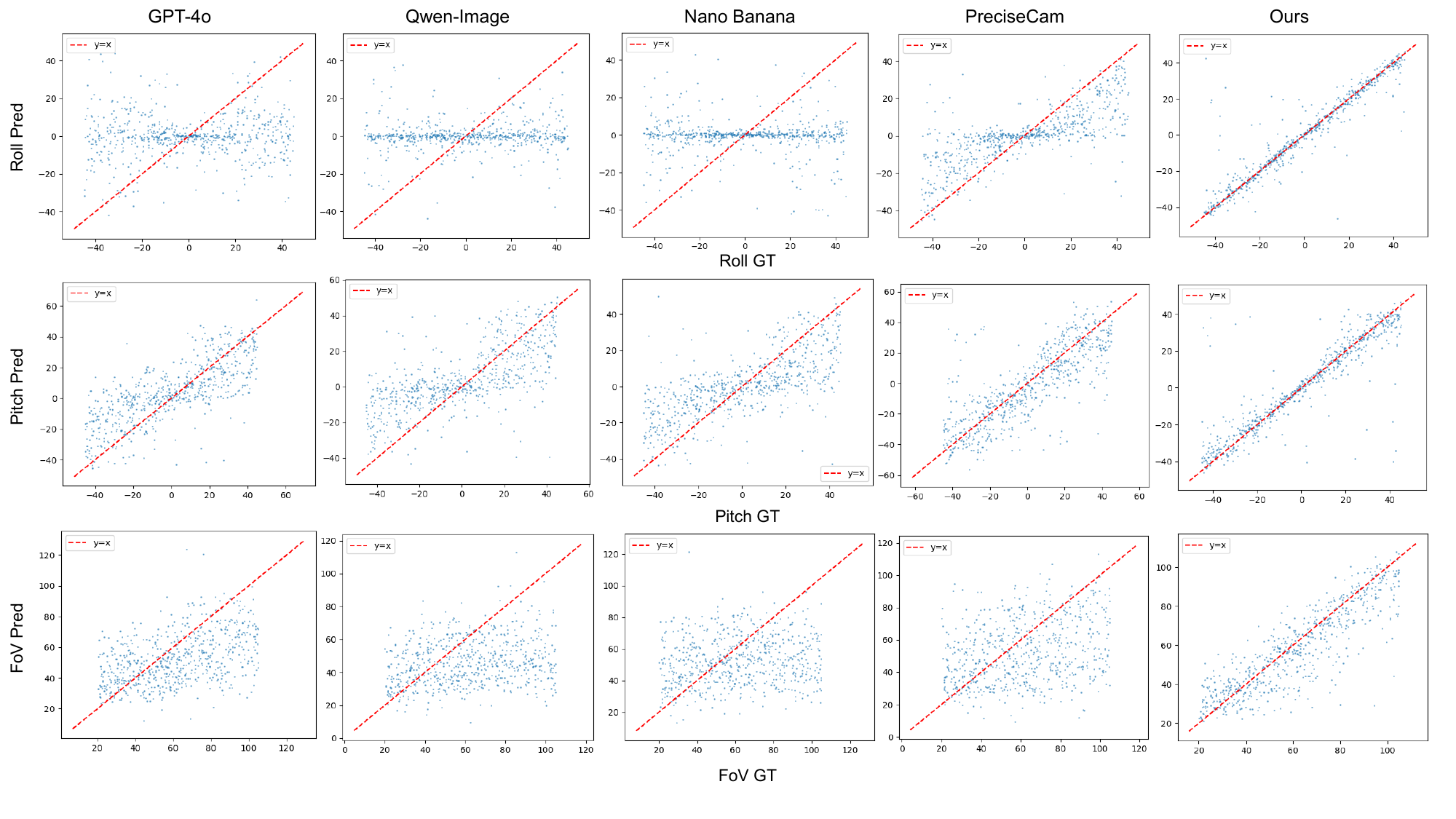}
    \caption{\textbf{The predicted \textit{vs.} ground truth camera parameters across all generated samples.} Compared with previous methods, our generated results well align with the distribution of the ground truth camera parameters.
    }
    \label{fig:plot_gen_comparison}
    %\vspace{-0.4cm}
\end{figure}

\noindent\textbf{Comparison Results.} 
We report quantitative and qualitative results in Table~\ref{tab:cam_gen_evaluation_fid} and Figure~\ref{fig:gen_vis}. Our method outperforms existing multimodal models by a large margin across all metrics. While these models produce high-quality and aesthetically pleasing images, they fail to ensure spatially consistent layouts under specific camera configurations. PreciseCam~\citep{bernal2025precisecam} provides effective control but often generates monotonous stylized outputs (\textit{e.g.}, anime) with limited diversity, and struggles with challenging configurations such as significantly tilted poses. In contrast, our method generalizes well across diverse scenarios and camera settings, demonstrating strong practicality for real-world image generation. Additional generated results are shown in Figure~\ref{fig:gallery} and parameter-specific controls in Figure~\ref{fig:camctrl_t2i}.

\noindent\textbf{Discussion.}
We further conduct an in-depth analysis to understand why existing image generation models fail to achieve accurate spatial simulation. Specifically, we decouple the spatial distributions of the generated images with respect to three camera parameters: roll, pitch, and FoV. As illustrated in Figure~\ref{fig:plot_gen_comparison}, we visualize scatter plots of the predicted \textit{vs.} ground truth camera parameters (with the reference line $y=x$) across all generated samples. For fairness, the predicted camera parameters are obtained using the state-of-the-art vision-based camera calibration method~\citep{veicht2024geocalib}.

Interestingly, we observe a reversed role of camera parameters in controllable generation compared with camera understanding. Specifically, images generated by previous methods~\citep{GTP-4o, wu2025qwen, gemini25flashimage, bernal2025precisecam} exhibit poor simulation accuracy on roll compared to pitch, where the predicted roll values fail to align with the target ground truth. In contrast, roll estimation in camera understanding is generally easier than pitch, due to its explicit link with geometric structures.

We attribute this discrepancy to two main factors: (i) Most existing image generation models are trained on datasets curated for high visual aesthetics. Both professional and amateur photographers tend to prefer near-level shots, as humans are sensitive to horizontal perturbations~\citep{he2013content, Howard1966SpatialOrientation, dyde2006subjective}. Consequently, variations in roll often conflict with aesthetic preferences, leading to a skewed dataset distribution with far fewer roll variants compared to pitch or FoV. (ii) Roll directly alters the perceived gravity direction in an image, thereby reformulating the common sense of spatial layout. For instance, a strong Dutch angle can make the sea surface appear above the horizon line, creating an inverted spatial illusion. Such cases are inherently more difficult to simulate, whereas pitch and FoV changes typically only affect the viewing scope without fundamentally disrupting the physical law.

\begin{table}[t]
\centering
%\footnotesize
\scriptsize
\renewcommand{\arraystretch}{1.0}%
\setlength\tabcolsep{4.1pt}%
\caption{\textbf{Ablation study on camera understanding.} We evaluate our method with different architectures and the mode of thinking with camera.}
\begin{tabular}{clcccccccccccc}
\toprule
&\multirow{2}{*}[-.4em]{Approach} 
& \multicolumn{4}{c}{Roll [degrees]} 
& \multicolumn{4}{c}{Pitch [degrees]} 
& \multicolumn{4}{c}{FoV [degrees]}\\
\cmidrule(lr){3-6}
\cmidrule(lr){7-10}
\cmidrule(lr){11-14}
&& error\,$\downarrow$ & \multicolumn{3}{c}{AUC\,$\triangleright$\,1/5/10\degree\,$\uparrow$}
& error\,$\downarrow$ & \multicolumn{3}{c}{AUC\,$\triangleright$\,1/5/10\degree\,$\uparrow$}
& error\,$\downarrow$ & \multicolumn{3}{c}{AUC\,$\triangleright$\,1/5/10\degree\,$\uparrow$} \\
\midrule
\multirow{9}{*}{\begin{sideways}\textbf{}\end{sideways}}
&{InternVL3}~\citep{zhu2025internvl3}     &          \00.91 &          53.7 &          75.5 & \cthird  85.6 &          \01.72 &          31.9 &          59.7 &          76.3 &          \02.96 &          19.7 &           43.1 &          63.5\\
&{Qwen2.5-VL}~\citep{bai2025qwen2}    &          \00.79 &          58.8 &          78.0 & \cthird  86.5 &          \01.61 &          36.4 &          62.4 &          78.0 &          \02.91 &          19.4 &           42.5 &          62.5\\
&{Vision Encoder}~\citep{heinrich2025radiov2}      &          \00.55 &          69.0 &          86.2 & \cthird  92.6 &          \01.00 &          49.8 &          74.1 &          85.9 &          \01.87 &          28.3 &           57.9 &          75.9 \\
&{Ours}             &          \00.47 &          75.6 &          89.7 &          94.6 &          \00.91 &          54.2 &          77.5 &          87.9 &           \01.48 &          38.0 &          66.2 &          81.5 \\
&{Ours \textit{w/} Thinking}        & \cfirst  \00.41 & \cfirst  78.3 & \cfirst  91.0 & \cfirst  95.2 & \cfirst  \00.74 & \cfirst  60.2 & \cfirst  81.2 & \cfirst  90.0 & \cfirst  \01.21 & \cfirst  42.4 & \cfirst   70.5 & \cfirst  84.3 \\
\bottomrule
\end{tabular}
\label{tab:cam_und_ablation}
\end{table}

\subsection{Ablation Studies}
\label{sec:ablation_study}

\noindent\textbf{Architecture.}
As discussed in Section~\ref{sec:thinking_und}, directly fine-tuning the existing VLMs yields a significant performance bottleneck since their vision encoders learn overly condensed high-level features and language models have little prior knowledge of spatial perception. As listed in Table~\ref{tab:cam_und_ablation}, directly finetuning the current VLMs~\citep{bai2025qwen2, zhu2025internvl3} even underperforms the vision-only network. To this end, we carefully pair an LLM~\citep{qwen2024qwen2} with the vision encoder~\citep{heinrich2025radiov2}; both of them are first aligned and then fine-tuned on the camera understanding dataset. By jointly integrating the geometric perception and context understanding in a staged optimization manner, our method (Ours) outperforms the above approaches on all evaluation metrics.

\begin{figure}[t]
    \centering
    \includegraphics[width=1\linewidth]{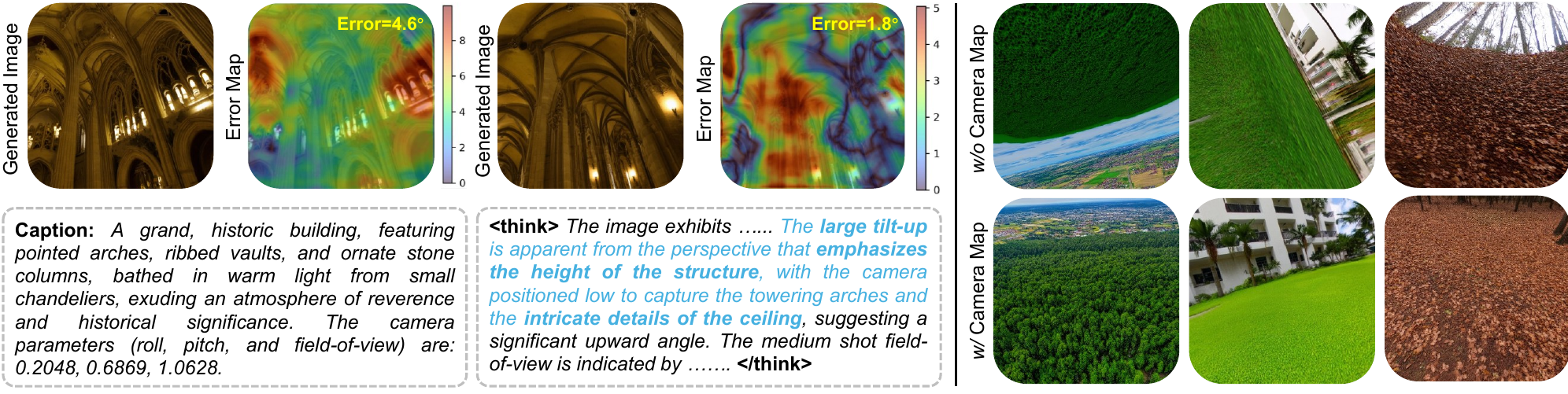}
    \caption{\textbf{Ablation study on the camera-controllable generation.} We evaluate the effectiveness of the thinking mode (left) and the precise geometric encoding provided by camera map (right).}
    \label{fig:reasoning_ablation_gen}
    %\vspace{-0.4cm}
\end{figure}

\noindent\textbf{Thinking with Camera.}
To mitigate the modality gap between camera and vision–language, we introduce thinking with camera. For camera understanding, we align spatially grounded visual cues with photographic terms across geometric context, enabling LMMs to predict camera parameters through structured spatial reasoning. As shown in Table~\ref{tab:cam_und_ablation}, this design (Ours \textit{w/} Thinking) consistently improves performance, especially for pitch and FoV prediction that depend on broader contextual priors, demonstrating the framework’s ability to capture hierarchical spatial context beyond localized geometric cues. Thinking with camera also enhances camera-controllable generation: given a prompt with scene descriptions and target parameters, the model infers potential spatial cues and uses them as a semantic planning stage to guide synthesis. As illustrated in Figure~\ref{fig:reasoning_ablation_gen} (left), it emphasizes visual cues such as ceilings under a large tilt-up, yielding more accurate spatial simulation.

\noindent\textbf{Camera Parameters \textit{vs.} Camera Map.}
Beyond discrete camera tokens derived from explicit numerical parameters, we further introduce a continuous representation of camera geometry via pixel-wise camera maps for controllable image generation. We show the effectiveness of the precise geometric encoding provided by camera map in Figure~\ref{fig:reasoning_ablation_gen} (right). Compared to numerical values of the camera parameters, the camera map encodes the local geometric context at each pixel, including orientation and spatial displacement clues, offering precise control over spatial layout and viewpoint. Without the camera map as conditions, generated images may exhibit severe geometric distortions and inverted spatial illusions under challenging camera configurations.

\begin{wrapfigure}{r}{0.48\textwidth}
	%\vspace{-9pt}
	\centering
	\includegraphics[width=\linewidth]{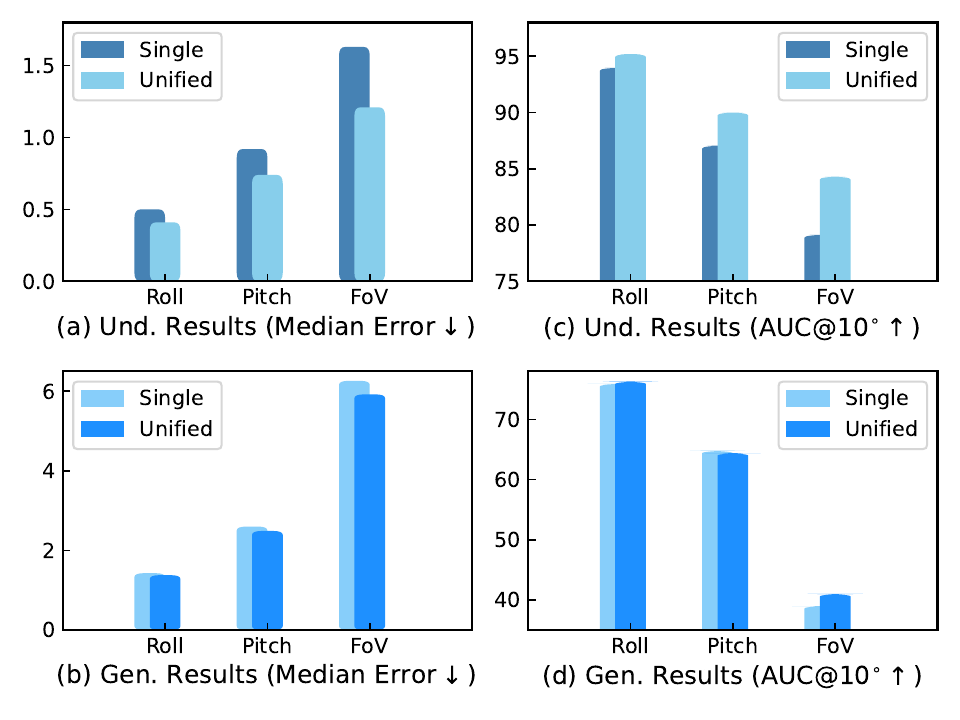}
	%\vspace{-16pt}
	\caption{\small{\textbf{Ablation study on the mutual effect} between camera understanding (a)(c) and camera-controllable generation (b)(d) supervision.
	}}\label{fig:mutual_effect}
	%\vspace{-15pt}
\end{wrapfigure}
\noindent\textbf{Single Task \textit{vs.} Unified Training.}
In addition to performing multimodal tasks within a unified framework, we aim to exploit the mutual benefits between understanding and generation through joint training. Unlike previous works~\citep{pan2025transfer, wu2025openuni}, we jointly optimize both the LLM and the diffusion model across understanding and generation tasks. This strategy avoids the representational bottleneck imposed by frozen modules and fosters a bidirectional synergy between the two tasks. As illustrated in Figure~\ref{fig:mutual_effect}(a)(c), training the camera understanding component in isolation underperforms compared to the unified framework, as the generation process contributes auxiliary diffusion loss at the low-level appearance, which implicitly enhances detailed geometric perception. While the performance gain for generation is less pronounced than for understanding in Figure~\ref{fig:mutual_effect}(b)(d), notable improvements emerge in challenging scenarios such as FoV simulation, which requires prior knowledge regarding precise and holistic spatial understanding within an image.

For other general-purpose unified models, the understanding tasks mainly target high-level concepts such as recognition and semantic comprehension. As a result, the domain gaps across multimodal tasks are more pronounced in these models, likely requiring more delicate architectural designs to harmonize representations across different modalities.

\subsection{Applications}

\begin{figure}
    \centering
    \includegraphics[width=1\linewidth]{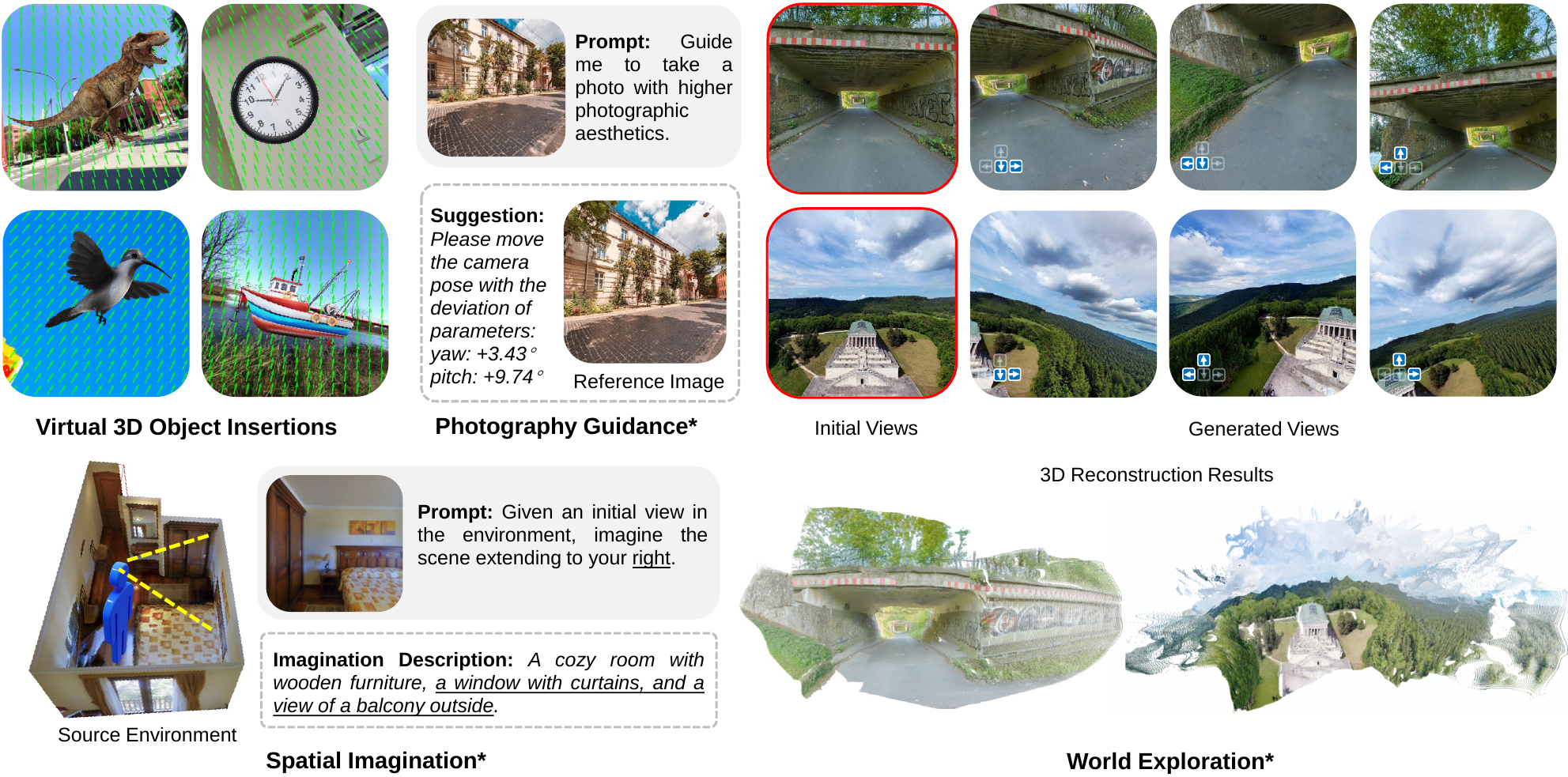}
    \caption{\textbf{Applications of the proposed Puffin.} Our method can help 3D object insertion into a wild image by predicting its camera parameters. Additionally, it can flexibly extend to various cross-view tasks such as the spatial imagination, world exploration, and photographic guidance, by instruction tuning*.
    %* denotes our instruction tuning tasks.
    }
    \label{fig:application}
    %\vspace{-0.4cm}
\end{figure}

We illustrate the versatile capabilities of our Puffin in Figure~\ref{fig:application}. Similar to previous methods~\citep{hold2018perceptual, jin2023perspective}, Puffin can support virtual 3D object insertion into in-the-wild images by accurately predicting camera parameters. Furthermore, it can be flexibly extended to a range of cross-view tasks through instruction tuning, such as spatial imagination, world exploration, and photographic guidance. For both the initial and generated views in world exploration, we visualize 3D reconstruction results with VGGT~\citep{wang2025vggt}, showing proper spatial consistency across viewpoints. Additional results are presented in Figure~\ref{fig:cross_view_gen_iter},~\ref{fig:cross_view_gen_vggt},~\ref{fig:photo_spatial_imagination}.

\subsection{Limitation and Future Work}

Because our training dataset is constructed at a fixed resolution of $512\times512$, Puffin’s image generation is currently restricted to a single scale. For camera understanding, we applied central cropping followed by resizing (Section~\ref{sec:eva_und}), an operation that may discard semantically valid content and degrade performance, particularly when the aspect ratio deviates significantly from unity. While these limitations are orthogonal to our main focus, they could be addressed in future work by building multi-scale training datasets. Beyond data design, our evaluation of camera-controllable generation relied on an offline vision-based calibration method~\citep{veicht2024geocalib}. Although this approach reflects the best available practice, the calibration errors it reports can be ambiguous, especially for generated images exhibiting only subtle spatial differences. Accurately evaluating spatial simulation thus remains an open challenge and is crucial for advancing camera-controllable generation. We plan to address this by incorporating stronger camera understanding models as evaluators and by designing benchmarks that more precisely capture geometric consistency. In addition, we aim to further enhance Puffin’s cross-view capability and extend it to camera-centric video generation and understanding, paving the way for broader applications in dynamic and immersive environments.

\section{Conclusion}
We introduce Puffin, a unified multimodal model that jointly performs camera-centric understanding and generation across arbitrary viewpoints. These two tasks have been commonly treated as isolated problems and independently explored by the research community. Yet, in essence, they represent two complementary sides: decoding the geometry of the world and encoding it back into controllable, perceptually consistent visual content. Unlike previous unified models restricted to oversimplified front-view assumptions, Puffin eliminates the modality gap by interpreting the camera as language and leverages the notion of thinking with camera. We argue that unifying camera-centric understanding and generation anchors perception and synthesis to a shared representation of camera geometry, allowing machines to reason about space more holistically and interactively. Such a unified camera-centric model underpins robust spatial intelligence and fosters more versatile applications.
\section*{Acknowledgment}
This study is supported under the RIE2020 Industry Alignment Fund Industry Collaboration Projects (IAF-ICP) Funding Initiative, as well as cash and in-kind contribution from the industry partner(s). It is also supported by Singapore MOE AcRF Tier 2 (MOE-T2EP20224-0003). We thank Zhouxia Wang, Zongsheng Yue, Haiwen Diao, Zhaoxi Chen, Zhijie Shen, Xiang Li, Qiqi Gong, Edurne Bernal-Berdun, and Koki Maeda for their insightful discussions. We thank Xu Song and Juncheng Zhou for their help with the comparison methods.

%\newpage
\normalem
\bibliography{iclr2026_conference}
\bibliographystyle{iclr2026_conference}

\newpage
\appendix
\renewcommand{\thesection}{A\arabic{section}}
\renewcommand{\thefigure}{A\arabic{figure}}
\renewcommand{\thetable}{A\arabic{table}}
\setcounter{section}{0}
\setcounter{figure}{0}
\setcounter{table}{0}

\section*{Appendix}

\newcolumntype{L}{>{\raggedright\arraybackslash}X}
\newcolumntype{C}{>{\centering\arraybackslash}X}
\renewcommand\cellalign{tc}
\renewcommand\theadalign{tc}
\renewcommand{\arraystretch}{1.0}
\setlength{\tabcolsep}{5.8pt}

%\svgpath{{figures/icon/}}
%\newcommand{\inlinesvg}[2][1em]{%
%  \raisebox{-0.2ex}{\includesvg[height=#1]{#2}}%
%}

\begin{table}[th!]
\centering
%\footnotesize
\scriptsize
\caption{\textbf{Camera parameter-to-term mapping.}
To align camera parameters (roll, pitch, and FoV) with the prior knowledge space of LMMs, their numerical ranges are mapped to professional photographic terms.}
\begin{tabular}{@{} l p{1.5cm} *{5}{c} @{}}
\toprule
&  & \multicolumn{5}{c}{\textbf{Roll}} \\
\cmidrule(lr){3-7}
& \textbf{Term ($t$)}
  & \makecell{Large counterclockwise\\Dutch angle}
  & \makecell{Small counterclockwise\\Dutch angle}
  & Near level shot
  & \makecell{Small clockwise\\Dutch angle}
  & \makecell{Large clockwise\\Dutch angle} \\
& \textbf{Example}
  & \multicolumn{1}{c}{\includegraphics[height=1.85em]{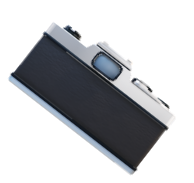}} & \multicolumn{1}{c}{\includegraphics[height=1.55em]{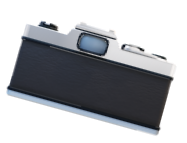}} & \multicolumn{1}{c}{\includegraphics[height=1.99em]{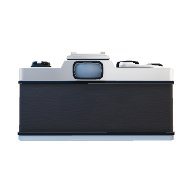}} & \multicolumn{1}{c}{\includegraphics[height=1.75em]{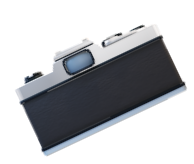}} & \multicolumn{1}{c}{\includegraphics[height=1.99em]{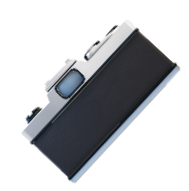}} \\
\addlinespace[4pt]
& \textbf{Parameter ($p$)}
  & $[-45^{\circ},-20^{\circ})$
  & $[-20^{\circ},-5^{\circ})$
  & $[-5^{\circ},5^{\circ}]$
  & $(5^{\circ},20^{\circ}]$
  & $(20^{\circ},45^{\circ}]$ \\
\midrule
&  & \multicolumn{5}{c}{\textbf{Pitch}} \\
\cmidrule(lr){3-7}
& \textbf{Term ($t$)}
  & Large tilt-down
  & Small tilt-down
  & Near straight-on shot
  & Small tilt-up
  & Large tilt-up \\
& \textbf{Example}
  & \multicolumn{1}{c}{\includegraphics[height=2.3em]{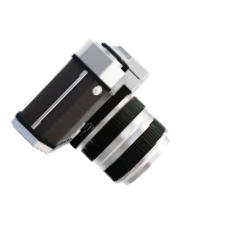}} & \multicolumn{1}{c}{\includegraphics[height=2.3em]{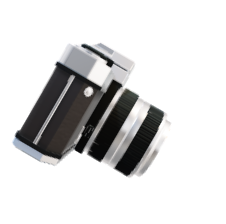}} & \multicolumn{1}{c}{\includegraphics[height=2.68em]{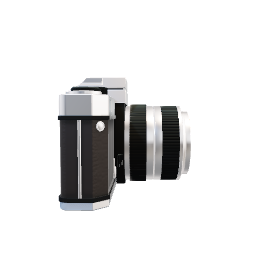}} & \multicolumn{1}{c}{\includegraphics[height=2.1em]{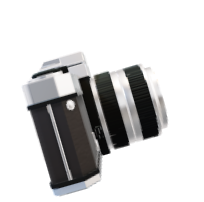}} & \multicolumn{1}{c}{\includegraphics[height=2.2em]{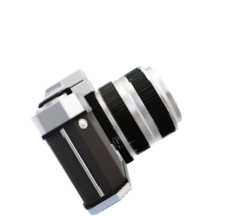}} \\
\addlinespace[4pt]
& \textbf{Parameter ($p$)}
  & $[-45^{\circ},-20^{\circ})$
  & $[-20^{\circ},-5^{\circ})$
  & $[-5^{\circ},5^{\circ}]$
  & $(5^{\circ},20^{\circ}]$
  & $(20^{\circ},45^{\circ}]$ \\
\midrule
&  & \multicolumn{5}{c}{\textbf{FoV}} \\
\cmidrule(lr){3-7}
& \textbf{Term ($t$)}
  & Close-up
  & Medium shot
  & Wide-angle
  & Ultra wide-angle
  &  \\
& \textbf{Example}
  & \multicolumn{1}{c}{\includegraphics[height=2.4em]{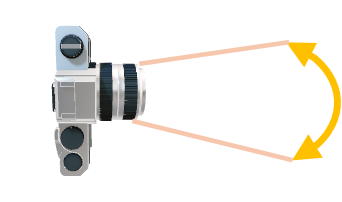}} & \multicolumn{1}{c}{\includegraphics[height=2.4em]{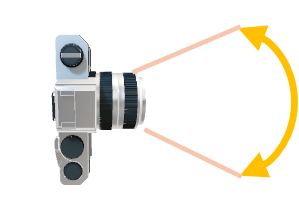}} & \multicolumn{1}{c}{\includegraphics[height=2.58em]{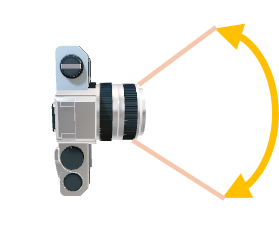}} & \multicolumn{1}{c}{\includegraphics[height=2.58em]{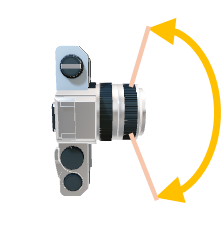}} \\
\addlinespace[4pt]
& \textbf{Parameter ($p$)}
  & $[20^{\circ},35^{\circ})$
  & $[35^{\circ},65^{\circ})$
  & $[65^{\circ},90^{\circ})$
  & $[90^{\circ},105^{\circ}]$
  &  \\
\bottomrule
\end{tabular}
\label{tab:photographic_term}
\end{table}
\section{Implementation Details}
\label{sec:appendix_implementation}

\subsection{Camera Parameters to Photographic Terms} To bridge the gap between the detailed numerical values of camera parameters and the highly abstracted understanding capability learned by LMMs, we propose using professional photographic terms as intermediate supervision for our framework. Specifically, we quantize the range of each camera parameter and map them to the following photographic terms: (i) Roll: large counterclockwise Dutch angle, small counterclockwise Dutch angle, near level shot, small clockwise Dutch angle, large clockwise Dutch angle; (ii) Pitch: large tilt-down, small tilt-down, near straight-on shot, small tilt-up, large tilt-up; (iii) FoV: close-up, medium shot, wide-angle, ultra wide-angle. As quantized abstractions of camera parameters, these terms are combined with textual scene descriptions to express global spatial arrangements in a linguistically accessible form. The detailed mapping relationship between the camera parameters and photographic terms is listed in Table~\ref{tab:photographic_term}.

\subsection{Prompt Design for Multimodal Tasks} 
\label{sec:appendix_implementation_prompt}
For text-to-image controllable generation, we use the following prompt template to format user instructions: \texttt{User: Generate an image: <caption>\textbackslash n Assistant:}''. The \texttt{<caption>} includes both the image description and the numerical camera parameters (roll, pitch, FoV, all in radians). For camera understanding, we employ the following prompt template to format the basic user instruction: \texttt{User: <image><question>\textbackslash n Assistant:}''. The \texttt{<question>} can be set as ``\texttt{Describe the image in detail. Then reason its spatial distribution and estimate its camera parameters (roll, pitch, and field-of-view)}''. For cross-view camera-controllable generation, the prompt template is formatted as: ``\texttt{User: Generate a target image given an initial view: <image><caption>\textbackslash n Assistant:}''. Here, \texttt{<image>} denotes the initial view token from the image tokenizer, while \texttt{<caption>} represents the target image description along with the target camera parameters (roll, pitch, yaw, and FoV, all in radians). During cross-view instruction tuning, we randomly set the \texttt{<caption>} to null with a probability of 0.5, thereby enabling both text-free and text-conditioned image-to-image generation. When applying the spatial reasoning paradigm, we switch to a new \texttt{<question>} for camera understanding: ``\texttt{Reason the spatial distribution of this image in a thinking mode, and then estimate its camera parameters (roll, pitch, and field-of-view)}''. For generation, we first enrich the vanilla prompt using our model with the template: ``\texttt{User: <caption><question>\textbackslash n Assistant:}''. Here, \texttt{<caption>} refers to the vanilla image description, and \texttt{<question>} is ``\texttt{Given a scene description and corresponding camera parameters, merge them into a coherent prompt and generate an accurate visualization that highlights visual cues for spatial reasoning}''. For other instruction tuning tasks, \texttt{<question>} is set to ``\texttt{Given the initial view and the camera parameters of the target view with the deviation yaw angle, how would you describe the target image to build a replica of the scene?}'' for spatial imagination, and ``\texttt{Estimate the camera parameters (roll, pitch, and field-of-view) of this image. And then predict the deviation camera yaw angle and pitch angle of the target view with high photographic aesthetics.}'' for photographic guidance.

\begin{figure}[t]
    \centering
    \begin{subfigure}{1\textwidth}
        \centering
        \includegraphics[width=1\linewidth]{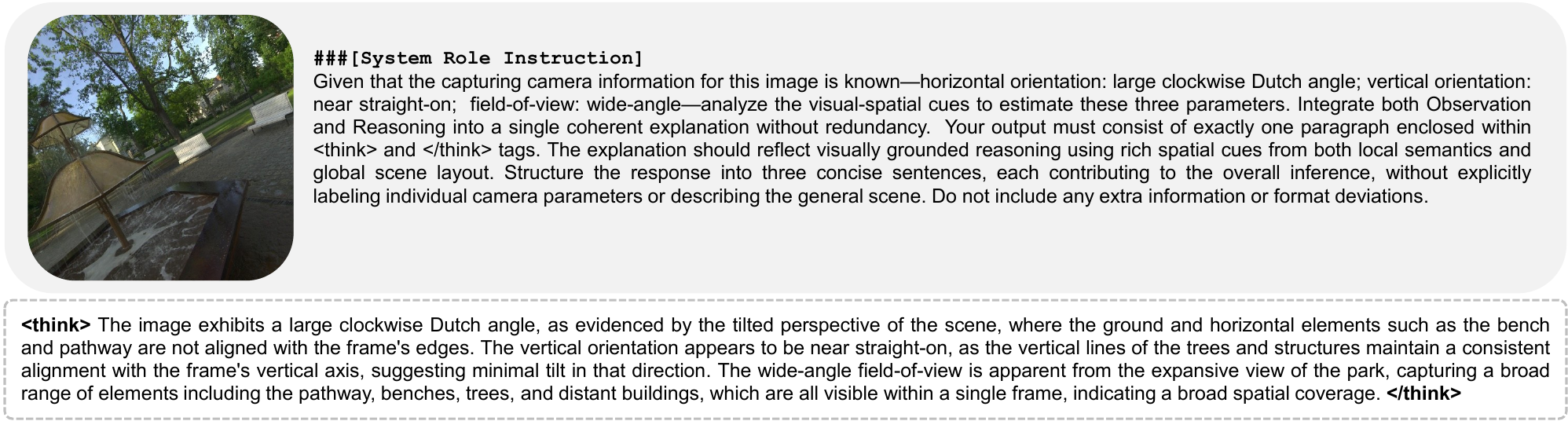}
        \caption{The prompt to generate the reasoning caption for \textit{thinking with camera}.}
    \end{subfigure}
    
    \begin{subfigure}{1\textwidth}
        \centering
        \includegraphics[width=1\linewidth]{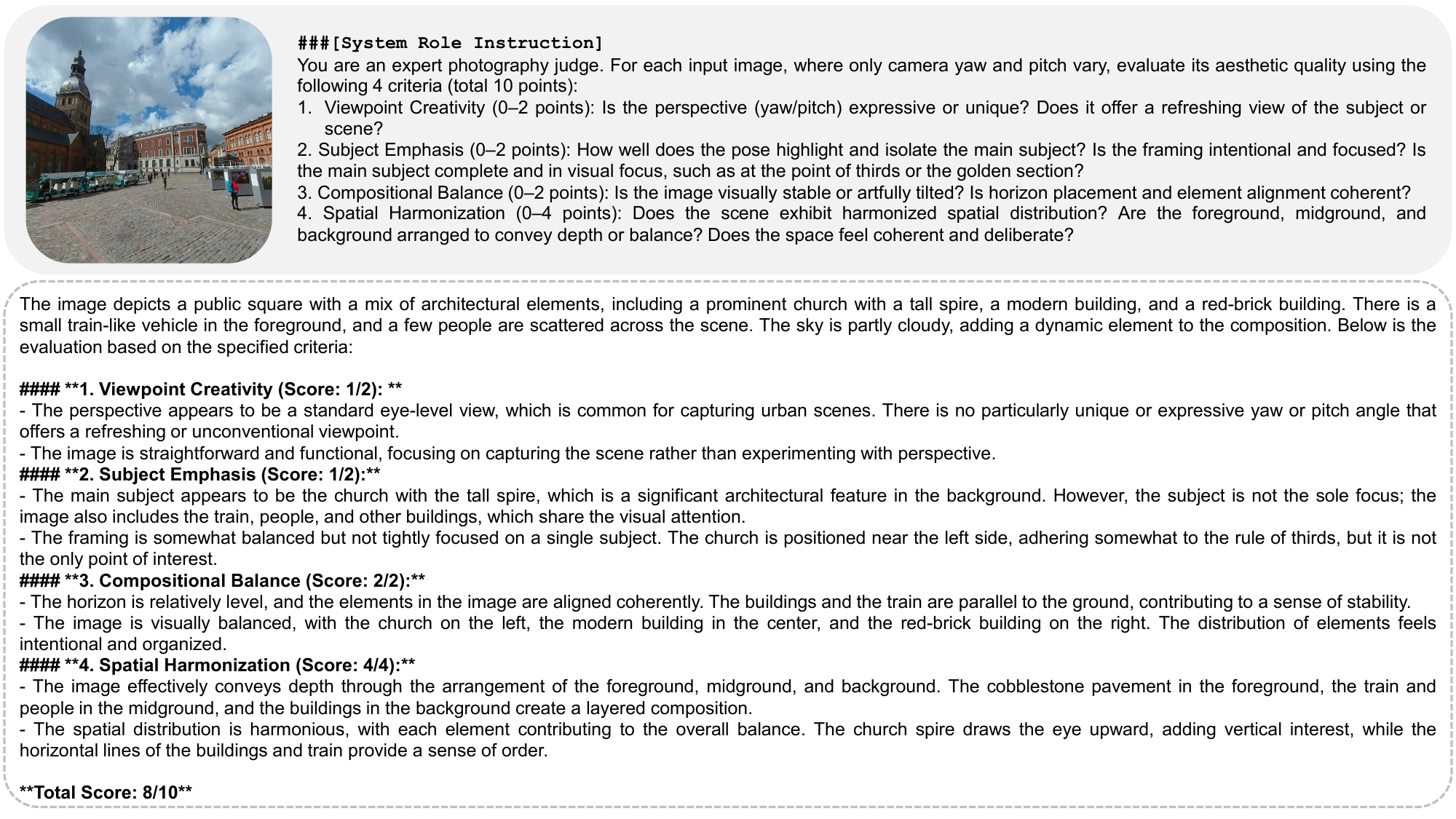}
        \caption{The prompt to generate the photographic aesthetic score for the photographic guidance task.}
    \end{subfigure}

    \caption{\textbf{Examples of the designed prompts for captioning our dataset: (a) reasoning caption, (b) photographic aesthetic caption.} For each sample, we visualize the input image, the prompt template for captioning, and the caption results from LMMs~\citep{bai2025qwen2}.}
    \label{fig:prompt_dataset}
\end{figure}

\begin{table}[t]
\centering
%\footnotesize
\scriptsize
\renewcommand{\arraystretch}{1.0}%
\setlength\tabcolsep{4.1pt}%
\caption{\textbf{Additional evaluation results on camera understanding.} The comparison methods are evaluated on the Stanford2D3D~\citep {Stanford2d3d} dataset.
\label{tab:cam_und_evaluation_Stanford2D3D}
}
\begin{tabular}{clcccccccccccc}
\toprule
&\multirow{2}{*}[-.4em]{Approach} 
& \multicolumn{4}{c}{Roll [degrees]} 
& \multicolumn{4}{c}{Pitch [degrees]} 
& \multicolumn{4}{c}{FoV [degrees]}\\
\cmidrule(lr){3-6}
\cmidrule(lr){7-10}
\cmidrule(lr){11-14}
&& error\,$\downarrow$ & \multicolumn{3}{c}{AUC\,$\triangleright$\,1/5/10\degree\,$\uparrow$}
& error\,$\downarrow$ & \multicolumn{3}{c}{AUC\,$\triangleright$\,1/5/10\degree\,$\uparrow$}
& error\,$\downarrow$ & \multicolumn{3}{c}{AUC\,$\triangleright$\,1/5/10\degree\,$\uparrow$} \\
\midrule
\multirow{9}{*}{\begin{sideways}\textbf{Stanford2D3D}\end{sideways}}
&{DeepCalib}~\citep{lopez2019deep}      &          \01.59 &          33.8 &          63.9 & \cthird  79.2 &          \02.58 &          21.6 &          46.9 &          65.7 &          \06.67 &          \08.1 &           20.6 &          37.6 \\
&{Perceptual}~\citep{hold2018perceptual}              &          \02.08 &          26.8 &          53.8 &          70.7 &          \03.17 &          21.5 &          41.8 &          57.8 &           13.84 &          \02.8 &          \07.7 &          16.1 \\
&{CTRL-C}~\citep{lee2021ctrl}                       &          \03.04 &          23.2 &          43.0 &          56.9 &          \03.43 &          18.3 &          38.6 &          53.8 &          \08.50 &          \07.7 &           18.2 &          31.5 \\
&{MSCC}~\citep{Song2024MSCC}                  &          \03.43 &          13.5 &          36.8 &          57.3 &          \02.64 &          22.6 &          45.0 &          60.5 & \cthird  \05.81 & \cthird  \09.6 & \cthird   23.8 & \cthird  41.6 \\
&{ParamNet}~\citep{jin2023perspective} & \cthird  \01.14 & \cthird  44.6 & \cthird  73.9 & \cthird 84.8 & \cthird  \01.94 & \cthird  29.2 & \cthird  56.7 & \cthird 73.1 &          \09.01 &          \05.8 &           14.3 &          27.8 \\
&{SVA}~\citep{lochman2021minimal}                            &              -  &          21.7 &          24.6 &          25.8 &              -  &          15.4 &          19.9 &          22.4 &              -  &          \06.2 &           11.5 &          15.2 \\
&{UVP}~\citep{pautrat2023vanishing}    & \cthird \00.52 & \cthird 65.3 & \cthird 74.6 &          79.1 & \cthird \00.95 & \cthird 51.2 & \cthird 63.0 & \cthird  69.2 & \cthird \03.65 & \csecond   22.2 & \cthird  39.5 & \cthird 51.3 \\
&{GeoCalib}~\citep{veicht2024geocalib}                                   & \csecond  \00.40 & \csecond  83.1 & \csecond  91.8 & \csecond  94.8 & \csecond  \00.93 & \csecond  52.3 & \csecond  74.8 & \csecond  84.6 & \csecond  \03.21 & \cthird  17.4 & \csecond   40.0 & \csecond  59.4 \\
&\textbf{Puffin (Ours)}                                   & \cfirst  \00.26 & \cfirst  96.6 & \cfirst  99.0 & \cfirst  99.4 & \cfirst  \00.48 & \cfirst  82.0 & \cfirst  93.6 & \cfirst  96.7 & \cfirst  \02.30 & \cfirst  23.4 & \cfirst   51.2 & \cfirst  71.4 \\
\bottomrule
\end{tabular}
\end{table}
\section{Additional Experiments}
\label{sec:apendix_experiments}
\subsection{Camera Understanding}
\label{sec:appendix_exp_cam_und}
Note that our training dataset consists of images rendered from the source panoramas in Stanford2D3D~\citep{Stanford2d3d}. Although the sampled perspective images and camera parameters differ from those in the test set~\citep{Stanford2d3d}, we exclude these results from the main evaluation to ensure rigor and report them in Table~\ref{tab:cam_und_evaluation_Stanford2D3D} only for reference. 

\begin{figure}[t]
    \centering
    \begin{subfigure}{1\textwidth}
        \centering
        \includegraphics[width=1\linewidth]{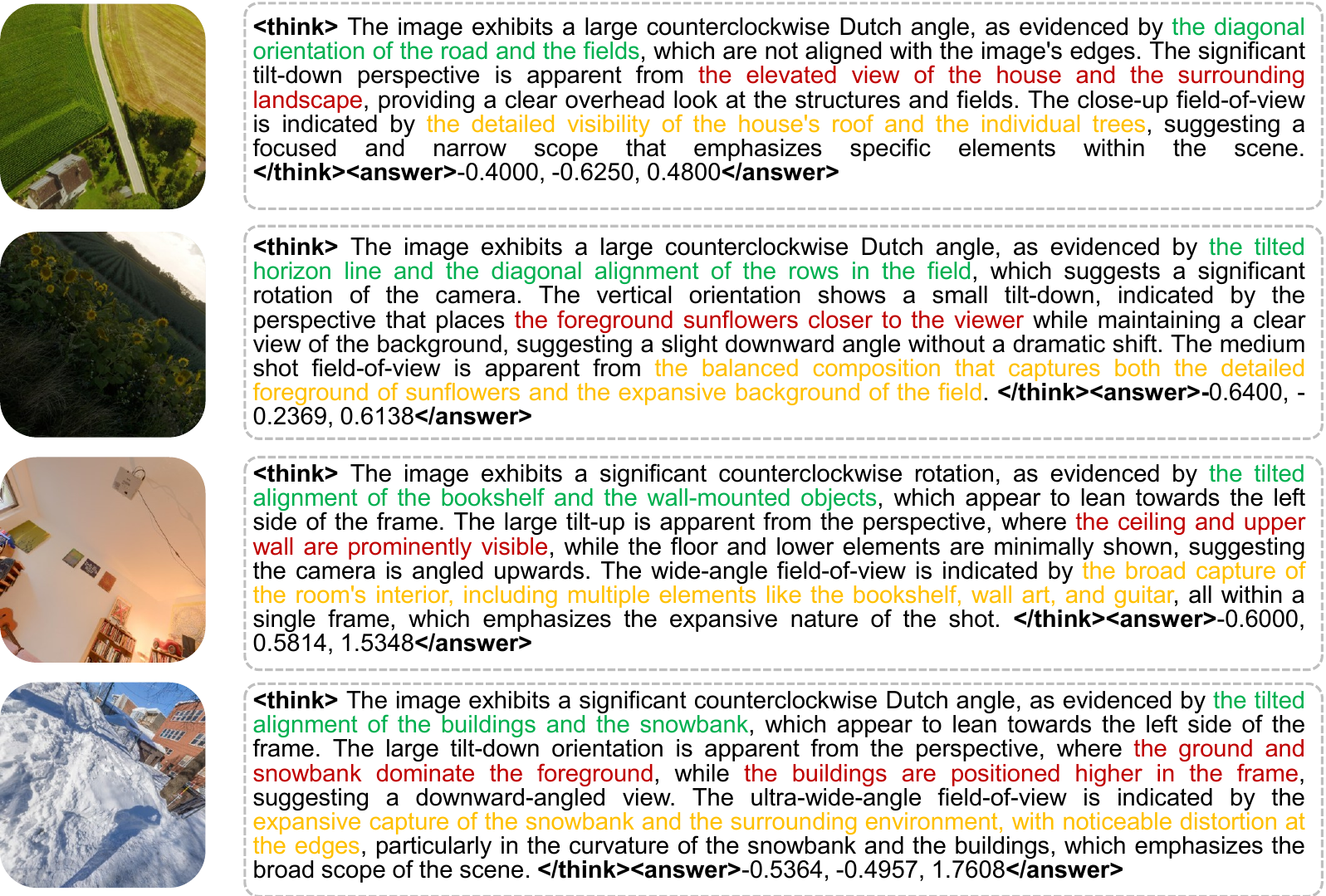}
    \end{subfigure}
    
    \vspace{0.1cm}
    
    \begin{subfigure}{1\textwidth}
        \centering
        \includegraphics[width=1\linewidth]{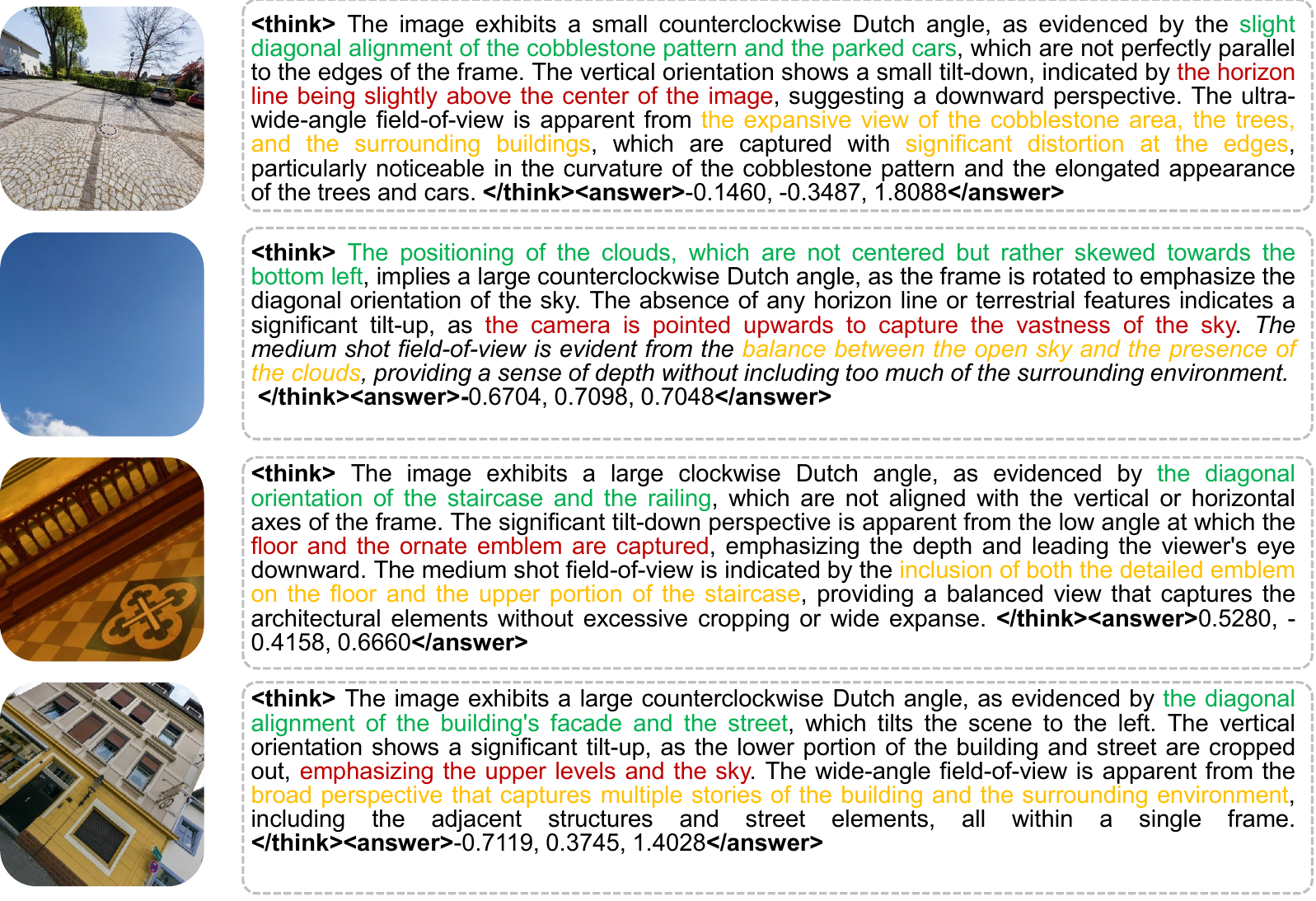}
    \end{subfigure}
    %\vspace{-0.3cm}
       \caption{\textbf{Visualization on our spatial reasoning process for camera understanding.} We highlight the reasoned spatially grounded visual cues regarding each camera parameter using different colors: \textcolor{Green}{roll}, \textcolor{darkred}{pitch}, and \textcolor{Orange}{FoV}.
    }
    \label{fig:und_reasoning_vis}
    
\end{figure}

\begin{figure}[t]
    \centering
    \begin{subfigure}{1\textwidth}
        \centering
        \includegraphics[width=1\linewidth]{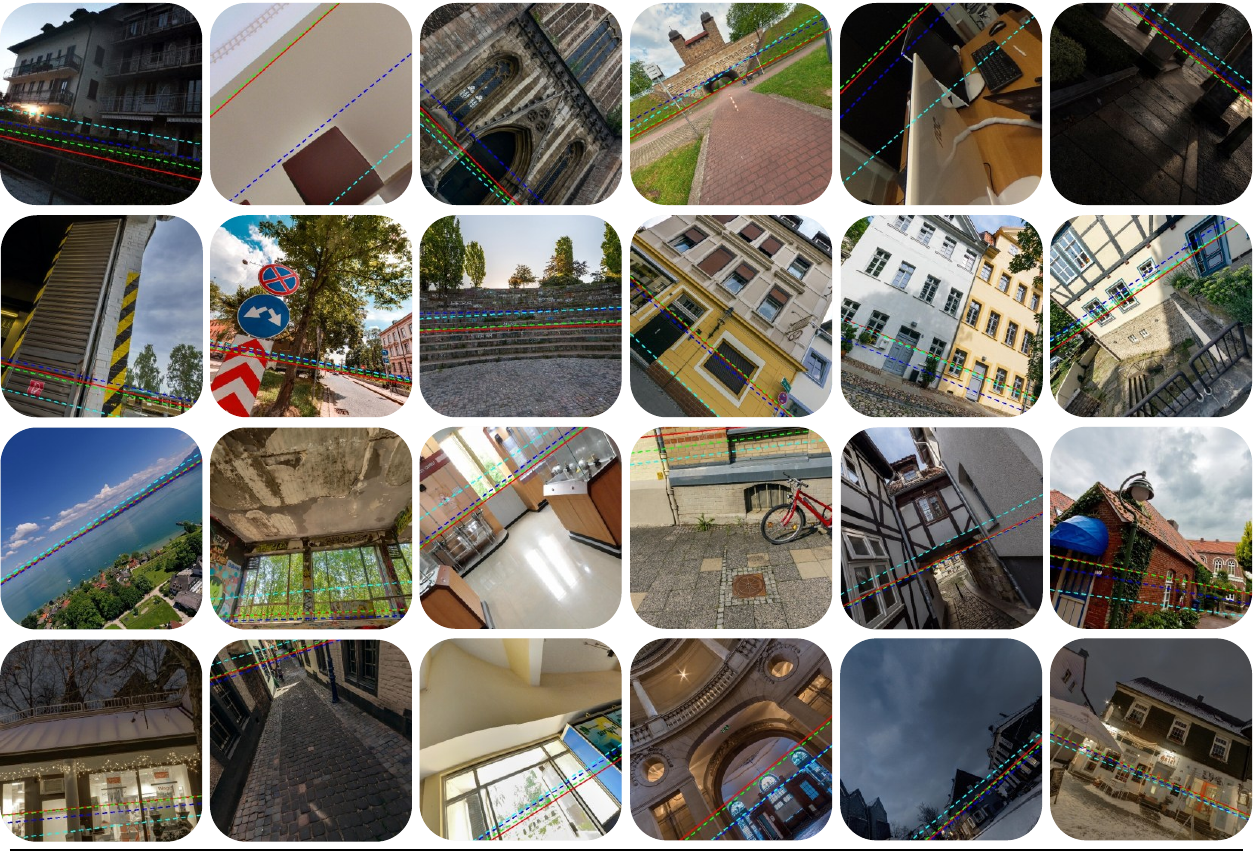}
    \end{subfigure}
    
    %\vspace{-0.3cm}
    
    \begin{subfigure}{1\textwidth}
        \centering
        \includegraphics[width=1\linewidth]{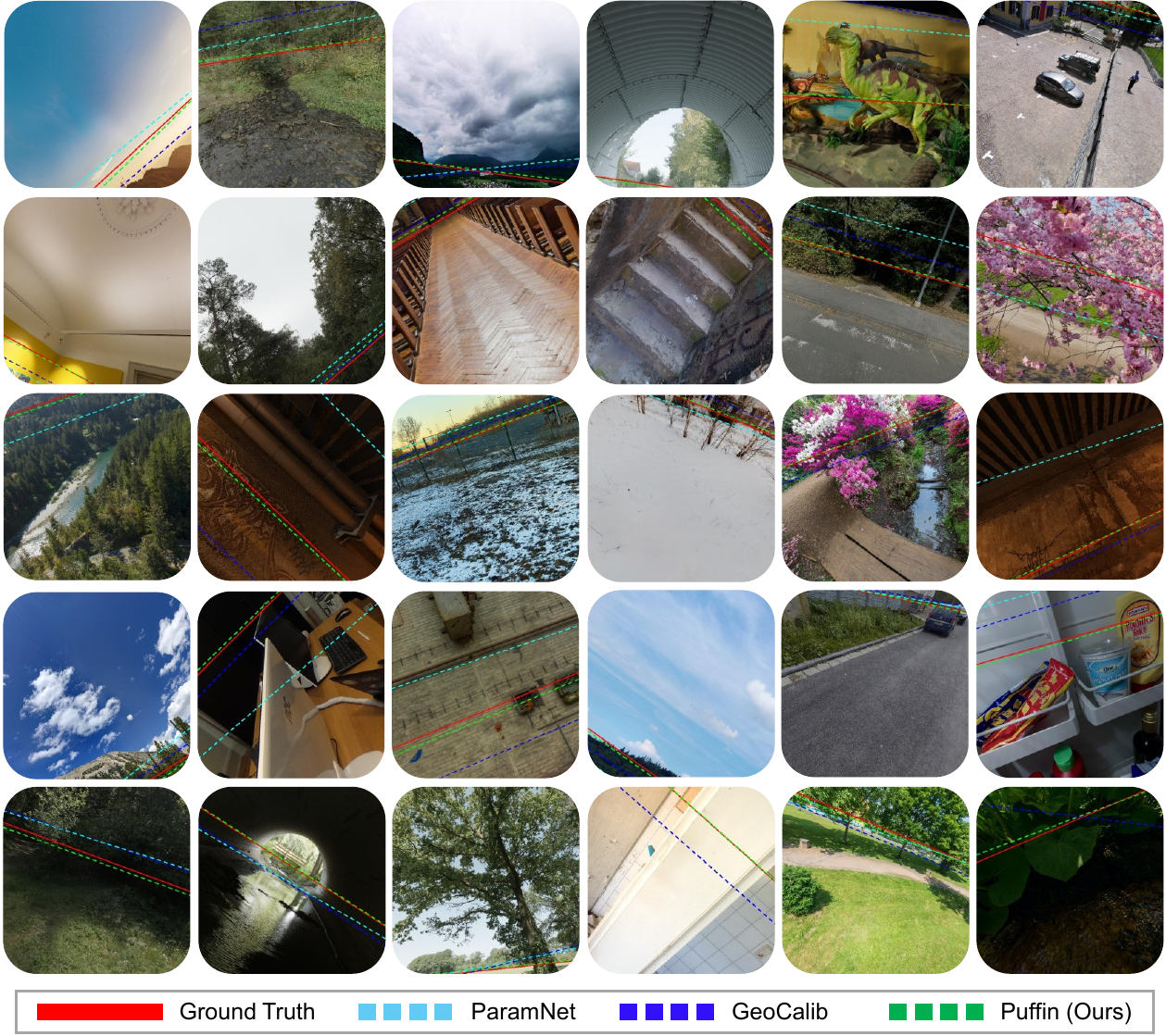}
    \end{subfigure}
    %\vspace{-0.3cm}
       \caption{\textbf{Qualitative evaluations on the camera understanding methods with horizon line visualization.} We show the common cases (with architectures or indoor) and challenging cases (with few geometric features or significant tilted camera poses) at the top and bottom, respectively.
    }
    \label{fig:horizon}
    
\end{figure}

\begin{figure}[t]
    \centering
    \begin{subfigure}{0.98\textwidth}
        \centering
        \includegraphics[width=1\linewidth]{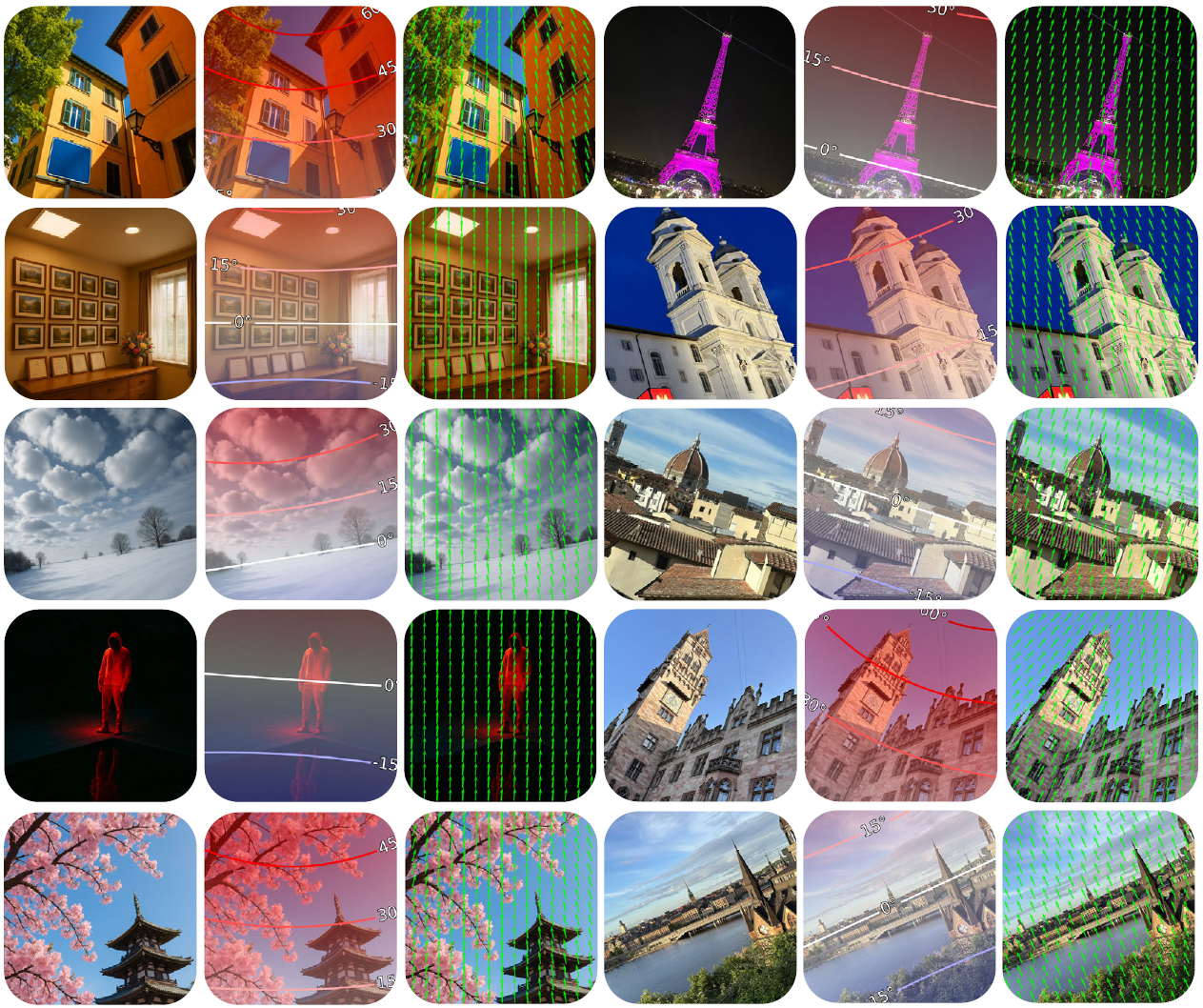}
    \end{subfigure}
    
    %\vspace{-0.3cm}
    
    \begin{subfigure}{0.98\textwidth}
        \centering
        \includegraphics[width=1\linewidth]{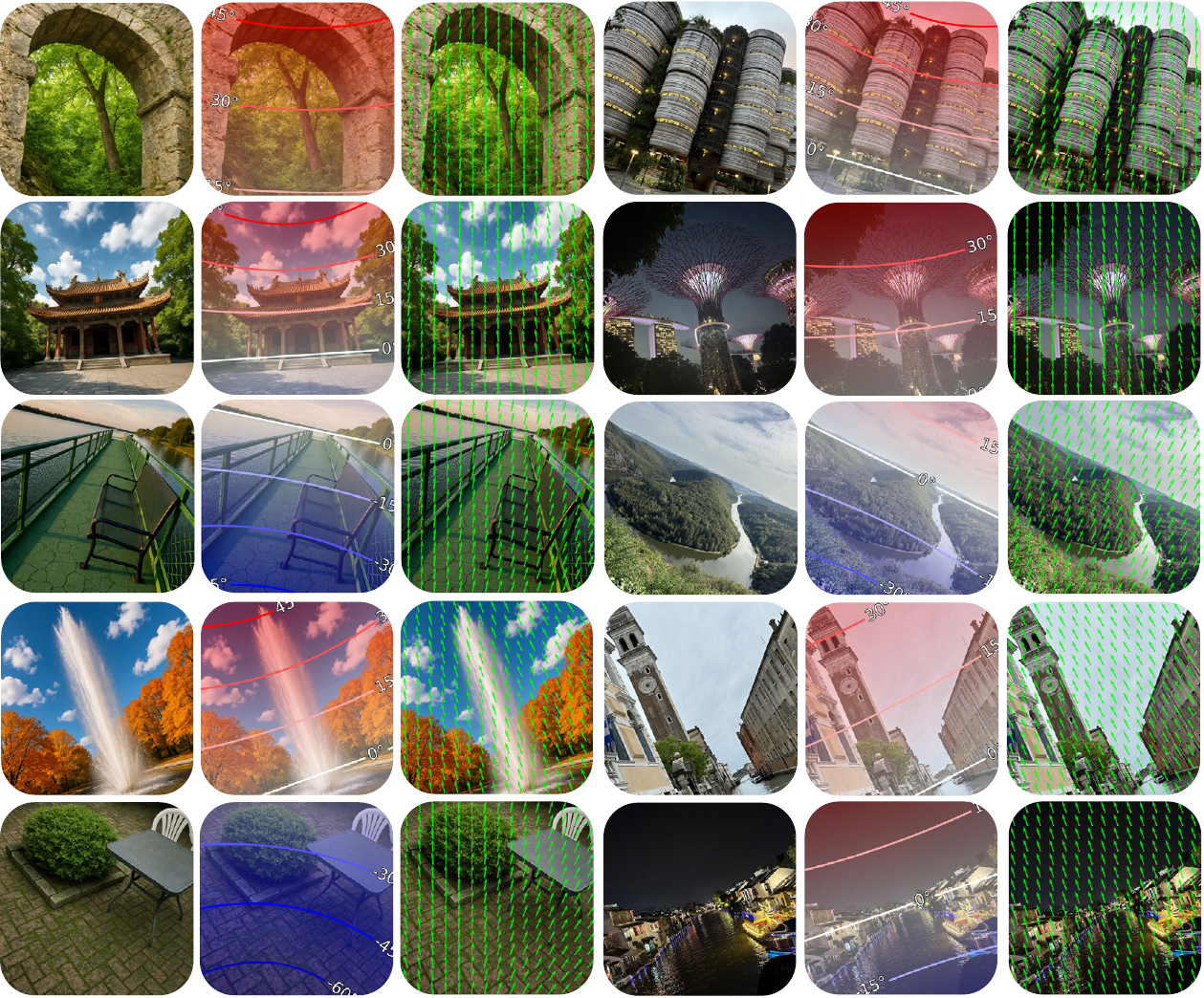}
    \end{subfigure}
    %\vspace{-0.3cm}
       \caption{\textbf{Our camera understanding on AIGC images~\citep{GTP-4o} (left) and real-world photographs (right).}
    }
    \label{fig:cam_und_map}
    
\end{figure}

We show more visualization results on the proposed thinking with camera for camera understanding in Figure~\ref{fig:und_reasoning_vis}. Qualitative evaluations of the camera understanding methods with horizon line visualization are illustrated in Figure~\ref{fig:horizon}. We visualize additional camera understanding results (with camera maps converted from the predicted camera parameters) on diverse inputs, including AIGC images~\citep{GTP-4o} and real-world photographs, in Figure~\ref{fig:cam_und_map}.

\subsection{Camera-Controllable Generation}
\begin{figure}
    \centering
    \includegraphics[width=1\linewidth]{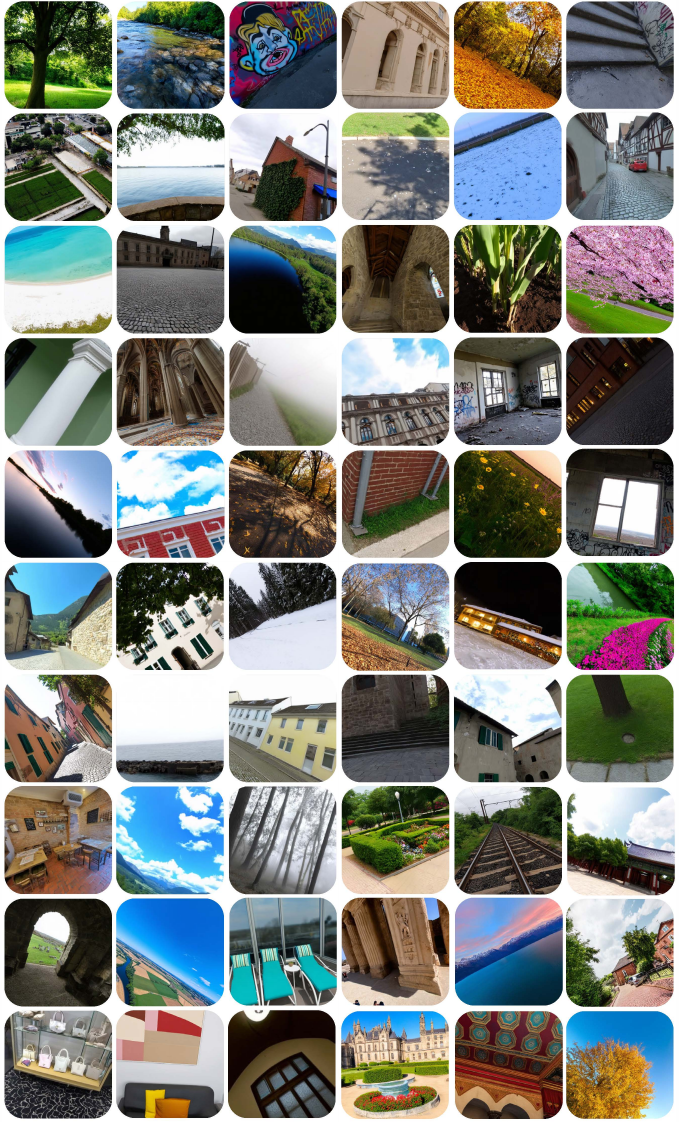}
    \caption{\textbf{Our camera-controllable generation results with various camera configurations.}
    }
    \label{fig:gallery}
    %\vspace{-0.3cm}
\end{figure}

\begin{figure}[t]
    \centering
    \begin{subfigure}{0.98\textwidth}
        \centering
        \includegraphics[width=0.98\linewidth]{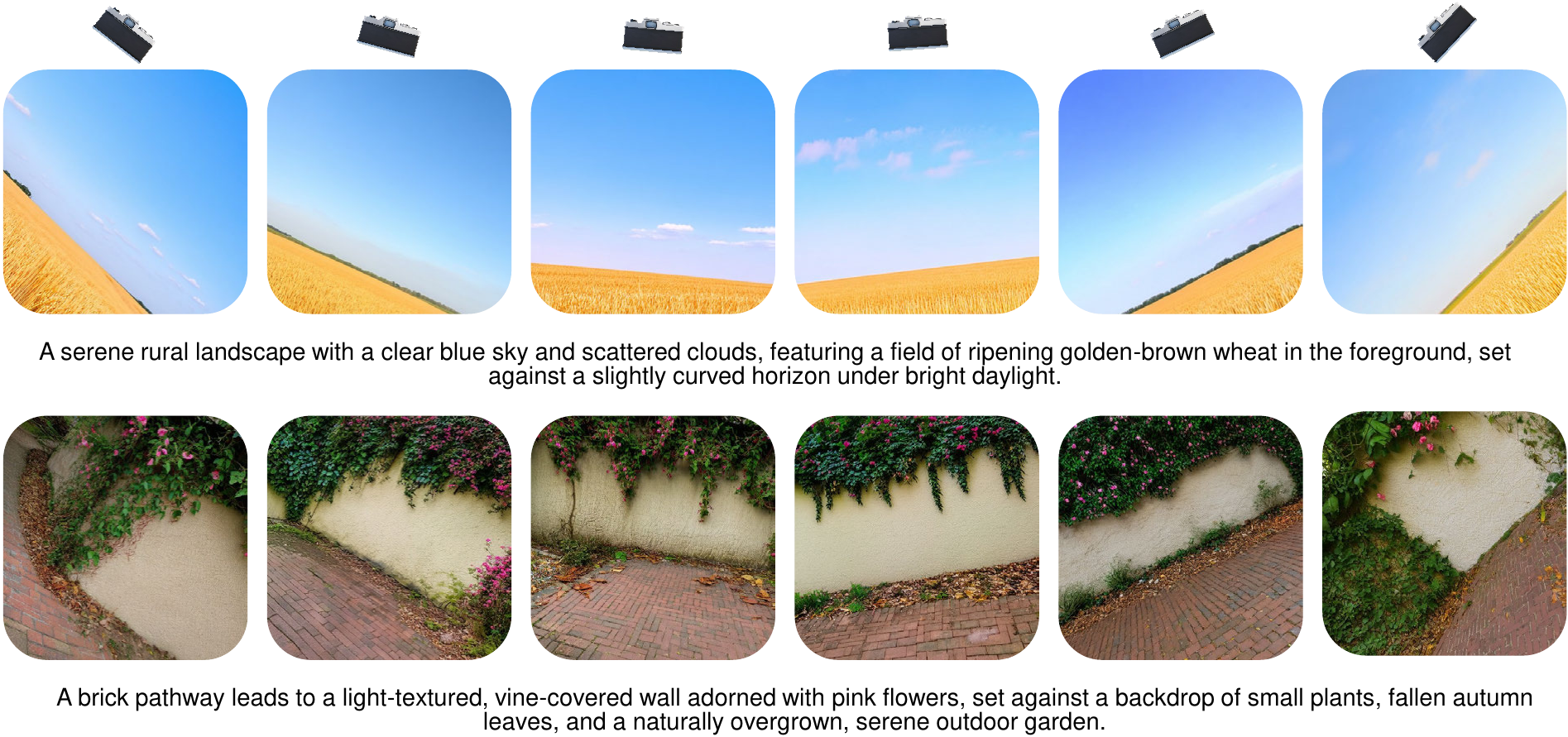}
        \caption{Text-to-image generation with varying roll angles.}
    \end{subfigure}
    
    \vspace{0.5em}
    
    \begin{subfigure}{0.98\textwidth}
        \centering
        \includegraphics[width=0.98\linewidth]{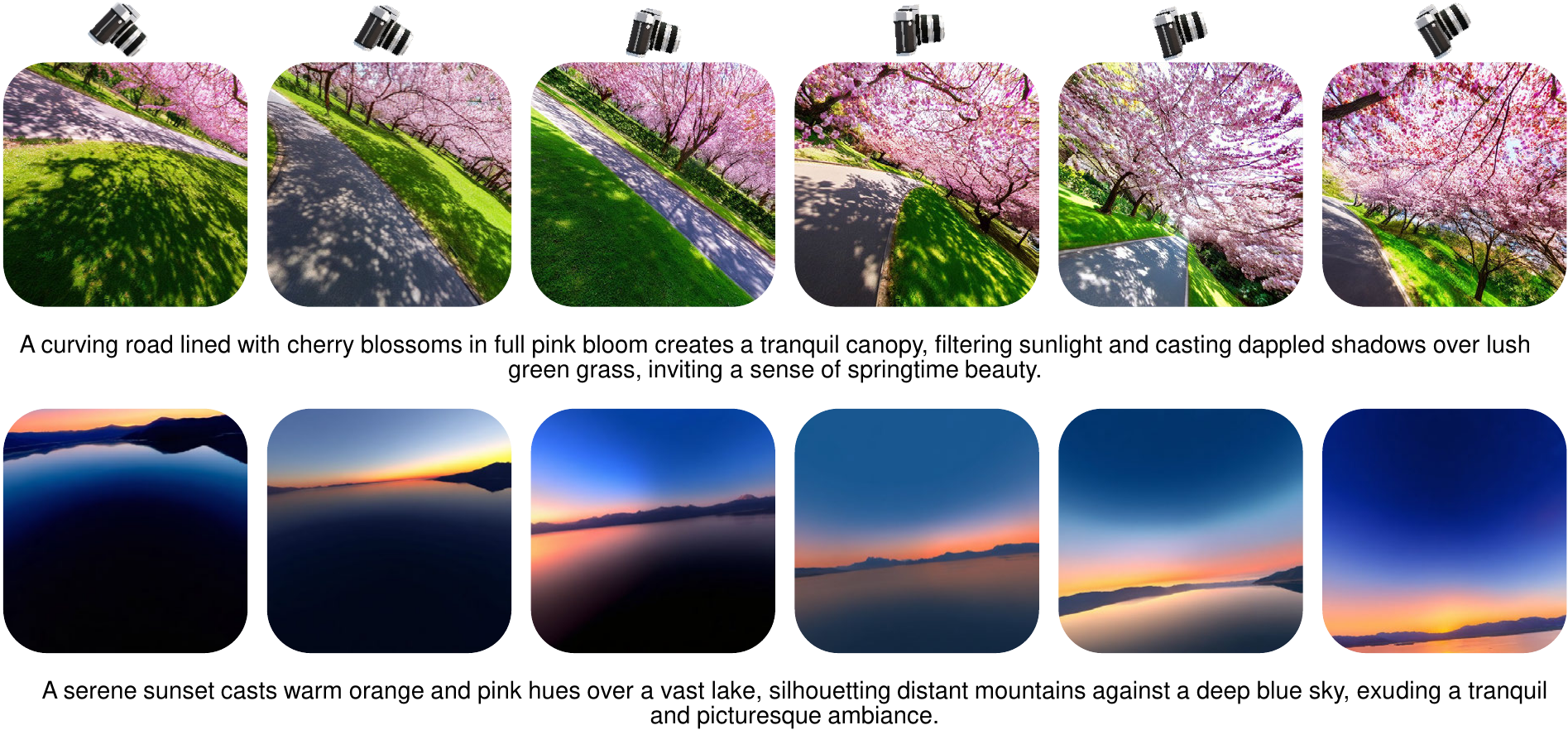}
        \caption{Text-to-image generation with varying pitch angles.}
    \end{subfigure}
    
    \vspace{0.5em}
    
    \begin{subfigure}{0.98\textwidth}
        \centering
        \includegraphics[width=0.98\linewidth]{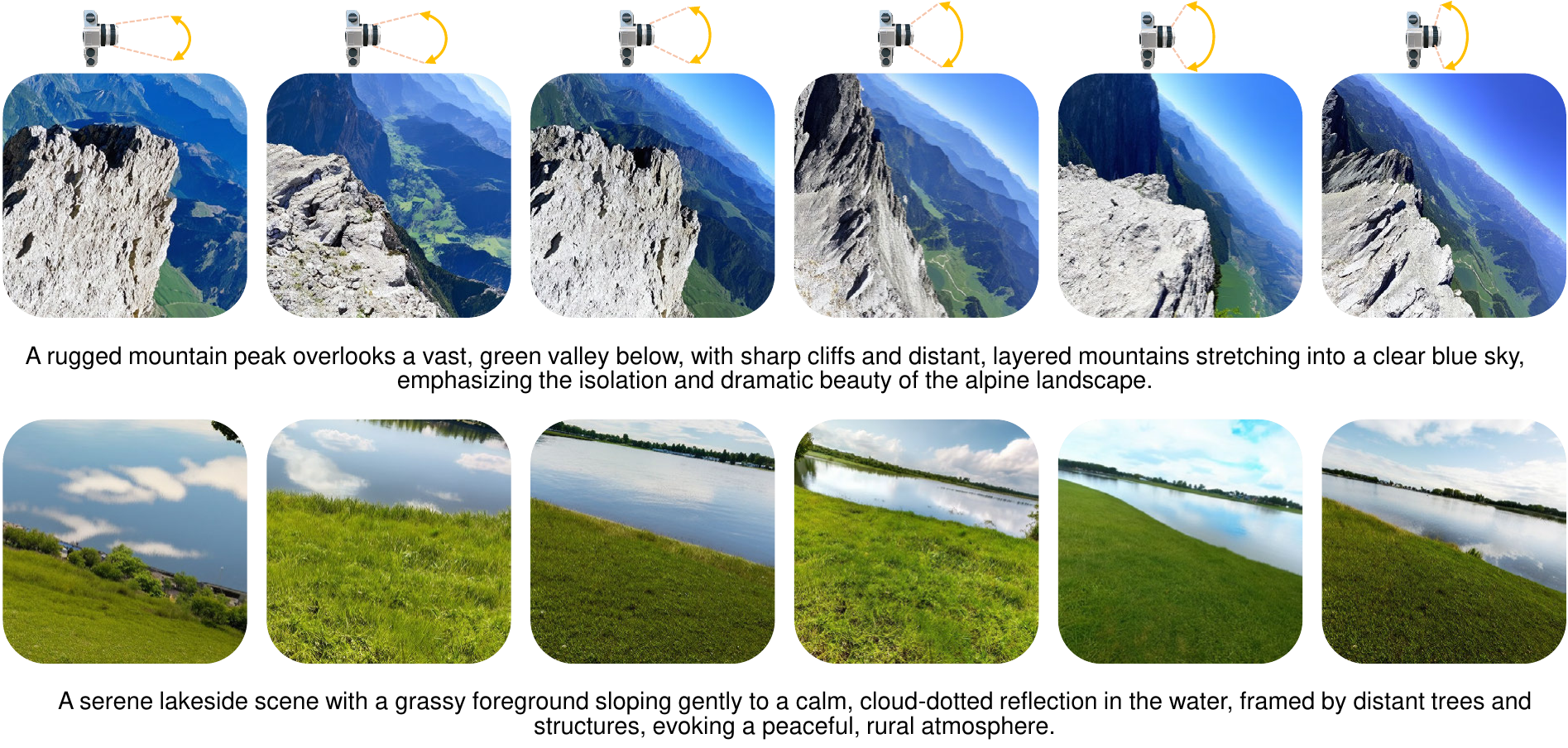}
        \caption{Text-to-image generation with varying FoVs.}
    \end{subfigure}

    \caption{\textbf{Text-to-image generation (single-view) with specific controls for each camera parameter.}}
    \label{fig:camctrl_t2i}
\end{figure}

Our camera-controllable generation results with various camera configurations are shown in Figure~\ref{fig:gallery}, and the text-to-image generation (single-view) results with specific controls for each camera parameter are presented in Figure~\ref{fig:camctrl_t2i}.

\subsection{Analysis}

\begin{table*}[t]
\captionsetup{aboveskip=2pt}
\centering
\scriptsize
    \caption{\textbf{Model size comparisons:} the specialized understanding and generation models (GeoCalib~\citep{veicht2024geocalib} and PreciseCam~\citep{bernal2025precisecam}), and the proposed unified camera-centric model.}
{%
\begin{tabular}{lccc}
\toprule
\multirow{1}{*}{\textbf{Model}} & \multirow{1}{*}{\textbf{Type}} & \multirow{1}{*}{\textbf{Parameters}} & \multirow{1}{*}{\textbf{GFLOPs}} 
\\ \midrule
GeoCalib~\citep{veicht2024geocalib}             & Specialized Model (Understanding)   & 28.9M & $3.47 \times 10^2$  \\
PreciseCam~\citep{bernal2025precisecam}     & Specialized Model (Generation) &  \01.3B & $2.67 \times 10^3$   \\
\textbf{Puffin-4M (Ours)}       & Unified Model & \04.4B & $2.92 \times 10^5$  \\
\bottomrule
\end{tabular}%
}% end resizebox
\label{tab:model_size}
\captionsetup{justification=raggedright,singlelinecheck=false}
\end{table*}
\noindent\textbf{Model Size.}
We show the comparison results on the total parameter count and FLOPs with previous understanding and generation models in Table~\ref{tab:model_size}, such as GeoCalib~\cite{veicht2024geocalib} and PreciseCam~\cite{jin2023perspective}. While Puffin is larger than previous specialized models, it replaces separate understanding and generation networks with a single unified model that handles both tasks within one framework. This design not only simplifies deployment, but also allows us to fully exploit a high-capacity backbone when training on large-scale multimodal datasets. In terms of overall parameter count and FLOPs, Puffin remains substantially more affordable than recent general-purpose unified multimodal models such as Bagel (14B)~\cite{deng2025emerging} and Qwen-Image (20B)~\cite{wu2025qwen}.

\begin{figure}[h]
    \centering
    \includegraphics[width=1\linewidth]{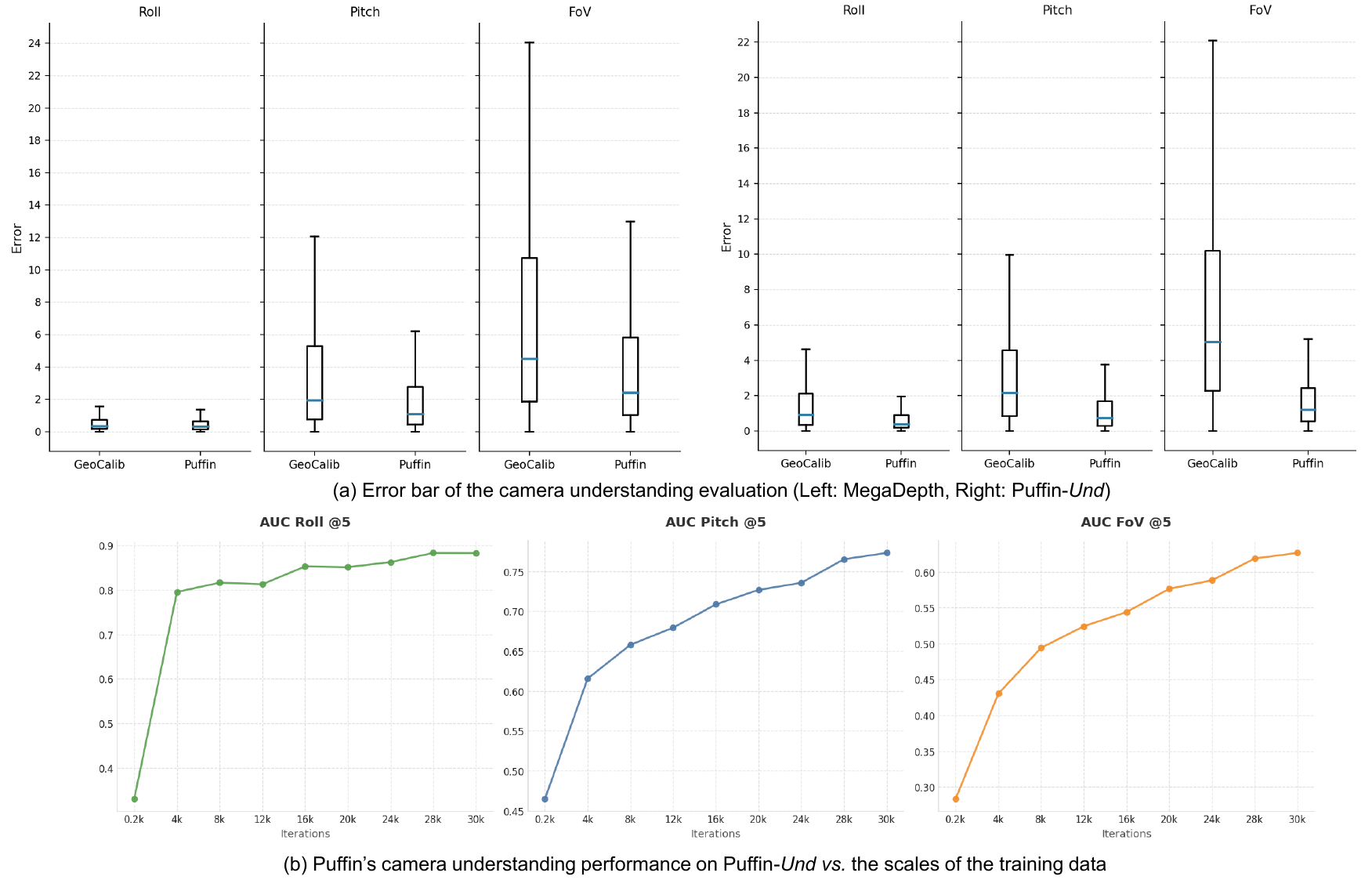}
    \caption{\textbf{More data analysis of the experiments.} (a) We show the error bar of the camera understanding evaluation on GeoCalib~\citep{veicht2024geocalib} and Puffin. (b) We show how the performance of Puffin scales with data size across different camera parameters.  
    }
    \label{fig:error_bar_data_scale}
    %\vspace{-0.4cm}
\end{figure}

\noindent\textbf{Error Analysis.}
In addition to the metrics (error and AUC) used in the main experiments, we also add the error bars for the representative baselines and our method in Figure~\ref{fig:error_bar_data_scale}(a) to provide a clearer and more comprehensive comparison. The results show that Puffin exhibits better robustness across the entire data distribution. The improvement is consistent across all intrinsics/extrinsics components and remains stable even under challenging camera configurations.

\noindent\textbf{Model \textit{vs}. Data.}
We conduct experiments where the comparison method~\citep{veicht2024geocalib} is re-trained on the same dataset (Puffin-4M) as ours, strictly following its original training recipe. Interestingly, we find that the re-trained GeoCalib~\citep{veicht2024geocalib} on our 4M dataset slightly underperforms the original GeoCalib model trained on its 40K dataset. The detailed evaluation results on the Puffin-\textit{Und} test set are reported in Table~\ref{tab:cam_und_evaluation_puffin4m}.

\begin{table}[t]
\centering
%\footnotesize
\scriptsize
\renewcommand{\arraystretch}{1.0}%
\setlength\tabcolsep{4.1pt}%
\caption{\textbf{Additional evaluation results on camera understanding} by re-training the comparison method~\citep{veicht2024geocalib} on our constructed Puffin-4M dataset*.}
\label{tab:cam_und_evaluation_puffin4m}
\begin{tabular}{clcccccccccccc}
\toprule
&\multirow{2}{*}[-.4em]{Approach} 
& \multicolumn{4}{c}{Roll [degrees]} 
& \multicolumn{4}{c}{Pitch [degrees]} 
& \multicolumn{4}{c}{FoV [degrees]}\\
\cmidrule(lr){3-6}
\cmidrule(lr){7-10}
\cmidrule(lr){11-14}
&& error\,$\downarrow$ & \multicolumn{3}{c}{AUC\,$\triangleright$\,1/5/10\degree\,$\uparrow$}
& error\,$\downarrow$ & \multicolumn{3}{c}{AUC\,$\triangleright$\,1/5/10\degree\,$\uparrow$}
& error\,$\downarrow$ & \multicolumn{3}{c}{AUC\,$\triangleright$\,1/5/10\degree\,$\uparrow$} \\
\midrule
\multirow{3}{*}{\begin{sideways}\textbf{}\end{sideways}}
&{GeoCalib*}~\citep{veicht2024geocalib}                                   & \cthird  \01.12 & \cthird  46.0 & \cthird  69.3 & \cthird  80.1 & \cthird  \02.54 & \cthird  24.5 & \cthird  47.5 & \cthird  65.4 & \cthird  \05.47 & \cthird  10.8 & \cthird   25.5 & \cthird  43.8 \\
&{GeoCalib}~\citep{veicht2024geocalib}                                 & \csecond  \00.92 & \csecond   53.6 & \csecond   73.9 & \csecond   82.6 & \csecond  \02.18 & \csecond   28.9 & \csecond   52.5 & \csecond   69.6 & \csecond  \05.04 & \csecond   12.4 & \csecond   28.0 & \csecond   45.8 \\
&\textbf{Puffin (Ours)}                                   & \cfirst  \00.41 & \cfirst  78.3 & \cfirst  91.0 & \cfirst  95.2 & \cfirst  \00.74 & \cfirst  60.2 & \cfirst  81.2 & \cfirst  90.0 & \cfirst  \01.21 & \cfirst  42.4 & \cfirst   70.5 & \cfirst  84.3 \\
\bottomrule
\end{tabular}
\end{table}

By carefully analyzing these results, we offer two explanations for this phenomenon: (i) Model capacity \textit{vs.} data scale. GeoCalib~\citep{veicht2024geocalib} is a relatively lightweight CNN-like architecture with around 29M parameters. When trained on a significantly larger 4M-scale dataset with broad scene and distribution coverage, it tends to underfit: its limited capacity cannot fully model the entire distribution, so it only fits some sub-distributions well while inevitably neglecting others. (ii) Consistent observations from previous experiments. The GeoCalib authors report a similar trend in their ablation: when training the network on a 5$\times$ larger dataset, the camera understanding performance slightly degrades on most benchmarks, and no clear improvement is observed; they also show that more advanced architectures can further boost performance. These observations are fully consistent with our findings.

In contrast, Puffin is built on a high-capacity LLM backbone, which, like other large multimodal models (\textit{e.g.}, LLaVA, Qwen-VL, InternVL), requires sufficiently large and diverse training data to avoid overfitting and to learn an accurate joint distribution over images, camera geometry, and language. In this sense, Puffin-4M is not merely a ``bonus”, but rather a necessary data regime for such a unified multimodal model that jointly supports both understanding and generation. 

Furthermore, we also conduct additional experiments by re-training our model on Puffin-4M of different scales (Stage-II SFT). As shown in Figure~\ref{fig:error_bar_data_scale} (b), we observe clear and consistent improvements as the data scale increases. Overall, these results suggest that the dataset and the model contribute jointly and should not be viewed in isolation.

Beyond the above scaling results in Figure~\ref{fig:error_bar_data_scale} (b), we find an interesting difference in trends across camera parameters. For roll, the model learns quickly even with relatively limited data, since it mainly relies on low- to mid-level geometric cues that are easy to capture (\textit{e.g.}, strongly slanted lines indicating a large roll). By contrast, estimating pitch and FoV requires more holistic and high-level spatial understanding, which cannot be sufficiently captured by local visual patterns alone and therefore benefits more from larger-scale data to form robust spatial reasoning concepts. This observation is consistent with our discussion in Section~\ref{sec:appendix_exp_cam_und}. Based on these trends, we believe that further scaling the dataset would bring additional gains in camera geometry understanding, especially for pitch and FoV.

\begin{figure}
    \centering
    \includegraphics[width=1\linewidth]{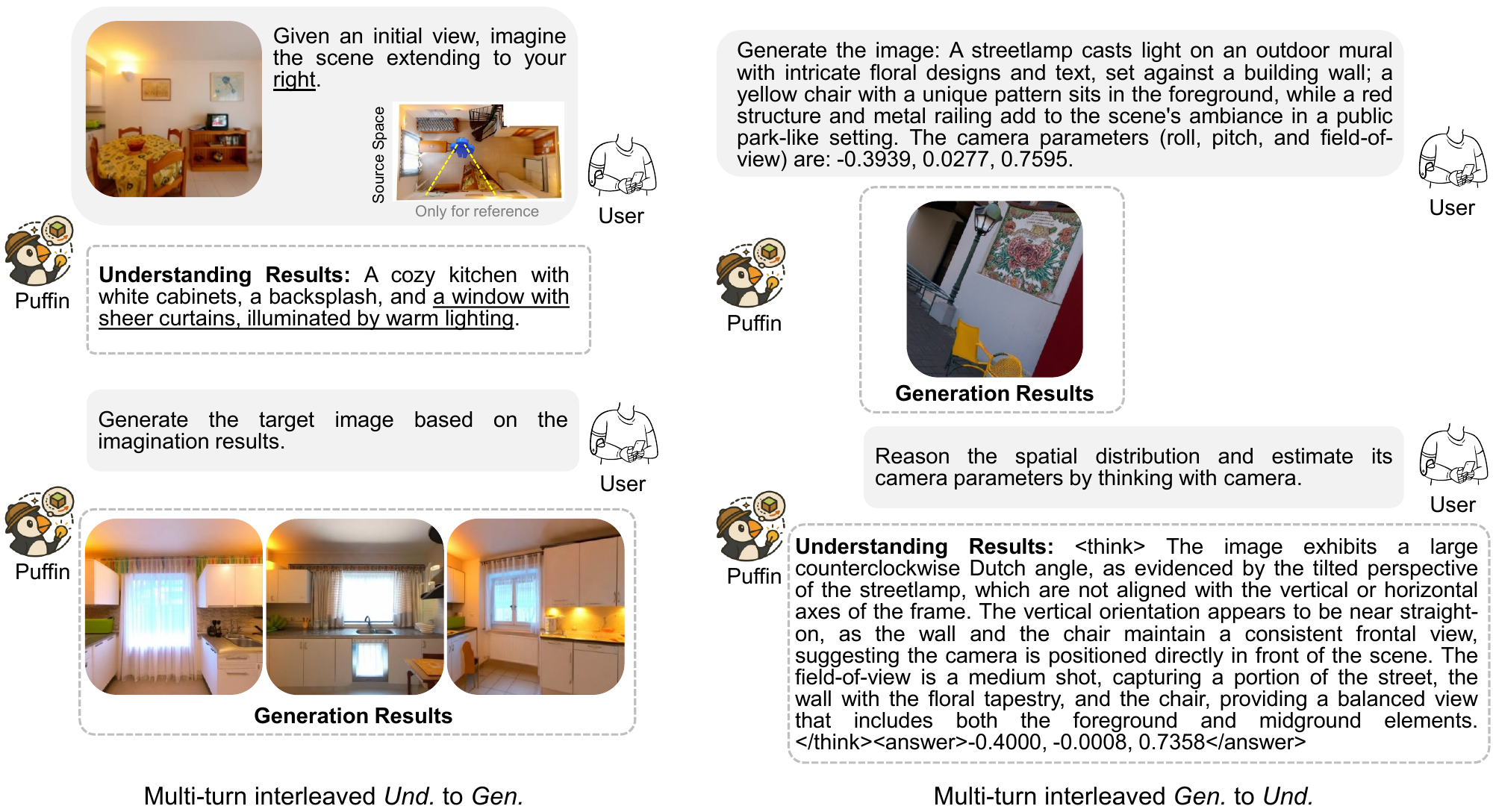}
    \caption{\textbf{Multi-turn interleaved capability of Puffin.}}
    \label{fig:multi_round_conversation}
    %\vspace{-0.4cm}
\end{figure}

\noindent\textbf{Multi-turn Interleaved Capability.}
Exploring the multi-round conversational capability of a unified multimodal model is meaningful. To this end, we conduct experiments on multi-turn interleaved conversations (generation $\rightarrow$ understanding and understanding $\rightarrow$ generation). As shown in Figure~\ref{fig:multi_round_conversation}, Puffin can carry out coherent interleaved dialogues conditioned on previous turns. Specifically, it produces consistent cross-view generation results based on its previous reasoning, and accurate camera understanding based on its own generated images. This demonstrates that Puffin not only supports both capabilities within a single framework but can also use them in an interactive way over multiple conversational rounds, without any task-specific switching or separate models.

\subsection{Downstream Applications}
\begin{figure}[t]
    \centering
    \begin{subfigure}{1\textwidth}
        \centering
        \includegraphics[width=1\linewidth]{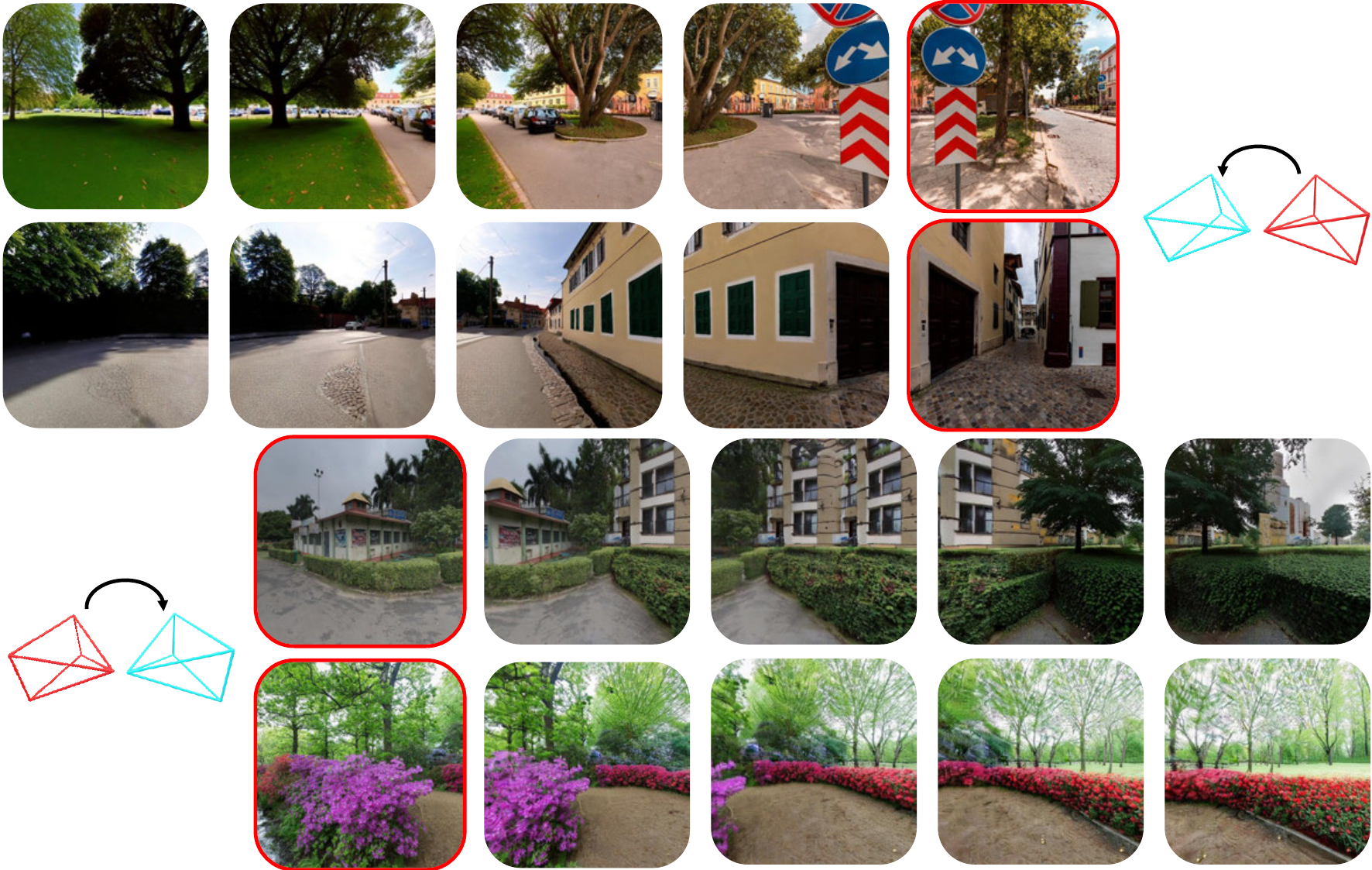}
    \end{subfigure}
    
    \vspace{0.3cm}
    
    \begin{subfigure}{1\textwidth}
        \centering
        \includegraphics[width=1\linewidth]{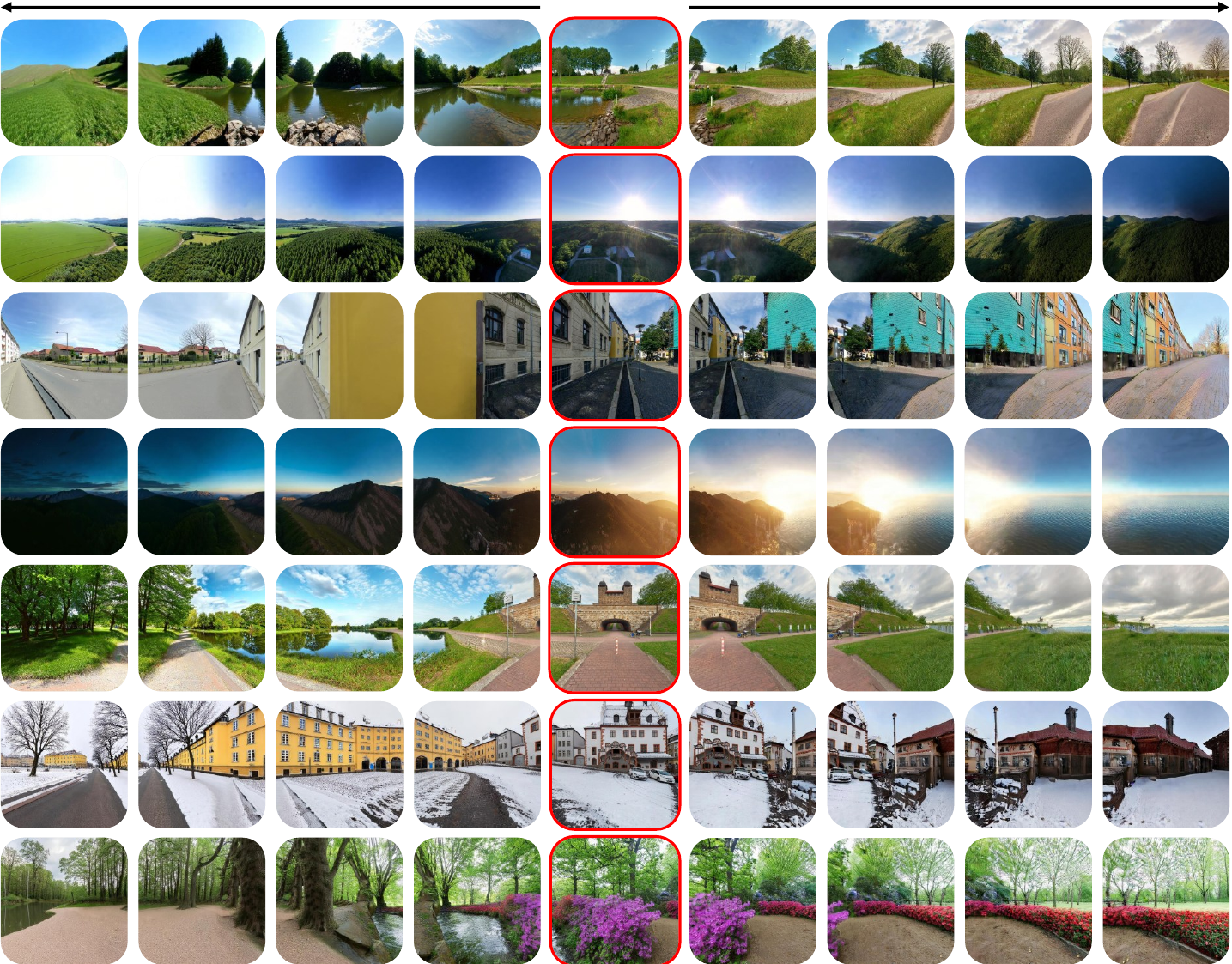}
    \end{subfigure}
    
       \caption{\textbf{Image-to-image generation (cross-view) with varying yaw angles.} The image with a red box denotes the initial view, and the others are the generated views based on the yaw deviation from the previous view.
    }
    \label{fig:cross_view_gen_iter}
    
\end{figure}

\begin{figure}[t]
    \centering
    \begin{subfigure}{.92\textwidth}
        \centering
        \includegraphics[width=1\linewidth]{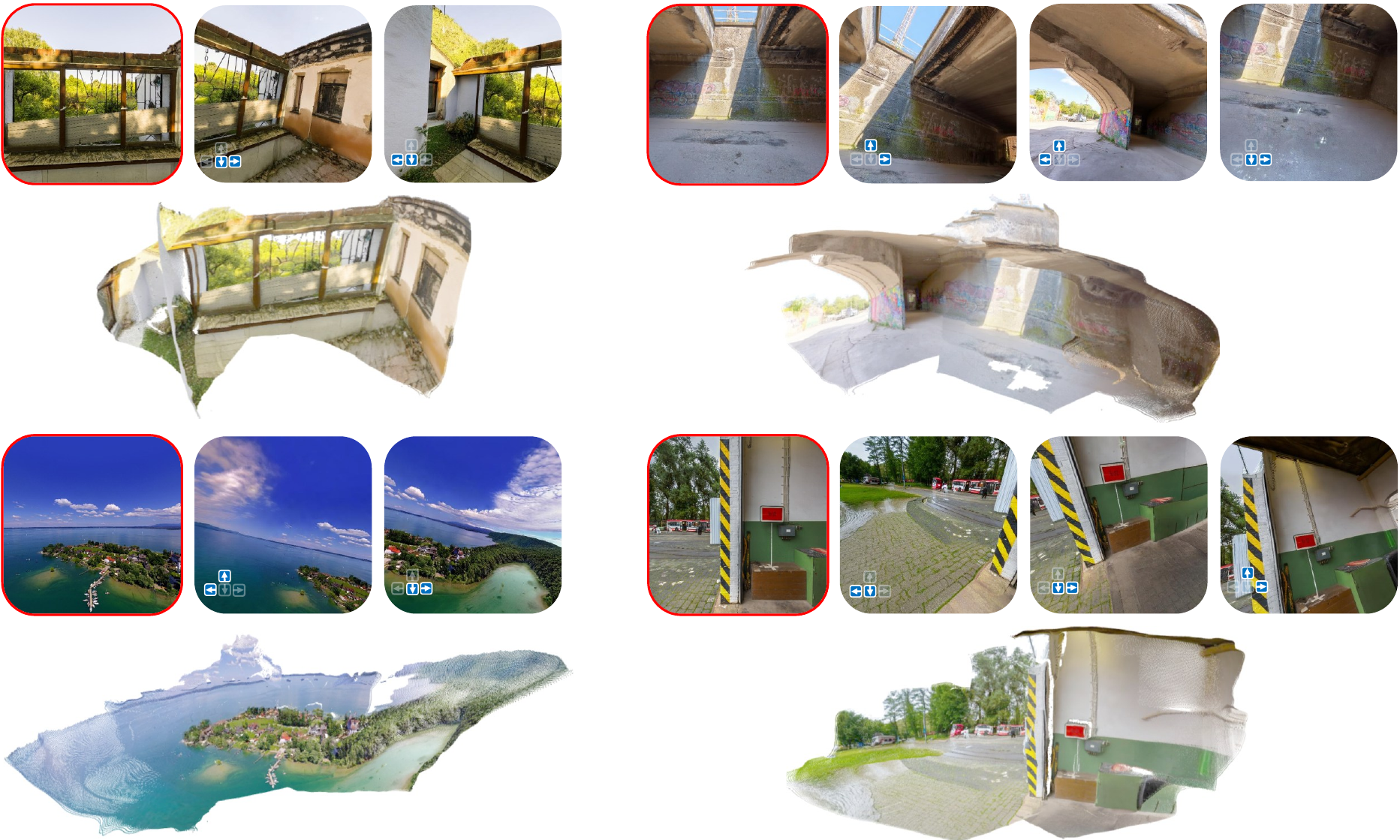}
    \end{subfigure}
    
    %\vspace{-0.1cm}
    
    \begin{subfigure}{.92\textwidth}
        \centering
        \includegraphics[width=1\linewidth]{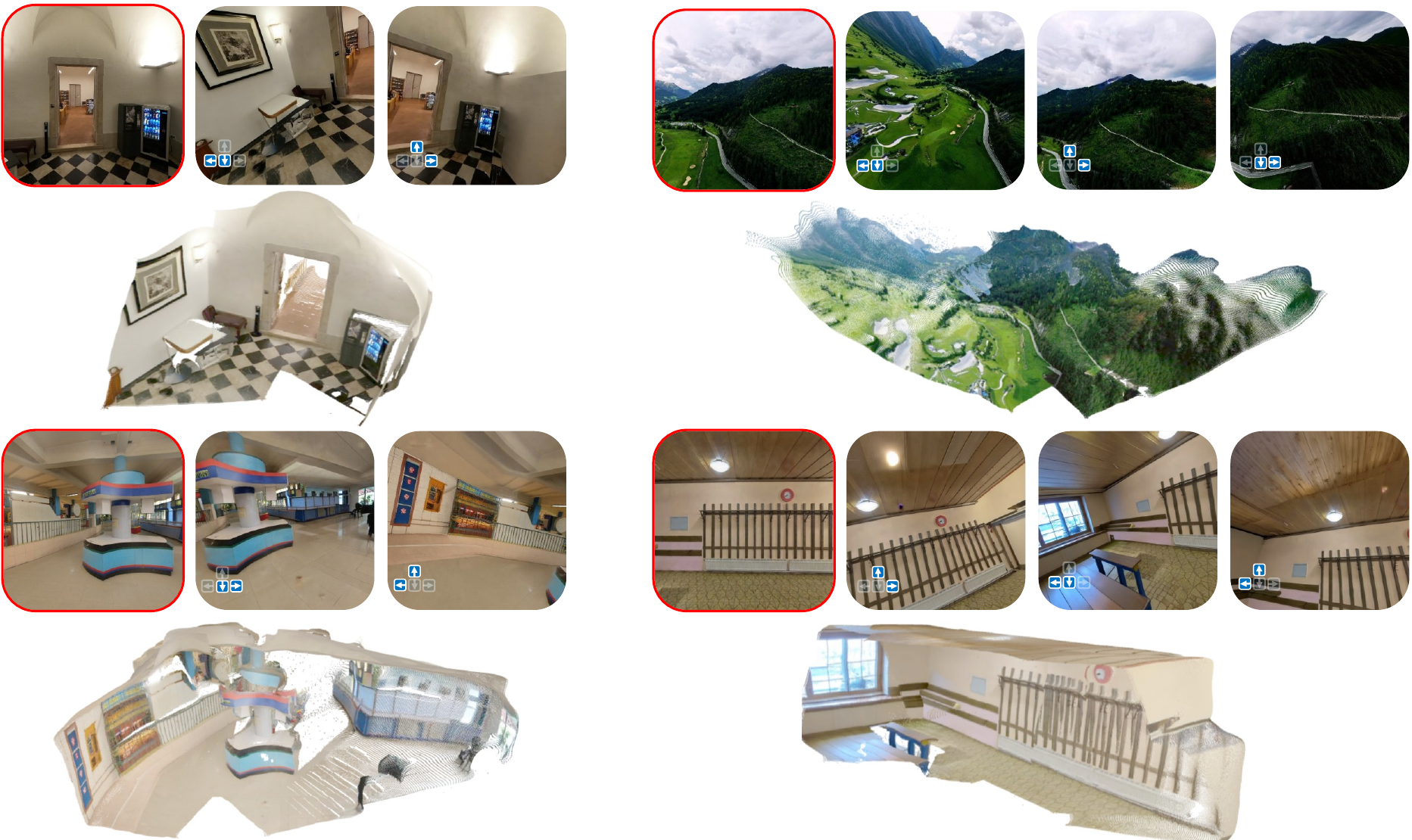}
    \end{subfigure}
    %\vspace{-0.1cm}
    
    \begin{subfigure}{.92\textwidth}
        \centering
        \includegraphics[width=1\linewidth]{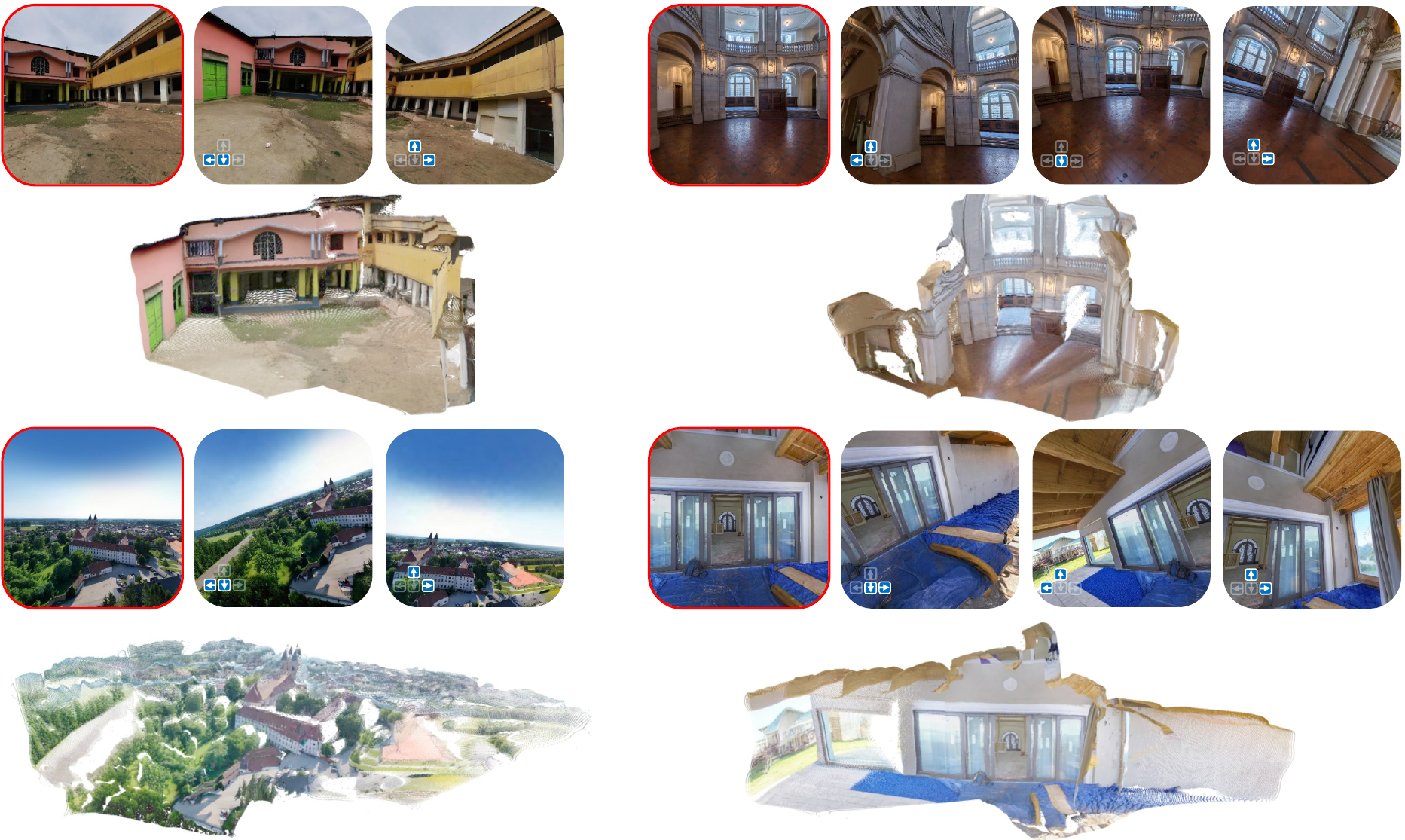}
    \end{subfigure}

       \caption{\textbf{World exploration results.} The 3D reconstruction results are obtained by VGGT.
    }
    \label{fig:cross_view_gen_vggt}
    
\end{figure}

\begin{figure}[t]
    \centering
    \begin{subfigure}{0.98\textwidth}
        \centering
        \includegraphics[width=0.98\linewidth]{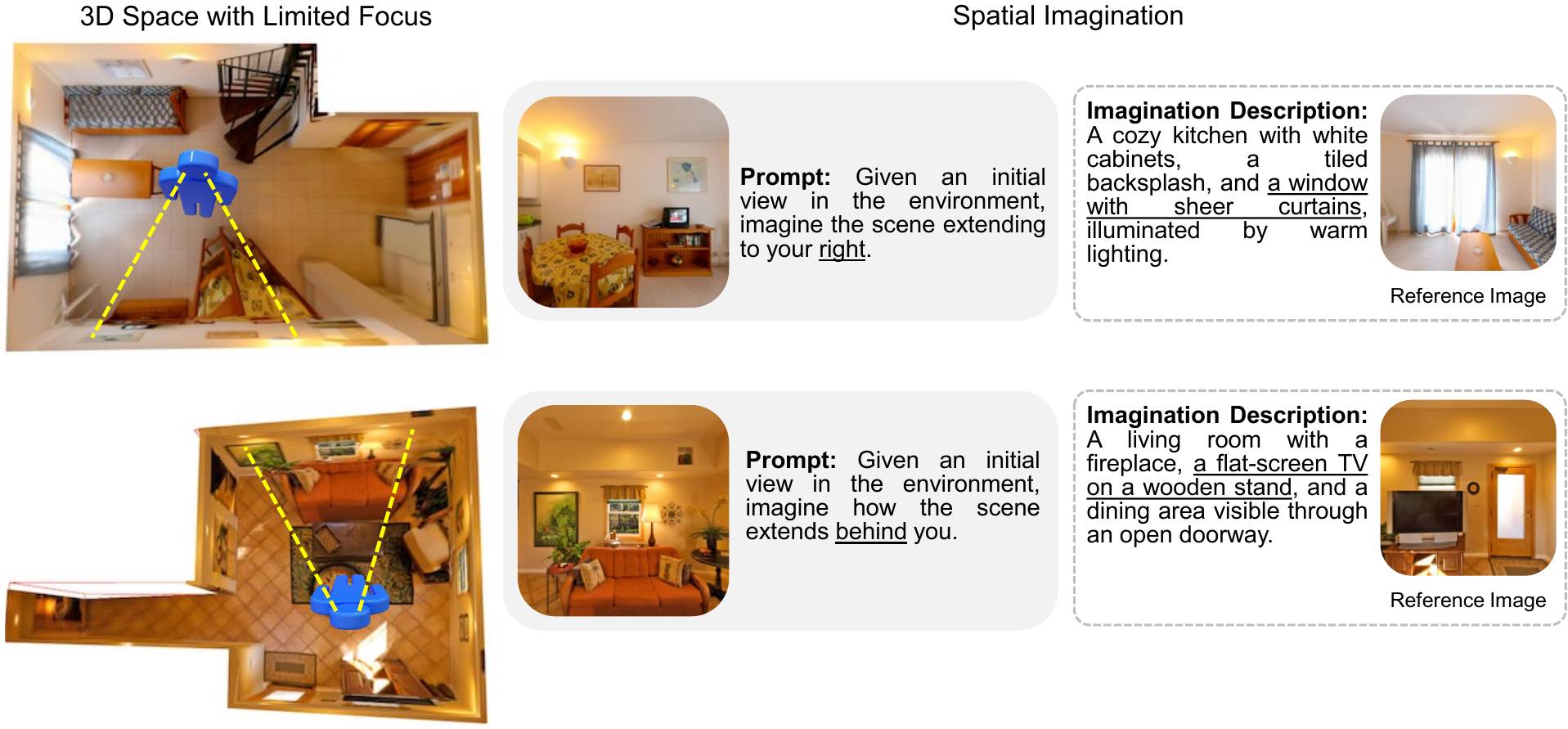}
    \end{subfigure}
    
    \vspace{-0.4cm}
    
    \begin{subfigure}{0.98\textwidth}
        \centering
        \includegraphics[width=0.98\linewidth]{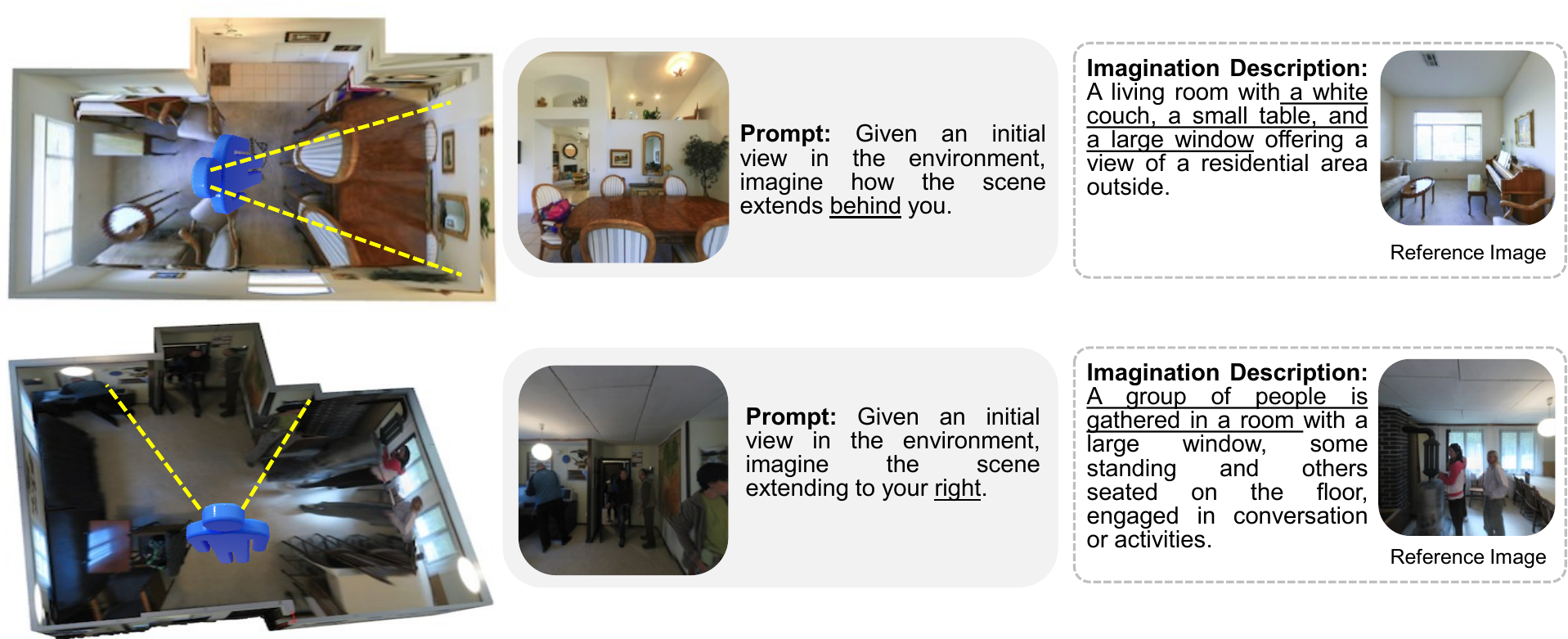}
        \caption{Spatial imagination. The plausible imagination results are \uline{highlighted}.}
    \end{subfigure}
    
    %\vspace{0.5em}
    
    \begin{subfigure}{0.98\textwidth}
        \centering
        \includegraphics[width=0.98\linewidth]{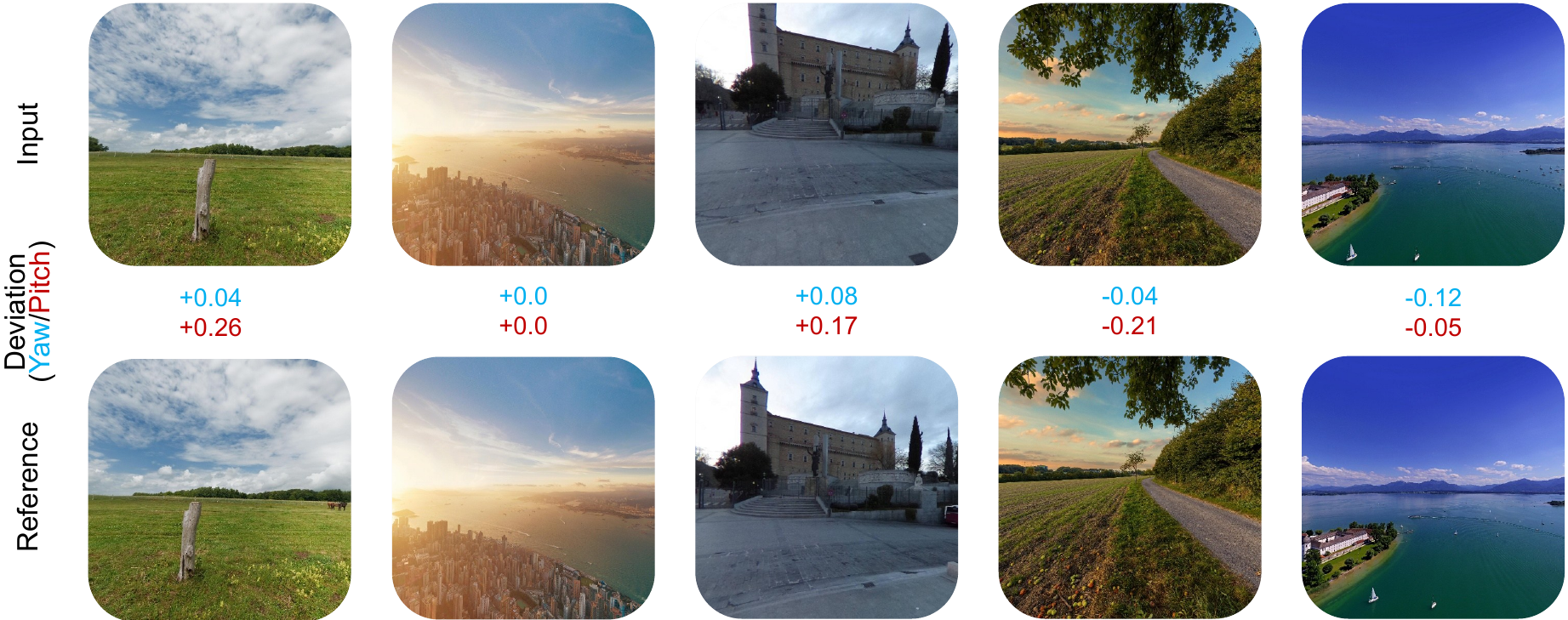}
    \end{subfigure}

    %\vspace{-0.1cm}
    
    \begin{subfigure}{0.98\textwidth}
        \centering
        \includegraphics[width=0.98\linewidth]{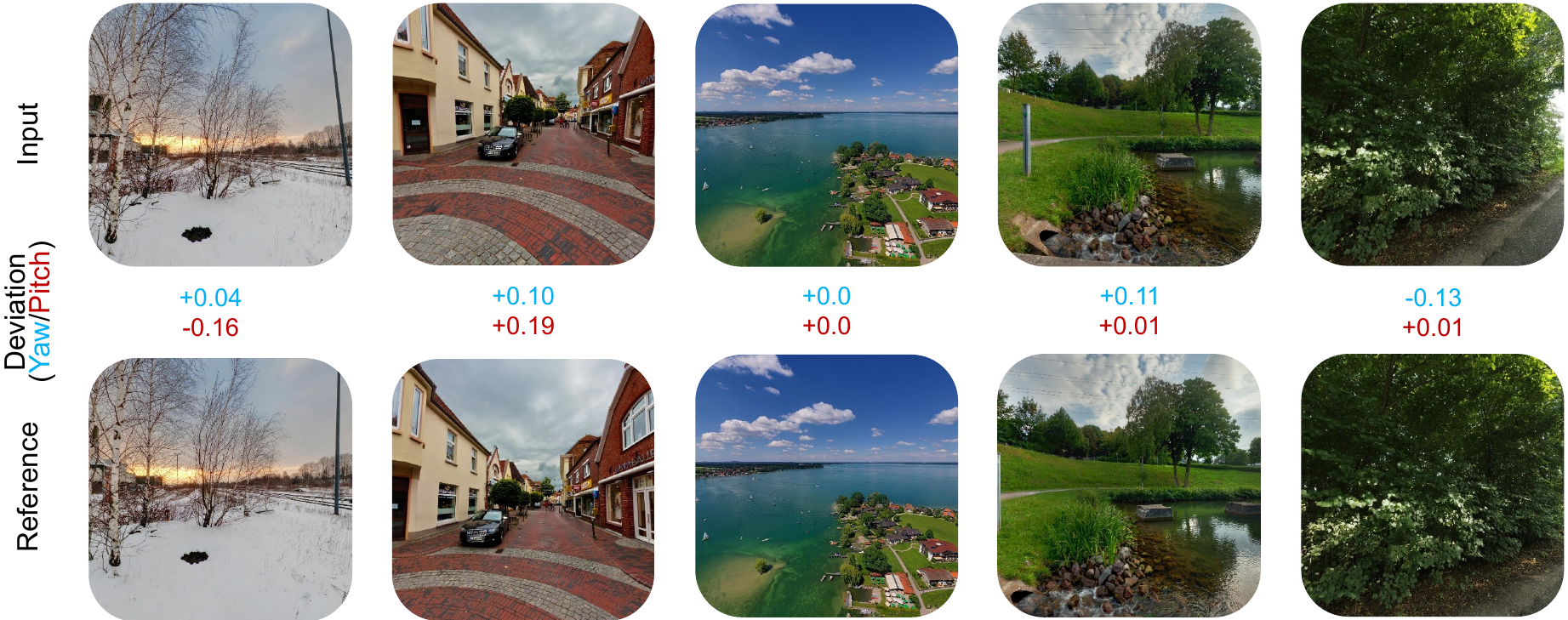}
        \caption{Photographic guidance. The suggested deviations of the camera parameters (\textcolor{cyan}{yaw}/\textcolor{darkred}{pitch}) are highlighted.}
        \label{fig:sub3}
    \end{subfigure}
    
    \caption{\textbf{Examples of the spatial imagination and photographic guidance.}}
    \label{fig:photo_spatial_imagination}
\end{figure}

We visualize more downstream application results by instruction tuning. Specifically, image-to-image generation (cross-view) results with varying yaw angles are shown in Figure~\ref{fig:cross_view_gen_iter}. World exploration results are provided in Figure~\ref{fig:cross_view_gen_vggt}. Examples of the spatial imagination and photographic guidance are shown in Figure~\ref{fig:photo_spatial_imagination}.

\newpage

\newpage

\end{document}